\documentclass[final,3p,times]{elsarticle}
\usepackage[T1]{fontenc}
\usepackage{geometry}
\usepackage{graphicx}

\usepackage{diagbox}
\usepackage{slashbox}

\usepackage{soul}
\usepackage{cancel}

\usepackage{enumitem}

\usepackage{commath}
\usepackage{amsthm}
\usepackage{amssymb}
\usepackage{amsmath}
\allowdisplaybreaks[1]
\usepackage{amsfonts}

\newtheorem{definition}{Definition}
\newtheorem{theorem}{Theorem}
\newtheorem{example}{Example}

\usepackage{hyperref}
\hypersetup{
colorlinks=true,
linkcolor=red,
anchorcolor=blue,
citecolor=green
}

\usepackage[dvipsnames,table]{xcolor}

\usepackage{multirow}

\usepackage{tikz}
\usetikzlibrary{shapes,arrows,matrix,positioning,patterns,through,calc,intersections}

\usepackage{subcaption}
\usepackage{subcaption}
\biboptions{sort&compress}

\newcommand*{\circled}[1]{\lower.7ex\hbox{\tikz\draw (0pt, 0pt) circle (.5em) node {\makebox[1em][c]{\small #1}};}}
\makeatletter
\newcommand{\bigcomp}{
  \DOTSB
  \mathop{\vphantom{\sum}\mathpalette\bigcomp@\relax}
  \slimits@
}
\newcommand{\bigcomp@}[2]{
  \begingroup\m@th
  \sbox\z@{$#1\sum$}
  \setlength{\unitlength}{0.9\dimexpr\ht\z@+\dp\z@}
  \vcenter{\hbox{
    \begin{picture}(1,1)
    \bigcomp@linethickness{#1}
    \put(0.5,0.5){\circle{0.3}}
    \end{picture}
  }}
  \endgroup
}
\newcommand{\bigcomp@linethickness}[1]{
  \linethickness{
      \ifx#1\displaystyle 2\fontdimen8\textfont\else
      \ifx#1\textstyle 1.65\fontdimen8\textfont\else
      \ifx#1\scriptstyle 1.65\fontdimen8\scriptfont\else
      1.65\fontdimen8\scriptscriptfont\fi\fi\fi 3
  }
}
\makeatother

\mathchardef\mhyphen="2D
\makeatletter
\newcommand*\bigcdot{\mathpalette\bigcdot@{.5}}
\newcommand*\bigcdot@[2]{\mathbin{\vcenter{\hbox{\scalebox{#2}{$\m@th#1\bullet$}}}}}
\makeatother

\journal{Elsevier}
\begin{document}
\begin{frontmatter}

\title{Three-Way Conflict Analysis Based on Alliance and Conflict Functions}
\author[A]{Junfang Luo}
\ead{junfangluo@163.com}
\author[B]{Mengjun Hu}
\ead{mengjun.hu@uregina.ca}
\author[C]{Guangming Lang\corref{cor1}}
\ead{langguangming1984@126.com}
\author[A]{Xin Yang}
\ead{yangxin@swufe.edu.cn}
\author[D]{Keyun Qin}
\ead{keyunqin@126.net}

\address[A]{School of Economic Information Engineering, Southwestern University of Finance and Economics, Chengdu, 611130, P.R. China}
\address[B]{Department of Computer Science, University of Regina, Regina, SK, S4S 0A2, Canada}
\address[C]{School of Mathematics and Statistics, Changsha University of Science and Technology, Changsha, 611756, P.R. China}
\address[D]{College of Mathematics, Southwest Jiaotong University, Chengdu 610031, P.R. China}
\cortext[cor1]{Corresponding author.}

\begin{abstract}
Trisecting agents, issues, and agent pairs are essential topics of three-way conflict analysis. They have been commonly studied based on either a rating or an auxiliary function. A rating function defines the positive, negative, or neutral ratings of agents on issues. An auxiliary function defines the alliance, conflict, and neutrality relations between agents. These functions measure two opposite aspects in a single function, leading to challenges in interpreting their aggregations over a group of issues or agents. For example, when studying agent relations regarding a set of issues, a standard aggregation takes the average of an auxiliary function concerning single issues. Therefore, a pair of alliance $+1$ and conflict $-1$ relations will produce the same result as a pair of neutrality 0 relations, although the attitudes represented by the two pairs are very different. To clarify semantics, we separate the two opposite aspects in an auxiliary function into a pair of alliance and conflict functions. Accordingly, we trisect the agents, issues, and agent pairs and investigate their applications in solving a few crucial questions in conflict analysis. Particularly, we explore the concepts of alliance sets and strategies. A real-world application is given to illustrate the proposed models.
\end{abstract}
\end{frontmatter}

\section{Introduction}
\label{sec:introduction}

Conflict analysis focuses on understanding and solving the conflict that arises from different attitudes held by agents or organizations towards a set of issues. Pawlak~\cite{Pawlak_1984,Pawlak_1998} proposes a three-valued situation table to formally represent the attitudes, opinions, or ratings of a set of agents on a set of issues. In particular, each cell in the table takes a value of $+1$, $-1$, or $0$, indicating that an agent is favourable/positive, against/negative, or neutral towards an issue. Based on such a situation table, Pawlak~\cite{Pawlak_1984,Pawlak_1998} further investigates the relations between agents, including the alliance, conflict, and neutrality relations. Following Pawlak's formulation, many researchers have studied conflict analysis models from different perspectives~\cite{Deja_2002,Jiang_2011,Jabbour_2017,Liu_2015,Pawlak_1984,Pawlak_2005,Silva_2016,Skowron_2002,Skowron_2006,Zhi_2019}.

The use of three ratings and three types of agent relations naturally links conflict analysis to the idea of thinking in threes that underlies the theory of three-way decision introduced by Yao~\cite{Yao_2012,Yao_2018,Yao_2021}. In a broad sense, three-way decision involves a philosophy of thinking in threes, a methodology of working in threes, and a mechanism of processing in threes~\cite{Yao_2021}. A popular concrete model is the Trisecting-Acting-Outcome (TAO) model~\cite{Yao_2018}, which involves dividing a whole into three parts, devising and applying strategies to process the three parts, and evaluating and optimizing the outcome. Three-way decision and its underlying philosophy of thinking in threes have been widely applied to many disciplines and fields~\cite{Luo_2020,Yao_Hu_2018,Zhang_2021,Deng_2014,Hu_2021}. Three-way classification~\cite{Hu_2014,Hu_2019,Maldonado_2020,Xu_2020,Yue_2020,Zhao_2020,Liu_2020} is one of the earliest studied specific topics of three-way decision. It focuses on trisecting a universe into positive instances, negative instances, and those in the boundary that cannot be determined to be positive or negative. Such a trisection is commonly based on the evaluation of objects. Yao~\cite{Yao_2012} summarizes two evaluation-based three-way classification models. One uses a single evaluation function to perform both the positive and negative evaluations, and the other uses two separate evaluation functions. The idea of computing with threes also appears in a few other topics, such as shadowed sets and bipolar representations. Pedrycz~\cite{Pedrycz_1998,Pedrycz_2005,Pedrycz_2009} proposes the concept of a shadowed set that transforms a fuzzy set into three disjoint elevated, reduced, and shadowed areas by applying two opposite operations of elevation and reduction. The idea of using three areas indicates a close connection between shadowed sets and three-way decision, which has been studied by a few researchers~\cite{Yang_2021,Yao_Wang_2017,Zhang_2020,Zhao_2019,Zhou_2021}. Dubois and Prade~\cite{Dubois_2008,Dubois_2012} propose the concept of bipolarity, which refers to the propensity of the human mind for reasoning and decision-making based on the two opposite positive and negative effects. Although bipolarity shares many common features with three-way decision, their connections are not well explored in the literature.

Conflict analysis is one of the many topics that share a common idea of using threes with three-way decision. A combination of the two topics leads to three-way conflict analysis that studies a conflict situation via trisecting agents, issues, and agent pairs. Specifically, most research on three-way conflict analysis~\cite{Lang_2020_unification,Lang_2020_general,Sun_2015,Sun_2016,Sun_2020,Fan_2018,Lang_2017,Lang_2019,Yao_2019,Li_2021} follows the three-way classification model based on a single evaluation function. Sun, Ma, and Zhao~\cite{Sun_2016} introduce a model of trisecting agents and issues by generalizing the concepts of lower and upper approximations over two universes in rough set theory. This work is further extended by Sun et al.~\cite{Sun_2020} based on the probabilistic rough set approximations over two universes. Lang, Miao, and Cai~\cite{Lang_2017} apply a pair of thresholds $(\alpha,\beta)$ on an evaluation function to trisect agent pairs instead of a single threshold of $0.5$ in Pawlak's model~\cite{Pawlak_1984,Pawlak_1998,Pawlak_2005}. Fan, Qi, and Wei~\cite{Fan_2018} propose a quantitative model of three-way conflict analysis based on an evaluation function generalized from derivation operators in formal concept analysis. Yao~\cite{Yao_2019} argues about the inconsistency of interpreting the neutral rating $0$ in Pawlak's model and discusses trisections of agents and agent pairs based on a single evaluation function.

A potential issue in most existing models is that they use a single evaluation function to reflect two opposite aspects. For instance, an auxiliary function commonly uses the positive value $+1$ for an alliance relation between two agents on a specific issue, the negative value $-1$ for a conflict relation, and the value 0 for a neutrality relation. This may lead to some difficulties in understanding and interpreting the computation of values from an auxiliary function. In particular, when investigating the relationships between agents on a set of issues, the average-based method is commonly used to aggregate the values from an auxiliary function on individual issues. Therefore, a pair of $+1$ and $-1$ gives the same result as a pair of 0 and 0. Consequently, two agents allied on one issue and in conflict on another have the same relation as two other agents that are neutral to each other on both issues. 

To solve the above semantic issues, we use two separate functions to manage the two opposite aspects represented by an auxiliary function. Specifically, we use an alliance function and a conflict function, respectively, to evaluate the alliance and conflict degrees between two agents on a single issue. When it comes to a set of issues, the alliance and conflict degrees on individual issues are aggregated separately into an aggregated alliance degree and an aggregated conflict degree. These two aggregated degrees are then considered together to define the final alliance, neutrality, and conflict relations between two agents. Intuitively, we define the alliance relation if the alliance degree suggests strong alliance and the conflict degree suggests non-conflict, the conflict relation if it is the other way around, and the neutrality relation otherwise. Many other crucial topics in conflict analysis could also be investigated via the two proposed functions. Particularly, we explore the definition of alliance sets and the description of their decisions based on the trisections of agent pairs. With the trisections of issues and agents, we present to formulate a strategy as a conjunction of issue-rating pairs and further explore the relationships between strategies and agents. With a separation of alliance and conflict functions, we can gain a more in-depth understanding of the relations regarding agents and issues in conflict analysis.

The remainder of this paper is organized as follows.
Section~\ref{sec:review} reviews basic concepts related to existing three-way conflict analysis with a single evaluation function, including a rating function and an auxiliary function. Section~\ref{sec:functions} presents the alliance and conflict functions that separate the two opposite aspects in an auxiliary function. Based on these two functions, we discuss the trisections of agents, issues, and agent pairs in Sections~\ref{sec:alliance_set_and_decision} and \ref{sec:strategy}. Particularly, Section~\ref{sec:alliance_set_and_decision} proposes the concept of alliance sets by trisecting agent pairs and investigates the representation of their decisions. Section~\ref{sec:strategy} presents a formal description of a strategy (i.e., a logic conjunction of issue-rating pairs) and further investigates the trisection of agents towards a specific strategy. In Section~\ref{sec:comparative}, we compare the proposed models with two evaluation functions and the existing models with a single evaluation function. An application with a real-world conflict situation is also given in Section~\ref{sec:comparative} to illustrate the proposed models. In the end, Section~\ref{sec:conclusion} concludes this work and discusses possible directions for future work.

\section{Three-way conflict analysis with a single evaluation function}
\label{sec:review}

This section reviews and summarizes basic concepts in three-way conflict analysis. Particularly, we discuss trisections with a single evaluation function, including trisections of agents, issues, and agent pairs. Examples are given at the end to illustrate the concepts.

\subsection{Trisecting agents and issues with a rating function}
\label{sec:review_rating}

A conflict problem is usually formulated by a three-valued situation table~\cite{Pawlak_1998,Yao_2019}, which gives the opinions, attitudes, or ratings of agents on issues. The attitudes are commonly considered to have three cases, namely, positive, negative, and neutral, represented by three numeric values of $+1$, $-1$, and 0, respectively. A three-valued situation table is formally defined as follows.

\begin{definition}
\label{def:situation_table}
A three-valued situation table is a triplet $S=(A,I,r)$,
where $A$ is a finite nonempty set of agents,
$I$ is a finite nonempty set of issues,
and $r:A\times I\rightarrow \{+1,-1,0\}$ is a rating function.
The value $r(x,i)$ is called the rating of an agent $x\in A$ on an issue $i\in I$ and interpreted as follows:
\begin{eqnarray*}
\left \{
\begin{array}{lcl} 
x {\rm~is~positive~on~} i, & \quad & {\rm iff~} r(x,i)=+1;\\
x {\rm~is~negative~on~} i, & \quad & {\rm iff~} r(x,i)=-1;\\
x {\rm~is~neutral~on~} i, & \quad & {\rm iff~} r(x,i)=0.\\
\end{array}
\right.
\end{eqnarray*}

\end{definition}

Yao~\cite{Yao_2019} extends a three-valued situation table to a many-valued situation table, which considers ratings from the interval $[-1,+1]$. We restrict our discussion to three-valued situation tables in this work. The generalization to many-valued situation tables might be a direction for our future work. According to the three ratings of agents on a specific issue, we can immediately construct a trisection of agents as follows.

\begin{definition}
\label{def:trisection_agent_i_existing}
For a subset of agents $X\subseteq A$ and an issue $i\in I$, the trisection of $X$ with respect to $i$ is defined as~\cite{Yao_2019}:
\begin{eqnarray}
X_i^+&=&\{x\in X\mid r(x,i)=+1\},\nonumber\\
X_i^-&=&\{x\in X\mid r(x,i)=-1\},\nonumber\\
X_i^0&=&\{x\in X\mid r(x,i)=0\}.
\end{eqnarray}
\end{definition}

The three sets $X_i^+$, $X_i^-$, and $X_i^0$ include the agents with a positive, negative, and neutral rating on a single issue $i$, respectively. When it comes to a set of issues, we need to aggregate the ratings of an agent on every single issue in the set. A commonly used method is to take the average~\cite{Yao_2019}. Formally, the rating of an agent $x \in A$ on a nonempty subset of issues $\emptyset\neq J\subseteq I$ is given by a rating function $r:~A\times (2^I - \{\emptyset\})\rightarrow [-1,+1]$ as follows:
\begin{eqnarray}
\label{equa:aggregated_rating_issues}
r(x,J)&=&\frac{\sum\limits_{i\in J}r(x,i)}{|J|},
\end{eqnarray}
where $2^I$ is the power set of $I$ and $|J|$ is the cardinality of $J$. It should be noted that we use the same symbol $r$ to represent different rating functions for simplicity. One may simply differentiate the functions by their parameters.

The aggregated rating $r(x,J)$ can be any value in the interval $[-1,+1]$. Accordingly, we may apply two thresholds on the aggregated ratings to construct a trisection of agents with respect to $J$.

\begin{definition}
\label{def:trisection_agent_J_existing}
For a subset of agents $X\subseteq A$ and a nonempty subset of issues $\emptyset\neq J\subseteq I$, the trisection of $X$ with respect to $J$ is defined as~\cite{Yao_2019}:
\begin{eqnarray}
X_J^{+}&=&\{x\in X\mid r(x,J)\geq h\},\nonumber\\
X_J^{-}&=&\{x\in X\mid r(x,J)\leq l\},\nonumber\\
X_J^{0}&=&\{x\in X\mid l<r(x,J)<h\},
\end{eqnarray}
where $(l,h)$ is a pair of thresholds satisfying $-1\leq l<0<h\leq +1$.
\end{definition}

Intuitively, if the aggregated rating $r(x,J)$ is high enough (i.e., $r(x,J)\geq h$), the agent $x$ is considered to have a positive attitude on $J$ as a whole. Similarly, if $r(x,J)$ is low enough (i.e., $r(x,J)\leq l$), $x$ is considered to have a negative attitude on $J$. Otherwise, $r(x,J)$ is neither strong enough to suggest a positive nor a negative attitude and thus, $x$ is considered to have a neutral attitude on $J$.

One can investigate the trisections of issues by simply exchanging the roles of agents and issues. Due to this reason, the trisections of issues are usually not explicitly explored in the literature. Although the trisections of issues follow the same formulations as the trisections of agents, they imply different meanings and views in understanding the conflict situation. Furthermore, the trisections of issues will also be investigated in our proposed models in the following sections. Thus, we briefly discuss the trisections of issues with respect to a single agent and a subset of agents. The trisection of issues with respect to a single agent can be obtained based on the ratings of agents.

\begin{definition}
\label{def:trisection_issue_x_existing}
For a subset of issues $J\subseteq I$ and an agent $x\in A$, the trisection of $J$ with respect to $x$ is defined as~\cite{Yao_2019}:
\begin{eqnarray}
J_x^+&=&\{i\in J\mid r(x,i)=+1\},\nonumber\\
J_x^-&=&\{i\in J\mid r(x,i)=-1\},\nonumber\\
J_x^0&=&\{i\in J\mid r(x,i)=0\}.
\end{eqnarray}
\end{definition}

With respect to a nonempty set of agents, we aggregate their ratings on a single issue as the average. Formally, the aggregated rating function $r: (2^A - \{\emptyset\}) \times I\rightarrow [-1,+1]$ is defined as: for $\emptyset \ne X \subseteq A$ and $i \in I$,
\begin{eqnarray}
\label{equa:aggregated_rating_agents}
r(X,i)&=&\frac{\sum\limits_{x\in X}r(x,i)}{|X|}.
\end{eqnarray}
Accordingly, one can trisect a set of issues as given in the following definition.

\begin{definition}
\label{def:trisection_issue_X_existing}
For a nonempty subset of agents $\emptyset\neq X\subseteq A$ and a subset of issues $J\subseteq I$, the trisection of $J$ with respect to $X$ is defined as:
\begin{eqnarray}
J_X^{+}&=&\{i\in J\mid r(X,i)\geq h\},\nonumber\\
J_X^{-}&=&\{i\in J\mid r(X,i)\leq l\},\nonumber\\
J_X^{0}&=&\{i\in J\mid l<r(X,i)<h\},
\end{eqnarray}
where $(l,h)$ is a pair of thresholds satisfying $-1\leq l<0<h\leq +1$.
\end{definition}

Intuitively, if the aggregated rating from a group of agents $X$ on an issue $i$ is high enough (i.e., $r(X,i)\geq h$), then the group $X$ as a whole is considered to have a positive attitude on $i$. Similarly, if the aggregated rating is low enough (i.e., $r(X,i)\leq l$), then the group has a negative attitude on $i$. Otherwise, the group has a neutral attitude on $i$.

\subsection{Trisecting agent pairs with an auxiliary function}
\label{sec:review_auxiliary}

Another essential topic of three-way conflict analysis is investigating the relationships between agents. Basically, there are three relationships between agents, namely, the alliance, conflict, and neutrality relations~\cite{Pawlak_1984}. These relations are defined based on the ratings of agents on either a single issue or a subset of issues. An auxiliary function $\Phi$ is commonly used to help formulate the relations.
Intuitively, a positive value of $\Phi$ represents the tendency of alliance, and a negative value represents the tendency of conflict.

Formally, an auxiliary function with respect to a single issue $i\in I$ is a three-valued mapping $\Phi_i:~A\times A\rightarrow \{+1,-1,0\}$, where the three values are interpreted as follows:
\begin{eqnarray*}
\left \{
\begin{array}{lcl} 
x {\rm~and~} y {\rm~are~allied~on~} i, & \quad & {\rm iff~} \Phi_i(x,y)=+1;\\
x {\rm~and~} y {\rm~are~in~conflict~on~} i, & \quad & {\rm iff~} \Phi_i(x,y)=-1;\\
x {\rm~and~} y {\rm~are~neutral~on~} i, & \quad & {\rm iff~} \Phi_i(x,y)=0.\\
\end{array}
\right.
\end{eqnarray*}
The neutrality relation is simply interpreted as the two agents being neither allied nor in conflict. Accordingly, one can formally define a trisection of agent pairs with respect to a single issue.

\begin{definition}
\label{def:trisection_agentpair_i_existing}
Given an auxiliary function $\Phi_i$ with respect to an issue $i\in I$, the alliance relation $R^=_i$, conflict relation $R_i^{\asymp}$, and neutrality relation $R_i^{\approx}$ between agents with respect to $i$ are defined as:
\begin{eqnarray}
R^=_i&=&\{(x,y)\in A\times A\mid \Phi_i(x,y)=+1\},\nonumber\\
R^{\asymp}_i&=&\{(x,y)\in A\times A\mid \Phi_i(x,y)=-1\},\nonumber\\
R^{\approx}_i&=&\{(x,y)\in A\times A\mid \Phi_i(x,y)=0\}.
\end{eqnarray}
\end{definition}

There are different specific definitions of $\Phi_i$ as will be illustrated in Example~\ref{example:auxiliary_PY}. Commonly, the following properties should be satisfied by a meaningful auxiliary function $\Phi_i$:
\begin{eqnarray*}
(1)&&\Phi_i(x,x)=+1,\\
(2)&&\Phi_i(x,y)=\Phi_i(y,x).
\end{eqnarray*}
That is, an agent is always self-allied, and the value for $x$ and $y$ is equal to the value for $y$ and $x$. The latter implies that the alliance, conflict, and neutrality relations are symmetric. In addition, it is reasonable to consider two agents to be allied if they both have a positive or negative attitude on the issue $i$. Similarly, it is reasonable to consider two agents to be in conflict if one of them is positive and the other is negative. In the case where at least one agent holds a neutral rating, there are different views of defining the relations based on different understanding and interpretations of neutral ratings. From the above analysis, we define a general template of $\Phi_i$ as follows: for $x,y\in A$,
\begin{eqnarray}
\label{equa:auxiliary_ixy_general}
\Phi_i(x,y)&=&
\left \{
\begin{array}{lcl}
+1, & &r(x,i)\cdot r(y,i)=+1~\mbox{or}~x=y,\\
-1,& &r(x,i)\cdot r(y,i)=-1,\\
*,& &otherwise,\\
\end{array}
\right.
\end{eqnarray}
where the notation $*$ represents a value in $\{+1,-1,0\}$. Since $\Phi_i(x,y)$ is actually defined with respect to the ratings from $x$ and $y$, we may alternatively use the notation $\Phi(r(x,i),r(y,i))$ when we do not concern about the specific agents who hold the ratings. Table~\ref{tab:auxiliary_ixy_general} shows the values of $\Phi_{i}$ in the general template for two different agents.
\begin{table}[!ht]
\centering
\setlength{\tabcolsep}{1em}
\renewcommand{\arraystretch}{1.2}
\begin{tabular}{c|ccc}
\hline
\diagbox[width=6em,height=3em]{$r(x,i)$}{$r(y,i)$}&$+1$& $-1$    & $0$  \\ \hline
$+1$    &$+1$&$-1$&$*$\\
$-1$    &$-1$ &$+1$&$*$\\
$0$     &$*$&$*$&$*$\\
\hline
\end{tabular}
\caption{The values of $\Phi(r(x,i),r(y,i))$ with $x\neq y$}
\label{tab:auxiliary_ixy_general}
\end{table}

We can generalize the auxiliary function in Equation~\eqref{equa:auxiliary_ixy_general} with respect to a set of issues by simply taking the average. Formally, an auxiliary function $\Phi_J:~A\times A\rightarrow [-1,+1]$ with respect to a subset of issues $J\subseteq I$ can be defined as follows \cite{Pawlak_1998,Yao_2019}:
\begin{eqnarray}
\label{equa:auxiliary_Jxy_general}
&&\Phi_J(x,y)=
\left \{
\begin{array}{lcl}
\frac{\sum\limits_{i\in J}\Phi_i(x,y)}{|J|}, & &J \neq \emptyset,\\
& &\\
0,& &J= \emptyset.\\
\end{array}
\right.
\end{eqnarray}
While $\Phi_i$ is a three-valued function, $\Phi_J$ is a many-valued function that takes values in the interval $[-1,+1]$. Applying two thresholds on $\Phi_J$, one can immediately construct a trisection of agent pairs and accordingly, define the alliance, conflict, and neutrality relations.

\begin{definition}
\label{def:trisection_agentpair_J_existing}
Given an auxiliary function $\Phi_J$ with respect to a subset of issues $J\subseteq I$, the alliance relation $R_J^{=}$, conflict relation $R_J^{\asymp}$, and neutrality relation $R_J^{\approx}$ with respect to $J$ are defined as:
\begin{eqnarray}
R_J^{=}&=&\{(x,y)\in A\times A\mid \Phi_J(x,y)\geq h\},\nonumber\\
R_J^{\asymp}&=&\{(x,y)\in A\times A\mid \Phi_J(x,y)\leq l\},\nonumber\\
R_J^{\approx}&=&\{(x,y)\in A\times A\mid l<\Phi_J(x,y)<h\},
\end{eqnarray}
where $(l,h)$ is a pair of thresholds satisfying $-1\leq l<0<h\leq +1$.
\end{definition}
Intuitively, two agents $x$ and $y$ are allied on $J$ if they have a high tendency for alliance (i.e., $\Phi_J(x,y)\geq h$). They are in conflict if their tendency for conflict is strong enough (i.e., $\Phi_J(x,y)\leq l$). Otherwise, they are neutral on $J$.

Aggregating ratings with respect to a group of issues or agents leads to continuous values from the interval $[-1,+1]$ instead of the three discrete values $+1$, $-1$, and $0$. Accordingly, the trisections with respect to a group of issues or agents need a pair of thresholds $(l,h)$ to cut the continuous aggregated values, as in Definitions~\ref{def:trisection_agent_J_existing}, \ref{def:trisection_issue_X_existing}, and \ref{def:trisection_agentpair_J_existing}. This idea is also commonly used in other topics to transform a set of continuous quantitative values into several qualitative parts or levels. For example, a shadowed set~\cite{Pedrycz_1998,Pedrycz_2005,Pedrycz_2009} adopts a similar idea to transform a fuzzy set with continuous membership degrees in the unit interval $[0,1]$ into three qualitative levels of membership degrees. The transformation is performed through two operations of elevation and reduction with a pair of thresholds. Particularly, the elevation operation elevates the membership degrees at or above one threshold to 1, and the reduction operation reduces the membership degrees at or below another threshold to 0. The membership degrees between the two thresholds are mapped to the whole unit interval $[0,1]$. A shadowed set can also be considered as a three-way approximation of fuzzy sets~\cite{Yao_Wang_2017}. The trisections in Definitions~\ref{def:trisection_agent_J_existing}, \ref{def:trisection_issue_X_existing}, and \ref{def:trisection_agentpair_J_existing} can also be explained through similar elevation and reduction operations. We elevate an evaluation value from a rating function (or an auxiliary function) to $+1$ that represents a positive attitude (or an alliance relation) if it is higher than or equal to a threshold $h$. Similarly, we reduce an evaluation value from a rating function (or an auxiliary function) to $-1$ that represents a negative attitude (or a conflict relation) if it is lower than or equal to another threshold $l$. For a value in-between the two thresholds, we either elevate or reduce it to the neutral value 0. A similar idea of dealing with values between two thresholds is also discussed for shadowed sets by Yao, Wang, and Deng~\cite{Yao_Wang_2017}, where they either elevate or reduce the values to a middle membership degree of 0.5. It is an interesting direction of our future work to further investigate the relations between conflict analysis and shadowed sets in this regard.

\subsection{Examples}

We give two examples to illustrate the concepts discussed in this section. Particularly, Example~\ref{example:auxiliary_PY} illustrates specific auxiliary functions that fit into our general template in Equation~\eqref{equa:auxiliary_ixy_general} and Example~\ref{example:trisections_review} illustrates the trisections.

\begin{example}[\textbf{Pawlak's and Yao's auxiliary functions}]
\label{example:auxiliary_PY} 

We have proposed a general definition of an auxiliary function in Equation~\eqref{equa:auxiliary_ixy_general}. Here, we review two specific auxiliary functions proposed by Pawlak~\cite{Pawlak_1998} and Yao~\cite{Yao_2019} in terms of our general definition. Concretely, for any two agents $x,y\in A$ and an issue $i\in I$, Pawlak defines the auxiliary function $\Phi_i^P: A\times A\longrightarrow \{+1,-1,0\}$ as: 
\begin{eqnarray}
\label{equa:auxiliary_ixy_Pawlak}
\Phi_i^P(x,y)&=&
\left \{
\begin{array}{lcl}
+1, & &r(x,i)\cdot r(y,i)=+1~\mbox{or}~x=y,\\
-1,& &r(x,i)\cdot r(y,i)=-1,\\
0,& &r(x,i)\cdot r(y,i)=0~\mbox{and}~x\neq y.\\
\end{array}
\right.
\end{eqnarray}
Instead of defining an auxiliary function, Yao proposes a distance function regarding two agents on a single issue. An auxiliary function can be easily defined based on the distance function. For $x,y\in A$ and $i\in I$, the Yao's auxiliary function $\Phi_i^Y: A\times A\longrightarrow \{+1,-1,0\}$ can be defined as:
\begin{eqnarray}
\label{equa:auxiliary_ixy_Yao}
\Phi_i^Y(x,y)&=&\left \{
\begin{array}{lcl}
+1, & &r(x,i)=r(y,i),\\
-1,& &r(x,i)\cdot r(y,i)=-1,\\
0,&&r(x,i)\cdot r(y,i)=0~\mbox{and}~r(x,i)\neq r(y,i).\\
\end{array}
\right.
\end{eqnarray}
\begin{table}[!ht]
\centering
\begin{subtable}{.45\linewidth}
\centering
\setlength{\tabcolsep}{1em}
\renewcommand{\arraystretch}{1.2}
\begin{tabular}{c|ccc}
\hline
\diagbox[width=6em,height=3em]{$r(x,i)$}{$r(y,i)$}& $+1$& $-1$    & $0$  \\ \hline
$+1$    &$+1$&$-1$&$0$\\
$-1$    &$-1$ &$+1$&$0$\\
$0$     &$0$&$0$&$0$\\
\hline
\end{tabular}
\caption{$\Phi^P(r(x,i),r(y,i))$ with $x\neq y$}
\end{subtable}
~
\begin{subtable}{.45\linewidth}
\centering
\setlength{\tabcolsep}{1em}
\renewcommand{\arraystretch}{1.2}
\begin{tabular}{c|ccc}
\hline
\diagbox[width=6em,height=3em]{$r(x,i)$}{$r(y,i)$}& $+1$& $-1$    & $0$  \\ \hline
$+1$&$+1$&$-1$&$0$\\
$-1$&$-1$ &$+1$&$0$\\
$0$&$0$&$0$&$+1$\\
\hline
\end{tabular}
\caption{$\Phi^Y(r(x,i),r(y,i))$ with $x\neq y$}
\end{subtable}
\caption{The values of $\Phi^P(r(x,i),r(y,i))$ and $\Phi^Y(r(x,i),r(y,i))$ with $x\neq y$}
\label{tab:auxiliary_ixy_PY}
\end{table}

To clearly show the difference between Pawlak's and Yao's auxiliary functions, Table~\ref{tab:auxiliary_ixy_PY} gives the values of $\Phi_i^P$ and $\Phi_i^Y$ for two different agents. These two auxiliary functions differ in the understanding of the relations between two different agents with both neutral ratings on an issue $i$. Concretely, for two different agents $x,y\in A$ with $r(x,i)=r(y,i)=0$, in Pawlak's opinion, they are neutral on $i$, that is, $\Phi^P(r(x,i),r(x,i))=0$; in Yao's opinion, they are allied on $i$, that is, $\Phi^Y(r(x,i),r(x,i))=+1$.

With respect to a set of issues $J\subseteq I$, we could apply the general template in Equation~\eqref{equa:auxiliary_Jxy_general} with Pawlak's auxiliary function in Equation~\eqref{equa:auxiliary_ixy_Pawlak} and get the following aggregated auxiliary function:
\begin{equation}
\label{equa:auxiliary_Jxy_Pawlak}
\Phi_J^P(x,y)=
\left \{
\begin{array}{lcl}
\frac{\sum\limits_{i\in J}\Phi_i^P(x,y)}{|J|}, & &J \neq \emptyset,\\
& &\\
0,& &J= \emptyset.\\
\end{array}
\right.\\ 
\end{equation}
Similarly, we can define Yao's auxiliary function with respect to $J$ as:
\begin{equation}
\label{equa:auxiliary_Jxy_Yao}
\Phi_J^Y(x,y)=
\left \{
\begin{array}{lcl}
\frac{\sum\limits_{i\in J}\Phi_i^Y(x,y)}{|J|}, & &J \neq \emptyset,\\
& &\\
0,& &J= \emptyset.\\
\end{array}
\right.
\end{equation}

\end{example}

\begin{example}[\textbf{Trisections of agents, issues, and agent pairs}]
\label{example:trisections_review} 

We illustrate the trisections of agents and issues using the rating functions $r$ defined in Section \ref{sec:review_rating}, and the trisection of agent pairs using Pawlak's and Yao's auxiliary functions given in the above Example \ref{example:auxiliary_PY}. Consider the three-valued situation table given in Table~\ref{tab:situation_table_example1}. For simplicity, we consider a set of agents $X=A=\{x_1,x_2,x_3,x_4,x_5,x_6\}$ and a set of issues $J=I=\{i_1,i_2,i_3,i_4,i_5\}$ when constructing the trisections.

\begin{table}[ht]
\centering
\begin{tabular}{c|ccccc}
\hline
\diagbox[width=3em,height=2em]{$A$}{$I$}&$i_1$&$i_2$&$i_3$&$i_4$&$i_5$\\
\hline
$x_1$&$+1$&0&0&0&0\\
$x_2$&$-1$&0&0&0&0\\
$x_3$&$+1$ &$+1$ &$+1$ &$+1$&$+1$ \\
$x_4$&$-1$ &$+1$ &$+1$ &$+1$&$+1$\\
$x_5$&$+1$ &$+1$ &$+1$ &$-1$&$-1$\\
$x_6$&$+1$ &$+1$ &$-1$ &$-1$&$-1$\\
\hline
\end{tabular}
\caption{A three-valued situation table}
\label{tab:situation_table_example1}
\end{table}

Firstly, we construct the trisection of agents $X$ with respect to $J$. According to the aggregated rating function in Equation~\eqref{equa:aggregated_rating_issues}, we compute the ratings of every agent in $X$ on $J$ as:
\begin{eqnarray}
\label{equa:example_rating_aggregated_J}
r(x_1,J) &=& \frac{(+1)+0+0+0+0}{5} = +\frac{1}{5},\nonumber\\
r(x_2,J) &=& \frac{(-1)+0+0+0+0}{5} = -\frac{1}{5},\nonumber\\
r(x_3,J) &=& \frac{(+1)+(+1)+(+1)+(+1)+(+1)}{5} = +1,\nonumber\\
r(x_4,J) &=& \frac{(-1)+(+1)+(+1)+(+1)+(+1)}{5} = +\frac{3}{5},\nonumber\\
r(x_5,J) &=& \frac{(+1)+(+1)+(+1)+(-1)+(-1)}{5} = +\frac{1}{5},\nonumber\\
r(x_6,J) &=& \frac{(+1)+(+1)+(-1)+(-1)+(-1)}{5} = -\frac{1}{5}.
\end{eqnarray}
By applying a pair of thresholds $(-\frac{3}{5},+\frac{3}{5})$, we get the trisection of $X$ according to Definition~\ref{def:trisection_agent_J_existing} as:
\begin{eqnarray}
X_J^{+}&=&\{x\in X\mid r(x,J)\geq +\frac{3}{5}\} = \{x_3,x_4\},\nonumber\\
X_J^{-}&=&\{x\in X\mid r(x,J)\leq -\frac{3}{5}\} = \emptyset,\nonumber\\
X_J^{0}&=&\{x\in X\mid -\frac{3}{5}<r(x,J)<+\frac{3}{5}\} = \{x_1,x_2,x_5,x_6\}.
\end{eqnarray}

Secondly, we construct the trisection of issues $J$ with respect to $X$. According to the aggregated rating function in Equation~\eqref{equa:aggregated_rating_agents}, we compute the ratings of agents in $X$ as a whole on every issue in $J$ as:
\begin{eqnarray}
r(X,i_1) &=& \frac{(+1)+(-1)+(+1)+(-1)+(+1)+(+1)}{6} = +\frac{1}{3}, \nonumber\\
r(X,i_2) &=& \frac{0+0+(+1)+(+1)+(+1)+(+1)}{6} = +\frac{2}{3}, \nonumber\\
r(X,i_3) &=& \frac{0+0+(+1)+(+1)+(+1)+(-1)}{6} = +\frac{1}{3}, \nonumber\\
r(X,i_4) &=& \frac{0+0+(+1)+(+1)+(-1)+(-1)}{6} = 0, \nonumber\\
r(X,i_5) &=& \frac{0+0+(+1)+(+1)+(-1)+(-1)}{6} = 0.
\end{eqnarray}
By applying a pair of thresholds $(-\frac{1}{3},+\frac{1}{3})$, we get the trisection of $J$ according to Definition~\ref{def:trisection_issue_X_existing} as:
\begin{eqnarray}
J_X^{+}&=&\{i\in J\mid r(X,i)\geq +\frac{1}{3}\}=\{i_1,i_2,i_3\},\nonumber\\
J_X^{-}&=&\{i\in J\mid r(X,i)\leq -\frac{1}{3}\}=\emptyset,\nonumber\\
J_X^{0}&=&\{i\in J\mid -\frac{1}{3}<r(X,i)<+\frac{1}{3}\}=\{i_4,i_5\}.
\end{eqnarray}

Finally, we construct the trisection of agent pairs using Pawlak's and Yao's auxiliary functions. According to Equations~\eqref{equa:auxiliary_Jxy_Pawlak} and \eqref{equa:auxiliary_Jxy_Yao}, we compute Pawlak's and Yao's auxiliary functions with respect to $J$ as given in Table~\ref{tab:auxiliary_Jxy_PY}. For example, for the two agents $x_1$ and $x_2$, we have:
\begin{eqnarray}
&&\Phi_{J}^P(x_1,x_2)=\frac{\sum\limits_{i\in J}\Phi_i^P(x_1,x_2)}{|J|}=\frac{(-1)+0+0+0+0}{5}=-\frac{1}{5};\nonumber\\
&&\Phi_{J}^Y(x_1,x_2)=\frac{\sum\limits_{i\in J}\Phi_i^Y(x_1,x_2)}{|J|}=\frac{(-1)+(+1)+(+1)+(+1)+(+1)}{5}=+\frac{3}{5}.
\end{eqnarray}

\begin{table}[ht!]
\centering
\renewcommand\arraystretch{1.2}
\setlength{\tabcolsep}{1.5mm}
\scalebox{1}{
\begin{tabular}{c|rr|rr|rr|rr|rr|rr}
\hline
  \multirow{2}{*}{}&\multicolumn{2}{c|}{$x_1$}&\multicolumn{2}{c|}{$x_2$}&\multicolumn{2}{c|}{$x_3$}&\multicolumn{2}{c|}{$x_4$}&
  \multicolumn{2}{c|}{$x_5$}&
  \multicolumn{2}{c}{$x_6$}\\\cline{2-13}
&$\Phi_{J}^P$&$\Phi_{J}^Y$&$\Phi_{J}^P$&$\Phi_{J}^Y$&$\Phi_{J}^P$&$\Phi_{J}^Y$&$\Phi_{J}^P$&$\Phi_{J}^Y$ &$\Phi_{J}^P$&$\Phi_{J}^Y$&$\Phi_{J}^P$&$\Phi_{J}^Y$\\
\hline
$x_1$   &+1 &+1 &$-\frac{1}{5}$ &$+\frac{3}{5}$  &$+\frac{1}{5}$ &$+\frac{1}{5}$ &$-\frac{1}{5}$ &$-\frac{1}{5}$ &$+\frac{1}{5}$ &$+\frac{1}{5}$ &$+\frac{1}{5}$ &$+\frac{1}{5}$\\
$x_2$   &$-\frac{1}{5}$ &$+\frac{3}{5}$  &+1   &+1   &$-\frac{1}{5}$ &$-\frac{1}{5}$ &$+\frac{1}{5}$ &$+\frac{1}{5}$  &$-\frac{1}{5}$ &$-\frac{1}{5}$  &$-\frac{1}{5}$ &$-\frac{1}{5}$\\
$x_3$   &$+\frac{1}{5}$ &$+\frac{1}{5}$   &$-\frac{1}{5}$ &$-\frac{1}{5}$  &+1    &+1  &$+\frac{3}{5}$ &$+\frac{3}{5}$ &$+\frac{1}{5}$ &$+\frac{1}{5}$&$-\frac{1}{5}$ &$-\frac{1}{5}$\\
$x_4$   &$-\frac{1}{5}$ &$-\frac{1}{5}$ &$+\frac{1}{5}$ &$+\frac{1}{5}$ &$+\frac{3}{5}$ &$+\frac{3}{5}$ &$+1$  &$+1$  &$-\frac{1}{5}$ &$-\frac{1}{5}$ &$-\frac{3}{5}$ &$-\frac{3}{5}$\\
$x_5$   &$+\frac{1}{5}$ &$+\frac{1}{5}$   &$-\frac{1}{5}$ &$-\frac{1}{5}$  &$+\frac{1}{5}$ &$+\frac{1}{5}$ &$-\frac{1}{5}$ &$-\frac{1}{5}$ &$+1$&$+1$&$+\frac{3}{5}$ &$+\frac{3}{5}$\\
$x_6$   &$+\frac{1}{5}$ &$+\frac{1}{5}$   &$-\frac{1}{5}$ &$-\frac{1}{5}$  &$-\frac{1}{5}$ &$-\frac{1}{5}$ &$-\frac{3}{5}$ &$-\frac{3}{5}$ &$+\frac{3}{5}$ &$+\frac{3}{5}$&$+1$&$+1$\\
\hline
\end{tabular}}
\caption{The values of auxiliary functions $\Phi_{J}^P$ and $\Phi_{J}^Y$}
\label{tab:auxiliary_Jxy_PY}
\end{table}

By applying a pair of thresholds $(-\frac{1}{2},+\frac{1}{2})$, we construct the trisections of agent pairs according to Definition~\ref{def:trisection_agentpair_J_existing}. The trisection based on Pawlak's auxiliary function in Table~\ref{tab:auxiliary_Jxy_PY} is constructed as:
\begin{eqnarray}
R_J^{=P}&=&\{(x,y)\in A\times A\mid \Phi_J^P(x,y)\geq +\frac{1}{2}\},\nonumber\\
&=& \{(x_1,x_1),(x_2,x_2),(x_3,x_3),(x_3,x_4),(x_4,x_3),(x_4,x_4),(x_5,x_5),(x_5,x_6),(x_6,x_5),(x_6,x_6)\},\nonumber\\
R_J^{\asymp P}&=&\{(x,y)\in A\times A\mid \Phi_J^P(x,y)\leq -\frac{1}{2}\},\nonumber\\
&=& \{(x_4,x_6),(x_6,x_4)\},\nonumber\\
R_J^{\approx P}&=&\{(x,y)\in A\times A\mid -\frac{1}{2}<\Phi_J^P(x,y)<+\frac{1}{2}\},\nonumber\\
&=& \{(x_1,x_2),(x_1,x_3),(x_1,x_4),(x_1,x_5),(x_1,x_6),(x_2,x_1),(x_2,x_3),(x_2,x_4),(x_2,x_5),(x_2,x_6),\nonumber\\
&& ~(x_3,x_1),(x_3,x_2),(x_3,x_5),(x_3,x_6),(x_4,x_1),(x_4,x_2),(x_4,x_5),\nonumber\\
&& ~(x_5,x_1),(x_5,x_2),(x_5,x_3),(x_5,x_4),(x_6,x_1),(x_6,x_2),(x_6,x_3)\}.
\end{eqnarray}
The trisection based on Yao's auxiliary function in Table~\ref{tab:auxiliary_Jxy_PY} is constructed as:
\begin{eqnarray}
R_J^{=Y}&=&\{(x,y)\in A\times A\mid \Phi_J^Y(x,y)\geq +\frac{1}{2}\},\nonumber\\
&=& \{(x_1,x_1),(x_1,x_2),(x_2,x_1),(x_2,x_2),(x_3,x_3),(x_3,x_4),(x_4,x_3),(x_4,x_4),\nonumber\\
&& ~(x_5,x_5),(x_5,x_6),(x_6,x_5),(x_6,x_6)\},\nonumber\\
R_J^{\asymp Y}&=&\{(x,y)\in A\times A\mid \Phi_J^Y(x,y)\leq -\frac{1}{2}\},\nonumber\\
&=& \{(x_4,x_6),(x_6,x_4)\},\nonumber\\
R_J^{\approx Y}&=&\{(x,y)\in A\times A\mid -\frac{1}{2}<\Phi_J^Y(x,y)<+\frac{1}{2}\},\nonumber\\
&=&  \{(x_1,x_3),(x_1,x_4),(x_1,x_5),(x_1,x_6),(x_2,x_3),(x_2,x_4),(x_2,x_5),(x_2,x_6),\nonumber\\
&& ~(x_3,x_1),(x_3,x_2),(x_3,x_5),(x_3,x_6),(x_4,x_1),(x_4,x_2),(x_4,x_5),\nonumber\\
&& ~(x_5,x_1),(x_5,x_2),(x_5,x_3),(x_5,x_4),(x_6,x_1),(x_6,x_2),(x_6,x_3)\}.
\end{eqnarray}

\end{example}

\section{Alliance and conflict functions}
\label{sec:functions}

We present the alliance and conflict functions that separate the two opposite semantics measured in one auxiliary function, including both unaggregated and aggregated formulations. Examples are given at the end to illustrate the presented concepts.

\subsection{Unaggregated alliance and conflict functions}

As discussed in the previous section, the existing trisections of agents, issues, and agent pairs in three-way conflict analysis are commonly based on a single evaluation function, either a rating or an auxiliary function. A common feature of these functions is that they measure two opposite attitudes or relations in one formula. The rating function gives both positive and negative attitudes, and the auxiliary function defines both alliance and conflict relations. This may induce certain difficulties in understanding, interpreting, and applying these functions in analyzing conflict problems. For example, when using the average function to aggregate the ratings of an agent on a set of issues, two opposite ratings $+1$ and $-1$ on two issues have the same effect as two neutral ratings $0$ and $0$. This issue is actually illustrated in Example~\ref{example:trisections_review}, where the aggregated ratings of agents $x_1$ and $x_5$ on $J=\{i_1,i_2,i_3,i_4,i_5\}$ are $r(x_1,J)=r(x_5,J)=+\frac{1}{5}$ (see Equation (\ref{equa:example_rating_aggregated_J})). As a result, $x_1$ and $x_5$ are determined to have the same attitude toward the set of issues $J$, even though $x_1$ clearly expresses a non-neutral attitude on issue $i_{1}$ only, and $x_5$ has non-neutral attitudes on all issues in $J$. As a mirror image, aggregating the ratings of a set of agents on a single issue also faces the same challenge.

There is a similar difficulty in aggregating an auxiliary function $\Phi_i$ into $\Phi_J$. A pair of alliance and conflict relations on two issues has the same effect as a pair of neutrality relations. Consider a subset of issues $J=\{i_2,i_3,i_4,i_5\}$ from Table~\ref{tab:situation_table_example1}. We first use Yao's auxiliary function for a single issue (in Table \ref{tab:auxiliary_ixy_PY}(b)) to compute the relations between agents $x_1$ and $x_3$ on every single issue in $J$, as given in the first line of Table~\ref{tab:auxiliary_Jxy_PY_computation}. Then we take their average as the aggregated value of $\Phi_J^Y(x_1,x_3)$ which is 0. We perform the same computation for agents $x_3$ and $x_5$ which is given in the second line of Table~\ref{tab:auxiliary_Jxy_PY_computation}. Clearly, we have $\Phi_J^Y(x_1,x_3)=\Phi_J^Y(x_3,x_5)=0$. Therefore, the agent $x_3$ is determined to have the same relation with $x_1$ and $x_5$ on $J$ (i.e., neutrality), even though the agent $x_3$ is in conflict with $x_5$ on two issues but not with $x_1$ for any issue. A conflict relation on an issue is dissolved by an alliance relation on another. The same problem exists in the auxiliary function $\Phi_J^P$. For example, we have $\Phi_J^P(x_1,x_2)=\Phi_J^P(x_3,x_5)=0$ as given in Table~\ref{tab:auxiliary_Jxy_PY_computation}.

\begin{table}[ht!]
\centering
\begin{tabular}{c|cccc|c}
\hline
\diagbox[width=5.5em,height=2em]{$\Phi_i$}{$J$}&$i_2$&$i_3$&$i_4$&$i_5$ & $\Phi_J$\\
\hline
$\Phi_i^Y(x_1,x_3)$&0&0&0&0 & $\Phi_J^Y(x_1,x_3) = 0$\\
$\Phi_i^Y(x_3,x_5)$&$+1$&$+1$&$-1$&$-1$& $\Phi_J^Y(x_3,x_5)=0$\\
\hline
$\Phi_i^P(x_1,x_2)$&0&0&0&0 & $\Phi_J^P(x_1,x_2)=0$ \\
$\Phi_i^P(x_3,x_5)$&$+1$&$+1$&$-1$&$-1$& $\Phi_J^P(x_3,x_5)=0$\\
\hline
\end{tabular}
\caption{Computing some values of $\Phi_J^Y$ and $\Phi_J^P$ in Table~\ref{tab:situation_table_example1}}
\label{tab:auxiliary_Jxy_PY_computation}
\end{table}
 
The reason for the above issues is that three-way conflict models with a single evaluation function do not distinguish between two opposite impacts, that is, either positive/negative attitudes or alliance/conflict relations. Intuitively in decision-making, people usually weigh the positive/good aspect and the negative/bad aspect separately and then combine them to make a final decision. A similar idea has been used in the three-way classification model based on two evaluation functions of positive and negative~\cite{Yao_2012}, as well as in the concept of bipolarity~\cite{Dubois_2008,Dubois_2012}. Following the same idea, we propose to split the auxiliary function, which describes both alliance and conflict, into two separate functions. An alliance function evaluates the alliance degree, and a conflict function evaluates the conflict degree. To clearly show their connections to the existing works, we define the alliance and conflict functions in terms of a given auxiliary function. Since auxiliary functions are defined through ratings, one may equivalently define these two functions in terms of ratings.

\begin{definition}
\label{def:alliance_conflict_function_ixy}
Given an auxiliary function $\Phi_i$ regarding a single issue $i\in I$, the alliance function $\Phi_i^=:~A\times A\rightarrow \{0,1\}$ and the conflict function $\Phi_i^\asymp:~A\times A\rightarrow \{0,1\}$ with respect to $i$ are defined as:
\begin{eqnarray}
\Phi_i^=(x,y)&=&
\left \{
\begin{array}{lcl}
1, & &\Phi_i(x,y)=+1,\\
& &\\
0,& &\Phi_i(x,y)\neq +1;\\
\end{array}
\right.\nonumber\\
~\nonumber\\
\Phi_i^{\asymp}(x,y)&=&
\left \{
\begin{array}{lcl}
1,& &\Phi_i(x,y)=-1,\\
& &\\
0, & &\Phi_i(x,y)\neq -1.
\end{array}
\right.
\end{eqnarray}
The function $\Phi_i$ can be any specific auxiliary function that satisfies the general formulation in Equation~\eqref{equa:auxiliary_ixy_general}.
\end{definition}

Intuitively, we use the values $1$ and $0$ to represent an answer of ``yes'' and ``no'' to the alliance and conflict relations. Accordingly, the alliance function describes the alliance and non-alliance relations between agents. If $\Phi_i^=(x,y)=1$, then $x$ and $y$ are allied; otherwise, $x$ and $y$ are not allied, which does not necessarily mean that they are in conflict. Similarly, the conflict function gives the conflict and non-conflict relations between agents. If $\Phi_i^{\asymp}(x,y)=1$, then $x$ and $y$ are in conflict; otherwise, $x$ and $y$ are not in conflict, which does not necessarily mean that they are allied. 

As mentioned above, one may also define the alliance and conflict functions through ratings rather than a given auxiliary function. In that case, an auxiliary function and the pair of alliance and conflict functions can be interchangeably defined. Specifically, we may define an auxiliary function through a pair of alliance and conflict functions as:
\begin{equation}
\Phi_i(x,y)=\Phi_i^=(x,y)-\Phi_i^{\asymp}(x,y).
\end{equation}
Equivalently, we have:
\begin{eqnarray}
\Phi_i(x,y)&=&
\left \{
\begin{array}{lcl}
+1, & &\Phi_i^=(x,y)=1 \wedge \Phi_i^\asymp(x,y)=0,\\
-1,& &\Phi_i^=(x,y)=0 \wedge \Phi_i^\asymp(x,y)=1,\\
0,& &{\rm otherwise}.\\
\end{array}
\right.
\end{eqnarray}

In our following discussion, we may alternatively denote the alliance and conflict functions with ratings as their parameters instead of the agents, particularly when we emphasize the ratings rather than the agents who give the ratings. Formally, for two agents $x,y\in A$, we have:
\begin{eqnarray}
\Phi_i^=(x,y)&=&\Phi^=(r(x,i),r(y,i)),\nonumber\\
\Phi_i^{\asymp}(x,y)&=&\Phi^{\asymp}(r(x,i),r(y,i)).
\end{eqnarray}

\subsection{Aggregated alliance and conflict functions}

The alliance $\Phi_i^=$ and conflict $\Phi_i^{\asymp}$ functions can be aggregated with respect to both issues and agents. Specifically, we may consider the following four cases of aggregation:
\begin{enumerate}[label=(C\arabic*)]
\item a set of issues only;
\item a set of agents as the first parameter only;
\item a set of agents as the second parameter only;
\item any combination of the above three.
\end{enumerate}
We propose a general definition of the aggregated alliance and conflict functions that could cover all these four cases.

\begin{definition}
\label{def:alliance_conflict_function_JXY}
A general definition of the aggregated alliance and conflict functions is given as follows:
for $J\subseteq I$ and $X,Y\subseteq A$,
\begin{eqnarray}
\label{equa:alliance_conflict_function_JXY}
&&\Phi_J^{=}(X,Y)=
\left \{
\begin{array}{lcl}
\frac{\sum\limits_{i\in J,x\in X,y\in Y}\Phi_i^{=}(x,y)}{|J|\cdot|X|\cdot|Y|}, & &J \ne \emptyset, X \ne \emptyset, {\rm and~} Y \neq \emptyset,\\
& &\\
0,& &{\rm otherwise};\\
\end{array}
\right.\nonumber\\
\nonumber\\
&&\Phi_J^{\asymp}(X,Y)=\left \{
\begin{array}{lcl}\frac{\sum\limits_{i\in J,x\in X,y\in Y}\Phi_i^{\asymp}(x,y)}{|J|\cdot|X|\cdot|Y|}, & &J \ne \emptyset, X \ne \emptyset, {\rm and~} Y \neq \emptyset,\\
& &\\
0,& &{\rm otherwise}.\\
\end{array}
\right.
\end{eqnarray}
\end{definition}

\begin{figure}[!ht]
\centering
\scalebox{0.8}{
\begin{tikzpicture}[auto, >=stealth', node distance=1em,
block_multilines1/.style ={rectangle, text width=4em, text centered, minimum height=2em},
block_multilines2/.style ={rectangle, text width=6em, text centered},
noborder_center/.style ={rectangle, draw=white, fill=white, text width=10em,text centered, minimum height=2em}]

\node [block_multilines1](JXY){$\Phi_J^{=}(X,Y)$};

\node [block_multilines1, below left=4em and 2em of JXY](JXy){$\Phi_J^{=}(X,y)$};
\node [block_multilines1, below=4em of JXY](JxY){$\Phi_J^{=}(x,Y)$};
\node [block_multilines1, below right=4em and 2em of JXY](iXY){$\Phi_i^=(X,Y)$};

\node [block_multilines1, below = 4em of JXy](Jxy){$\Phi_J^{=}(x,y)$};
\node [block_multilines1, below = 4em of JxY](iXy){$\Phi_i^=(X,y)$};
\node [block_multilines1, below = 4em of iXY](ixY){$\Phi_i^=(x,Y)$};

\node [block_multilines1, below = 4em of iXy](ixy){$\Phi_i^=(x,y)$};

\draw [-] (JXY.south) -- (JXy.north);
\draw [-] (JXY.south) -- (JxY.north);
\draw [-] (JXY.south) -- (iXY.north);
\draw [-] (JXy.south) -- (Jxy.north);
\draw [-] (JXy.south) -- (iXy.north);
\draw [-] (JxY.south) -- (Jxy.north);
\draw [-] (JxY.south) -- (ixY.north);
\draw [-] (iXY.south) -- (ixY.north);
\draw [-] (iXY.south) -- (iXy.north);
\draw [-] (Jxy.south) -- (ixy.north);
\draw [-] (iXy.south) -- (ixy.north);
\draw [-] (ixY.south) -- (ixy.north);

\node [block_multilines1, right=18em of JXY](-JXY){$\Phi_J^{\asymp}(X,Y)$};

\node [block_multilines1, below left=4em and 2em of -JXY](-JXy){$\Phi_J^{\asymp}(X,y)$};
\node [block_multilines1, below=4em of -JXY](-JxY){$\Phi_J^{\asymp}(x,Y)$};
\node [block_multilines1, below right=4em and 2em of -JXY](-iXY){$\Phi_i^{\asymp}(X,Y)$};

\node [block_multilines1, below = 4em of -JXy](-Jxy){$\Phi_J^{\asymp}(x,y)$};
\node [block_multilines1, below = 4em of -JxY](-iXy){$\Phi_i^{\asymp}(X,y)$};
\node [block_multilines1, below = 4em of -iXY](-ixY){$\Phi_i^{\asymp}(x,Y)$};

\node [block_multilines1, below = 4em of -iXy](-ixy){$\Phi_i^{\asymp}(x,y)$};

\draw [-] (-JXY.south) -- (-JXy.north);
\draw [-] (-JXY.south) -- (-JxY.north);
\draw [-] (-JXY.south) -- (-iXY.north);
\draw [-] (-JXy.south) -- (-Jxy.north);
\draw [-] (-JXy.south) -- (-iXy.north);
\draw [-] (-JxY.south) -- (-Jxy.north);
\draw [-] (-JxY.south) -- (-ixY.north);
\draw [-] (-iXY.south) -- (-ixY.north);
\draw [-] (-iXY.south) -- (-iXy.north);
\draw [-] (-Jxy.south) -- (-ixy.north);
\draw [-] (-iXy.south) -- (-ixy.north);
\draw [-] (-ixY.south) -- (-ixy.north);

\draw [<->,thick] (-5,-0.5)--(-5,-5.8);
\node [block_multilines1, left=9.5em of JXY](G){General};
\node [block_multilines2, left=8.5em of ixy](S){Special};
\end{tikzpicture}}
\caption{The lattices for alliance and conflict functions}
\label{fig:alliance_conflict_function_relation}
\end{figure}
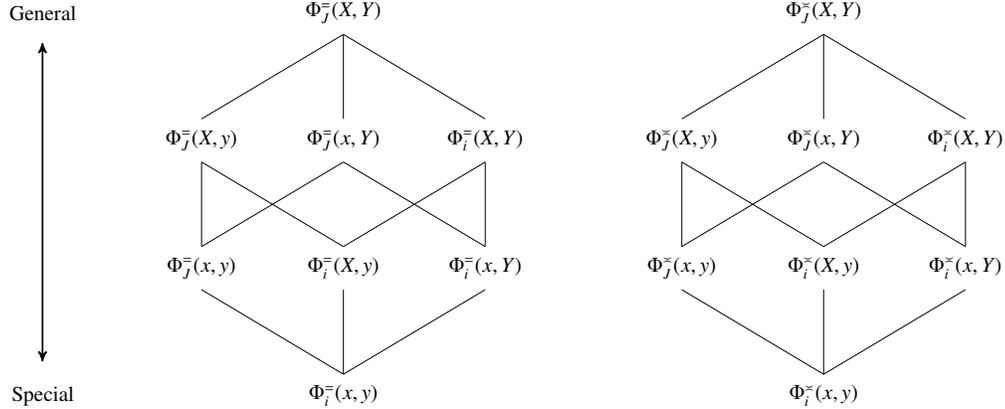

From the general definition, one may easily define other cases of aggregation mentioned above by degenerating one or more sets of $J$, $X$, and $Y$ into a singleton set. For example, for the first case of aggregation mentioned above (i.e., a set of issues only), we may simply degenerate the set $X$ into $\{x\}$ and $Y$ into $\{y\}$. Furthermore, for simplicity, we may use $x$ and $y$ instead of the singleton sets in the notations. Thus, we get the following aggregated functions:
\begin{eqnarray}
\label{equa:alliance_conflict_function_Jxy}
&&\Phi_J^{=}(x,y)=
\left \{
\begin{array}{lcl}
\frac{\sum\limits_{i\in J}\Phi_i^=(x,y)}{|J|}, & &J \neq \emptyset,\\
& &\\
0,& &J= \emptyset;\\
\end{array}
\right.\nonumber\\
\nonumber\\
&&\Phi_J^{\asymp}(x,y)=
\left \{
\begin{array}{lcl}
\frac{\sum\limits_{i\in J}\Phi_i^{\asymp}(x,y)}{|J|} & &J \neq \emptyset,\\
& &\\
0,& &J= \emptyset.\\
\end{array}
\right.
\end{eqnarray}
Similarly, we may also easily define the other cases of aggregated functions. Corresponding to all the cases of aggregation mentioned above, we arrive at the following 7 cases in total:
\begin{enumerate}[label=(C\arabic*)]
\item a set of issues only: $\Phi_J^{=}(x,y)$ and $\Phi_J^{\asymp}(x,y)$;
\item a set of agents as the first parameter only: $\Phi_i^=(X,y)$ and $\Phi_i^{\asymp}(X,y)$;
\item a set of agents as the second parameter only: $\Phi_i^=(x,Y)$ and $\Phi_i^{\asymp}(x,Y)$;
\item any combination of the above three:
\begin{itemize}
\item $\Phi_J^=(X,y)$ and $\Phi_J^{\asymp}(X,y)$;
\item $\Phi_J^=(x,Y)$ and $\Phi_J^{\asymp}(x,Y)$;
\item $\Phi_i^=(X,Y)$ and $\Phi_i^{\asymp}(X,Y)$;
\item $\Phi_J^=(X,Y)$ and $\Phi_J^{\asymp}(X,Y)$.
\end{itemize}
\end{enumerate}
We omit their definitions for simplicity. Following the two functions defined in Equation~(\ref{equa:alliance_conflict_function_JXY}), all these aggregated functions are defined in terms of the unaggregated functions $\Phi_i^=(x,y)$ and $\Phi_i^{\asymp}(x,y)$. Alternatively, one can also define an aggregated function in the case (C4) in terms of those in the cases from (C1) to (C3). For example, we can define $\Phi_i^=(X,Y)$ by aggregating $\Phi_i^=(x,Y)$ or $\Phi_i^=(X,y)$ as follows:
\begin{eqnarray}
&&\Phi_i^{=}(X,Y)=
\left \{
\begin{array}{lcl}
\frac{\sum\limits_{x\in X}\Phi_i^{=}(x,Y)}{|X|}, & &X \ne \emptyset,\\
& &\\
0,& &{\rm otherwise};\\
\end{array}
\right.\nonumber\\
\nonumber\\
&&\Phi_i^{=}(X,Y)=\left \{
\begin{array}{lcl}\frac{\sum\limits_{y\in Y}\Phi_i^{=}(X,y)}{|Y|}, & &Y \neq \emptyset,\\
& &\\
0,& &{\rm otherwise}.\\
\end{array}
\right.
\end{eqnarray}
Similarly, we can define $\Phi_J^=(X,Y)$ by aggregating $\Phi_i^=(X,Y)$ or $\Phi_J^=(x,Y)$ as follows:
\begin{eqnarray}
&&\Phi_J^{=}(X,Y)=
\left \{
\begin{array}{lcl}
\frac{\sum\limits_{i\in J}\Phi_i^{=}(X,Y)}{|J|}, & &J \ne \emptyset,\\
& &\\
0,& &{\rm otherwise};\\
\end{array}
\right.\nonumber\\
\nonumber\\
&&\Phi_J^{=}(X,Y)=\left \{
\begin{array}{lcl}\frac{\sum\limits_{x\in X}\Phi_J^{=}(x,Y)}{|X|}, & &X \neq \emptyset,\\
& &\\
0,& &{\rm otherwise}.\\
\end{array}
\right.
\end{eqnarray}
As a result, the relationships among alliance functions and those among conflict functions can be shown by the lattices in Figure~\ref{fig:alliance_conflict_function_relation}. A function can be defined in terms of any function below it as long as a path connects them. We get more general functions when going upwards and special cases when going downwards.

An essential issue in analyzing the agent relations is the choice of a subset of issues $J\subseteq I$. Two agents (or two groups of agents) may have different relations with respect to different subsets of issues. For a single agent $x\in A$, a subset of issues $J$ can be trisected into three disjoint parts $J^+_x$, $J^-_x$, and $J^0_x$ (as explained in Definition~\ref{def:trisection_issue_x_existing}). The agent has a neutral attitude towards issues in $J^0_x$, which can be interpreted as an unclear, uncertain, undetermined, or changeable attitude. 
In analyzing the relations between agents, it is reasonable to focus on the issues where the agents have clear, definite, and non-neutral attitudes. For an agent $x\in A$, we denote the set of issues on which $x$ has non-neutral attitudes as:
\begin{eqnarray}
J^{+-}_x&=&J^+_x\cup J^-_x,
\end{eqnarray}
which is called the set of non-neutral issues of $x$. One can define the alliance and conflict functions with respect to the non-neutral issues of $x$ by using $J^{+-}_x$ instead of $J$ in Equation~\eqref{equa:alliance_conflict_function_Jxy}:
\begin{eqnarray}
\label{equa:alliance_conflict_function_J+-xy}
\Phi_{J^{+-}_x}^=(x,y)&=&
\left \{
\begin{array}{lcl}
\frac{\sum\limits_{i\in J^{+-}_x}\Phi_i^=(x,y)}{|J^{+-}_x|}, & &J_x^{+-}\neq \emptyset,\\
& &\\
0,& &J_x^{+-}= \emptyset;\\
\end{array}
\right.\nonumber\\
~\nonumber\\
\Phi_{J^{+-}_x}^{\asymp}(x,y)&=&
\left \{
\begin{array}{lcl}
\frac{\sum\limits_{i\in J^{+-}_x}\Phi_i^{\asymp}(x,y)}{|J^{+-}_x|},& &J_x^{+-}\neq \emptyset,\\
& &\\
0, & &J_x^{+-}= \emptyset.
\end{array}
\right.
\end{eqnarray}

The function $\Phi_{J^{+-}_x}^=(x,y)$ represents the proportion of issues where $x$ and $y$ are allied out of the non-neutral issues in $J^{+-}_x$, which can be considered as the degree of $y$ supporting $x$ with respect to $J^{+-}_x$. Similarly, the function $\Phi_{J^{+-}_x}^{\asymp}(x,y)$ gives the proportion of issues where $x$ and $y$ are in conflict out of $J^{+-}_x$, which can be considered as the degree of $y$ opposing $x$ with respect to $J^{+-}_x$. The two functions in Equation (\ref{equa:alliance_conflict_function_J+-xy}) are not symmetric since different denominators are used in the computation, that is, $\Phi_{J^{+-}_x}^{=}(x,y) \ne \Phi_{J^{+-}_y}^{=}(y,x)$ and $\Phi_{J^{+-}_x}^{\asymp}(x,y) \ne \Phi_{J^{+-}_y}^{\asymp}(y,x)$. If the set of non-neutral issues $J^{+-}_x=\emptyset$, that is, the agent $x$ is neutral on all issues in $J$, then we have $\Phi_{J^{+-}_x}^{=}(x,y)=\Phi_{J^{+-}_x}^{\asymp}(x,y)=0$ for any other agent $y\in A$. In other words, an agent with all neutral ratings neither supports nor opposes other agents.

\subsection{Examples}

We give two examples to illustrate the concepts of alliance and conflict functions introduced in this section. Particularly, Example \ref{example:alliance_conflict_function_PY_definition} defines three pairs of alliance and conflict functions based on Pawlak's and Yao's auxiliary functions introduced in Example \ref{example:auxiliary_PY}, including both unaggregated and aggregated functions. Example \ref{example:alliance_conflict_function_PY_computation} further illustrates these three pairs of functions by showing the computation of their values with a specific situation table, including with respect to a set of issues $J$ and a set of non-neutral issues in $J$ regarding a specific agent.

\begin{example}[\textbf{Three pairs of alliance and conflict functions based on Pawlak's and Yao's auxiliary functions}]
\label{example:alliance_conflict_function_PY_definition}
 
We first illustrate the unaggregated alliance and conflict functions. By applying Pawlak's and Yao's auxiliary functions in Equations~\eqref{equa:auxiliary_ixy_Pawlak} and \eqref{equa:auxiliary_ixy_Yao} to Definition~\ref{def:alliance_conflict_function_ixy}, we get the alliance and conflict functions for individual agents with respect to a single issue $i\in I$ as follows:
\begin{eqnarray}
\label{equa:alliance_conflict_function_ixy_PY}
\Phi_i^{=P}(x,y)=\left\{ {\begin{array}{*{20}{lcl}}
1, & &\Phi_i^P(x,y)=+1,\\
& &\\
0,& &\Phi_i^P(x,y)\neq +1,\\
\end{array}} \right.\qquad\qquad
\Phi_i^{\asymp P}(x,y)=\left\{ {\begin{array}{*{20}{lcl}}
1, & &\Phi_i^P(x,y)=-1,\\
& &\\
0,& &\Phi_i^P(x,y)\neq -1,\\
\end{array}} \right.\nonumber\\\nonumber\\
\Phi_i^{=Y}(x,y)=\left\{ {\begin{array}{*{20}{lcl}}
1, & &\Phi_i^Y(x,y)=+1,\\
& &\\
0,& &\Phi_i^Y(x,y)\neq +1,\\
\end{array}} \right.\qquad\qquad
\Phi_i^{\asymp Y}(x,y)=\left\{ {\begin{array}{*{20}{lcl}}
1, & &\Phi_i^Y(x,y)=-1,\\
& &\\
0,& &\Phi_i^Y(x,y)\neq -1,\\
\end{array}} \right.
\end{eqnarray}
Tables~\ref{tab:alliance_conflict_function_ixy_P} and \ref{tab:alliance_conflict_function_ixy_Y} give the values of these four functions in terms of ratings as parameters instead of agents. The value $0^*$ means that the value is 1 instead of 0 if $x=y$. Apparently, Pawlak's and Yao's auxiliary functions induce the same conflict function, that is, we have $\Phi_i^{\asymp P}(x,y)=\Phi_i^{\asymp Y}(x,y)$. As discussed in Example \ref{example:auxiliary_PY}, Pawlak's and Yao's auxiliary functions express different opinions on the relations between two different agents with both neutral ratings. Thus, the two alliance functions only differ in the case of two different agents with both neutral ratings. While Pawlak considers them to be neutral to each other, Yao considers them to be allied.

\begin{table}[!ht]
\centering
\begin{subtable}{.45\linewidth}
\centering
\setlength{\tabcolsep}{1em}
\renewcommand{\arraystretch}{1}
\begin{tabular}{c|ccl}
\hline
\diagbox[width=6em,height=3em]{$r(x,i)$}{$r(y,i)$}& $+1$& $-1$    & $0$  \\ \hline
$+1$    &$1$&$0$&$0$\\
$-1$    &$0$ &$1$&$0$\\
$0$     &$0$&$0$&$0^*$\\
\hline
\end{tabular}
\caption{$\Phi^{= P}(r(x,i),r(y,i))$}
\end{subtable}
~
\begin{subtable}{.45\linewidth}
\centering
\setlength{\tabcolsep}{1em}
\renewcommand{\arraystretch}{1}
\begin{tabular}{c|ccc}
\hline
\diagbox[width=6em,height=3em]{$r(x,i)$}{$r(y,i)$}& $+1$& $-1$    & $0$  \\ \hline
$+1$&$0$&$1$&$0$\\
$-1$&$1$ &$0$&$0$\\
$0$&$0$&$0$&$0$\\
\hline
\end{tabular}
\caption{$\Phi^{\asymp P}(r(x,i),r(y,i))$}
\end{subtable}
\caption{The values of $\Phi^{= P}(r(x,i),r(y,i))$ and $\Phi^{\asymp P}(r(x,i),r(y,i))$}
\label{tab:alliance_conflict_function_ixy_P}
\end{table}

\begin{table}[!ht]
\centering
\begin{subtable}{.45\linewidth}
\centering
\setlength{\tabcolsep}{1em}
\renewcommand{\arraystretch}{1}
\begin{tabular}{c|ccc}
\hline
\diagbox[width=6em,height=3em]{$r(x,i)$}{$r(y,i)$}& $+1$& $-1$    & $0$  \\ \hline
$+1$    &$1$&$0$&$0$\\
$-1$    &$0$ &$1$&$0$\\
$0$     &$0$&$0$&$1$\\
\hline
\end{tabular}
\caption{$\Phi^{= Y}(r(x,i),r(y,i))$}
\end{subtable}
~
\begin{subtable}{.45\linewidth}
\centering
\setlength{\tabcolsep}{1em}
\renewcommand{\arraystretch}{1}
\begin{tabular}{c|ccc}
\hline
\diagbox[width=6em,height=3em]{$r(x,i)$}{$r(y,i)$}& $+1$& $-1$    & $0$  \\ \hline
$+1$&$0$&$1$&$0$\\
$-1$&$1$ &$0$&$0$\\
$0$&$0$&$0$&$0$\\
\hline
\end{tabular}
\caption{$\Phi^{\asymp Y}(r(x,i),r(y,i))$}
\end{subtable}
\caption{The values of $\Phi^{= Y}(r(x,i),r(y,i))$ and $\Phi^{\asymp Y}(r(x,i),r(y,i))$}
\label{tab:alliance_conflict_function_ixy_Y}
\end{table}

We then illustrate the aggregated alliance and conflict functions. Based on the functions in Equation (\ref{equa:alliance_conflict_function_ixy_PY}), we can define the aggregated alliance and conflict functions induced from Pawlak's and Yao's auxiliary functions. As an illustration, we consider the case {\rm (C1)} of aggregation, that is, a set of issues. We have given the general definition of $\Phi_J^{=}(x,y)$ and $\Phi_J^{\asymp}(x,y)$ in Equation~\eqref{equa:alliance_conflict_function_Jxy}. By simply replacing the general alliance and conflict functions $\Phi_i^{=}(x,y)$ and $\Phi_i^{\asymp}(x,y)$ with those in Equation (\ref{equa:alliance_conflict_function_ixy_PY}), we arrive at the following aggregated alliance and conflict functions with respect to a subset of issues $J\subseteq I$:
\begin{eqnarray}
\label{equa:alliance_conflict_function_Jxy_PY}
\Phi_J^{=P}(x,y)=\left\{ {\begin{array}{*{20}{lcl}}
\frac{\sum\limits_{i\in J}\Phi_i^{=P}(x,y)}{|J|}, & &J \neq \emptyset,\\
& &\\
0,& &J= \emptyset,\\
\end{array}} \right.\qquad\qquad
\Phi_J^{\asymp P}(x,y)=\left\{ {\begin{array}{*{20}{lcl}}
\frac{\sum\limits_{i\in J}\Phi_i^{\asymp P}(x,y)}{|J|} & &J \neq \emptyset,\\
& &\\
0,& &J= \emptyset,\\
\end{array}} \right.\nonumber\\\nonumber\\
\Phi_J^{=Y}(x,y)=\left\{ {\begin{array}{*{20}{lcl}}
\frac{\sum\limits_{i\in J}\Phi_i^{=Y}(x,y)}{|J|}, & &J \neq \emptyset,\\
& &\\
0,& &J= \emptyset,\\
\end{array}} \right.\qquad\qquad
\Phi_J^{\asymp Y}(x,y)=\left\{ {\begin{array}{*{20}{lcl}}
\frac{\sum\limits_{i\in J}\Phi_i^{\asymp Y}(x,y)}{|J|} & &J \neq \emptyset,\\
& &\\
0,& &J= \emptyset.\\
\end{array}} \right.
\end{eqnarray}

At last, we illustrate the aggregated alliance and conflict functions with respect to non-neutral issues. By replacing the general unaggregated alliance and conflict functions used in Equation (\ref{equa:alliance_conflict_function_J+-xy}) with those in Equation (\ref{equa:alliance_conflict_function_ixy_PY}), we can easily get the following aggregated functions with respect to the non-neutral issues of the first agent:
\begin{eqnarray}
\label{equa:alliance_conflict_function_J+-xy_PY}
\Phi_{J^{+-}_x}^{=P}(x,y)=\left\{ {\begin{array}{*{20}{lcl}}
\frac{\sum\limits_{i\in J^{+-}_x}\Phi_i^{=P}(x,y)}{|J^{+-}_x|}, & &J^{+-}_x \neq \emptyset,\\
& &\\
0,& &J^{+-}_x= \emptyset,\\
\end{array}} \right.\qquad\qquad
\Phi_{J^{+-}_x}^{\asymp P}(x,y)=\left\{ {\begin{array}{*{20}{lcl}}
\frac{\sum\limits_{i\in J^{+-}_x}\Phi_i^{\asymp P}(x,y)}{|J^{+-}_x|} & &J^{+-}_x \neq \emptyset,\\
& &\\
0,& &J^{+-}_x= \emptyset,\\
\end{array}} \right.\nonumber\\\nonumber\\
\Phi_{J^{+-}_x}^{=Y}(x,y)=\left\{ {\begin{array}{*{20}{lcl}}
\frac{\sum\limits_{i\in J^{+-}_x}\Phi_i^{=Y}(x,y)}{|J^{+-}_x|}, & &J^{+-}_x \neq \emptyset,\\
& &\\
0,& &J^{+-}_x= \emptyset,\\
\end{array}} \right.\qquad\qquad
\Phi_{J^{+-}_x}^{\asymp Y}(x,y)=\left\{ {\begin{array}{*{20}{lcl}}
\frac{\sum\limits_{i\in J^{+-}_x}\Phi_i^{\asymp Y}(x,y)}{|J^{+-}_x|} & &J^{+-}_x \neq \emptyset,\\
& &\\
0,& &J^{+-}_x= \emptyset.\\
\end{array}} \right.
\end{eqnarray}
These two pairs of aggregated alliance and conflict functions are actually equivalent, that is, we have $\Phi_{J^{+-}_x}^{=P}(x,y) = \Phi_{J^{+-}_x}^{=Y}(x,y)$ and $\Phi_{J^{+-}_x}^{\asymp P}(x,y) = \Phi_{J^{+-}_x}^{\asymp Y}(x,y)$. Recall that the unaggregated alliance functions induced from Pawlak's and Yao's functions only differ in the case where two different agents both hold a neutral rating on a single issue, as shown in Tables~\ref{tab:alliance_conflict_function_ixy_P}(a) and \ref{tab:alliance_conflict_function_ixy_Y}(a). Such an issue will not appear in $J^{+-}_x$. The unaggregated conflict functions induced from Pawlak's and Yao's function are exactly the same as shown in Tables~\ref{tab:alliance_conflict_function_ixy_P}(b) and \ref{tab:alliance_conflict_function_ixy_Y}(b). Thus, we must have $\Phi_{J^{+-}_x}^{\asymp P}(x,y) = \Phi_{J^{+-}_x}^{\asymp Y}(x,y)$.

\end{example}

\begin{example}[\textbf{Computing the three pairs of alliance and conflict functions in Example \ref{example:alliance_conflict_function_PY_definition}}]
\label{example:alliance_conflict_function_PY_computation}

We further illustrate the three pairs of alliance and conflict functions constructed in Example \ref{example:alliance_conflict_function_PY_definition} with the situation table given in Table~\ref{tab:situation_table_example2}.

\begin{table}[ht!]
\centering
\scalebox{0.83}{
\begin{tabular}{c|cccc}
\hline
\diagbox[width=3em,height=2em]{$A$}{$I$}&$i_1$&$i_2$&$i_3$&$i_4$\\
\hline
$x_1$&0&0&0&0\\
$x_2$&$+1$ &$+1$ &$-1$ &$-1$ \\
$x_3$&$+1$ &$+1$ &$-1$ &$-1$ \\
$x_4$&$+1$ &0&0&$-1$ \\
$x_5$&$+1$ &$+1$ &$+1$&$+1$\\
$x_6$&$+1$ &$+1$ &$+1$&$+1$\\
$x_7$&$-1$&$-1$&$-1$ &$-1$\\
$x_8$&$-1$&$-1$&$-1$ &$-1$\\
$x_{9}$ &$-1$&$-1$&$-1$ &$+1$\\
$x_{10}$ &$-1$&$-1$&0&0\\
$x_{11}$ &$-1$&$-1$&0&0\\
$x_{12}$ &$+1$ &$-1$&0&$-1$\\
\hline
\end{tabular}}
\caption{A three-valued situation table}
\label{tab:situation_table_example2}
\end{table}

For simplicity, let us consider a set of issues $J=I=\{i_1,i_2,i_3,i_4\}$. Based on the functions defined in Equation \eqref{equa:alliance_conflict_function_Jxy_PY}, one can easily compute the alliance and conflict degrees between agents with respect to $J$, which are shown in Tables~\ref{tab:example_alliance_conflict_function_Jxy_P_values} and \ref{tab:example_alliance_conflict_function_Jxy_Y_values}. We explicitly label the agents as the first or second parameter in the functions, which is important in those with respect to the non-neutral issues. As discussed in Example \ref{example:alliance_conflict_function_PY_definition}, with respect to the non-neutral issues, the alliance and conflict functions are equivalent in Pawlak's and Yao's cases. Thus, we give the values in both cases in Table \ref{tab:example_alliance_conflict_function_J+-xy_PY_values} where we omit the superscripts of $P$ and $Y$. We take the two agents $x_4$ and $x_{10}$ to illustrate the computation. The alliance and conflict degrees from the three pairs of functions are computed as: 
\begin{eqnarray}
\Phi_{J}^{=P}(x_4,x_{10})&=&\frac{\sum\limits_{i\in J}\Phi_i^{= P}(x_4,x_{10})}{|J|}=\frac{0+0+0+0}{4}=0,\nonumber\\
\Phi_{J}^{\asymp P}(x_4,x_{10})&=&\frac{\sum\limits_{i\in J}\Phi_i^{\asymp P}(x_4,x_{10})}{|J|}=\frac{1+0+0+0}{4}=\frac{1}{4};\nonumber\\
\Phi_{J}^{=Y}(x_4,x_{10})&=&\frac{\sum\limits_{i\in J}\Phi_i^{= Y}(x_4,x_{10})}{|J|}=\frac{0+0+1+0}{4}=\frac{1}{4},\nonumber\\
\Phi_{J}^{\asymp Y}(x_4,x_{10})&=&\frac{\sum\limits_{i\in J}\Phi_i^{\asymp Y}(x_4,x_{10})}{|J|}=\frac{1+0+0+0}{4}=\frac{1}{4};\nonumber\\
\Phi_{J^{+-}_{x_4}}^=(x_4,x_{10})&=&\frac{\sum\limits_{i\in J^{+-}_{x_4}}\Phi_i^{= P}(x_4,x_{10})}{|J^{+-}_{x_4}|}=\frac{\sum\limits_{i\in J^{+-}_{x_4}}\Phi_i^{= Y}(x_4,x_{10})}{|J^{+-}_{x_4}|}=\frac{\sum\limits_{i\in \{i_1,i_4\}}\Phi_i^{= Y}(x_4,x_{10})}{|\{i_1,i_4\}|}=\frac{0+0}{2}=0,\nonumber\\
\Phi_{J^{+-}_{x_4}}^{\asymp}(x_4,x_{10})&=&\frac{\sum\limits_{i\in J^{+-}_{x_4}}\Phi_i^{\asymp P}(x_4,x_{10})}{|J^{+-}_{x_4}|}=\frac{\sum\limits_{i\in J^{+-}_{x_4}}\Phi_i^{\asymp Y}(x_4,x_{10})}{|J^{+-}_{x_4}|}=\frac{\sum\limits_{i\in \{i_1,i_4\}}\Phi_i^{\asymp Y}(x_4,x_{10})}{|\{i_1,i_4\}|}=\frac{1+0}{2}=\frac{1}{2}.
\end{eqnarray}

\begin{table}[ht!]
\centering
\renewcommand\arraystretch{1.2}
\setlength{\tabcolsep}{1.5mm}
\scalebox{0.7}{
\begin{tabular}{c|cc|cc|cc|cc|cc|cc|cc|cc|cc|cc|cc|cc}
\hline
  \multirow{2}{*}{{\diagbox[width=4em,height=3.4em]{2nd}{1st}}}&\multicolumn{2}{c|}{$x_1$}&\multicolumn{2}{c|}{$x_2$}&\multicolumn{2}{c|}{$x_3$}&\multicolumn{2}{c|}{$x_4$}&
  \multicolumn{2}{c|}{$x_5$}&\multicolumn{2}{c|}{$x_6$}&\multicolumn{2}{c|}{$x_7$}&\multicolumn{2}{c|}{$x_8$}&\multicolumn{2}{c|}{$x_{9}$}
  &\multicolumn{2}{c|}{$x_{10}$}&\multicolumn{2}{c|}{$x_{11}$}&\multicolumn{2}{c}{$x_{12}$}\\\cline{2-25}
&$\Phi_{J}^{=P}$&$\Phi_{J}^{\asymp P}$&$\Phi_{J}^{=P}$&$\Phi_{J}^{\asymp P}$&$\Phi_{J}^{=P}$&$\Phi_{J}^{\asymp P}$&$\Phi_{J}^{=P}$&$\Phi_{J}^{\asymp P}$ &$\Phi_{J}^{=P}$&$\Phi_{J}^{\asymp P}$&$\Phi_{J}^{=P}$&$\Phi_{J}^{\asymp P}$&$\Phi_{J}^{=P}$&$\Phi_{J}^{\asymp P}$ &$\Phi_{J}^{=P}$&$\Phi_{J}^{\asymp P}$&$\Phi_{J}^{=P}$&$\Phi_{J}^{\asymp P}$&$\Phi_{J}^{=P}$&$\Phi_{J}^{\asymp P}$&$\Phi_{J}^{=P}$&$\Phi_{J}^{\asymp P}$&$\Phi_{J}^{=P}$&$\Phi_{J}^{\asymp P}$\\
\hline
$x_1$   &1 &$0$ &$0$ &$0$   &$0$  &$0$  &$0$  &$0$  &$0$  &$0$ &$0$  &$0$  &$0$  &$0$  &$0$  &$0$  &$0$ &$0$  &$0$  &$0$  &$0$  &$0$   &$0$  &$0$ \\
$x_2$   &0 &$0$ &1   &$0$   &1    &$0$  &$\frac{1}{2}$  &$0$  &$\frac{1}{2}$ &$\frac{1}{2}$&$\frac{1}{2}$ &$\frac{1}{2}$&$\frac{1}{2}$ &$\frac{1}{2}$&$\frac{1}{2}$ &$\frac{1}{2}$&$\frac{1}{4}$ &$\frac{3}{4}$&$0$ &$\frac{1}{2}$ &$0$  &$\frac{1}{2}$ &$\frac{1}{2}$&$\frac{1}{4}$\\
$x_3$   &0 &$0$ &1   &$0$   &1    &$0$  &$\frac{1}{2}$  &$0$  &$\frac{1}{2}$ &$\frac{1}{2}$&$\frac{1}{2}$ &$\frac{1}{2}$&$\frac{1}{2}$ &$\frac{1}{2}$&$\frac{1}{2}$ &$\frac{1}{2}$&$\frac{1}{4}$ &$\frac{3}{4}$&$0$ &$\frac{1}{2}$ &$0$  &$\frac{1}{2}$ &$\frac{1}{2}$&$\frac{1}{4}$\\
$x_4$   &0 &$0$ &$\frac{1}{2}$ &$0$   &$\frac{1}{2}$  &$0$  &$1$  &$0$  &$\frac{1}{4}$ &$\frac{1}{4}$&$\frac{1}{4}$ &$\frac{1}{4}$&$\frac{1}{4}$ &$\frac{1}{4}$&$\frac{1}{4}$ &$\frac{1}{4}$&0   &$\frac{1}{2}$&0   &$\frac{1}{4}$&0   &$\frac{1}{4}$&$\frac{1}{2}$&$0$ \\
$x_5$   &0 &$0$ &$\frac{1}{2}$ &$\frac{1}{2}$ &$\frac{1}{2}$ &$\frac{1}{2}$ &$\frac{1}{4}$ &$\frac{1}{4}$ &1   &0    &1   &0    &0   &$1$  &0   &$1$  &$\frac{1}{4}$ &$\frac{3}{4}$&0   &$\frac{1}{2}$  &0   &$\frac{1}{2}$  &$\frac{1}{4}$&$\frac{1}{2}$\\
$x_6$   &0 &$0$ &$\frac{1}{2}$ &$\frac{1}{2}$ &$\frac{1}{2}$ &$\frac{1}{2}$ &$\frac{1}{4}$ &$\frac{1}{4}$ &1   &0    &1   &0    &0   &$1$  &0   &$1$  &$\frac{1}{4}$ &$\frac{3}{4}$&0   &$\frac{1}{2}$  &0   &$\frac{1}{2}$  &$\frac{1}{4}$&$\frac{1}{2}$\\
$x_7$   &0 &$0$ &$\frac{1}{2}$ &$\frac{1}{2}$ &$\frac{1}{2}$ &$\frac{1}{2}$ &$\frac{1}{4}$ &$\frac{1}{4}$ &0   &$1$  &0   &$1$  &1   &0     &1   &0   &$\frac{3}{4}$ &$\frac{1}{4}$&$\frac{1}{2}$   &0     &$\frac{1}{2}$   &0     &$\frac{1}{2}$&$\frac{1}{4}$\\
$x_8$   &0 &$0$ &$\frac{1}{2}$ &$\frac{1}{2}$ &$\frac{1}{2}$ &$\frac{1}{2}$ &$\frac{1}{4}$ &$\frac{1}{4}$ &0   &$1$  &0   &$1$  &1   &0     &1   &0   &$\frac{3}{4}$ &$\frac{1}{4}$&$\frac{1}{2}$   &0     &$\frac{1}{2}$   &0     &$\frac{1}{2}$&$\frac{1}{4}$\\
$x_9$   &0 &$0$ &$\frac{1}{4}$ &$\frac{3}{4}$ &$\frac{1}{4}$ &$\frac{3}{4}$ &0   &$\frac{1}{2}$   &$\frac{1}{4}$ &$\frac{3}{4}$&$\frac{1}{4}$ &$\frac{3}{4}$&$\frac{3}{4}$ &$\frac{1}{4}$&$\frac{3}{4}$ &$\frac{1}{4}$&1  &0     &$\frac{1}{2}$   &0     &$\frac{1}{2}$   &0     &$\frac{1}{4}$&$\frac{1}{2}$\\
$x_{10}$&0 &$0$ &0   &$\frac{1}{2}$ &0   &$\frac{1}{2}$ &0   &$\frac{1}{4}$ &0   &$\frac{1}{2}$&0   &$\frac{1}{2}$&$\frac{1}{2}$ &0     &$\frac{1}{2}$ &0     &$\frac{1}{2}$ &0     &1   &0     &$\frac{1}{2}$   &0     &$\frac{1}{4}$&$\frac{1}{4}$\\
$x_{11}$&0 &$0$ &0   &$\frac{1}{2}$ &0   &$\frac{1}{2}$ &0   &$\frac{1}{4}$ &0   &$\frac{1}{2}$&0   &$\frac{1}{2}$&$\frac{1}{2}$ &0     &$\frac{1}{2}$ &0     &$\frac{1}{2}$ &0     &$\frac{1}{2}$   &0     &1   &0     &$\frac{1}{4}$&$\frac{1}{4}$\\
$x_{12}$&0 &$0$ &$\frac{1}{2}$ &$\frac{1}{4}$ &$\frac{1}{2}$ &$\frac{1}{4}$ &$\frac{1}{2}$   &0     &$\frac{1}{4}$ &$\frac{1}{2}$&$\frac{1}{4}$ &$\frac{1}{2}$&$\frac{1}{2}$ &$\frac{1}{4}$&$\frac{1}{2}$ &$\frac{1}{4}$&$\frac{1}{4}$ &$\frac{1}{2}$&$\frac{1}{4}$ &$\frac{1}{4}$&$\frac{1}{4}$ &$\frac{1}{4}$&$1$  &$0$\\
\hline
\end{tabular}}
\caption{The values of $\Phi_{J}^{= P}$ and $\Phi_{J}^{\asymp P}$ with respect to Table \ref{tab:situation_table_example2}}
\label{tab:example_alliance_conflict_function_Jxy_P_values}
\end{table}

\begin{table}[ht!]
\centering
\renewcommand\arraystretch{1.2}
\setlength{\tabcolsep}{1.5mm}
\scalebox{0.7}{
\begin{tabular}{c|cc|cc|cc|cc|cc|cc|cc|cc|cc|cc|cc|cc}
\hline
  \multirow{2}{*}{{\diagbox[width=4em,height=3.4em]{2nd}{1st}}}&\multicolumn{2}{c|}{$x_1$}&\multicolumn{2}{c|}{$x_2$}&\multicolumn{2}{c|}{$x_3$}&\multicolumn{2}{c|}{$x_4$}&
  \multicolumn{2}{c|}{$x_5$}&\multicolumn{2}{c|}{$x_6$}&\multicolumn{2}{c|}{$x_7$}&\multicolumn{2}{c|}{$x_8$}&\multicolumn{2}{c|}{$x_{9}$}
  &\multicolumn{2}{c|}{$x_{10}$}&\multicolumn{2}{c|}{$x_{11}$}&\multicolumn{2}{c}{$x_{12}$}\\\cline{2-25}
&$\Phi_{J}^{=Y}$&$\Phi_{J}^{\asymp Y}$&$\Phi_{J}^{=Y}$&$\Phi_{J}^{\asymp Y}$&$\Phi_{J}^{=Y}$&$\Phi_{J}^{\asymp Y}$&$\Phi_{J}^{=Y}$&$\Phi_{J}^{\asymp Y}$ &$\Phi_{J}^{=Y}$&$\Phi_{J}^{\asymp Y}$&$\Phi_{J}^{=Y}$&$\Phi_{J}^{\asymp Y}$&$\Phi_{J}^{=Y}$&$\Phi_{J}^{\asymp Y}$ &$\Phi_{J}^{=Y}$&$\Phi_{J}^{\asymp Y}$&$\Phi_{J}^{=Y}$&$\Phi_{J}^{\asymp Y}$&$\Phi_{J}^{=Y}$&$\Phi_{J}^{\asymp Y}$&$\Phi_{J}^{=Y}$&$\Phi_{J}^{\asymp Y}$&$\Phi_{J}^{=Y}$&$\Phi_{J}^{\asymp Y}$\\
\hline
$x_1$   &1 &$0$ &$0$ &$0$   &$0$  &$0$  &$\frac{1}{2}$  &$0$  &$0$  &$0$ &$0$  &$0$  &$0$  &$0$  &$0$  &$0$  &$0$ &$0$  &$\frac{1}{2}$  &$0$  &$\frac{1}{2}$  &$0$   &$\frac{1}{4}$  &$0$ \\
$x_2$   &0 &$0$ &1   &$0$   &1    &$0$  &$\frac{1}{2}$  &$0$  &$\frac{1}{2}$ &$\frac{1}{2}$&$\frac{1}{2}$ &$\frac{1}{2}$&$\frac{1}{2}$ &$\frac{1}{2}$&$\frac{1}{2}$ &$\frac{1}{2}$&$\frac{1}{4}$ &$\frac{3}{4}$&$0$ &$\frac{1}{2}$ &$0$  &$\frac{1}{2}$ &$\frac{1}{2}$&$\frac{1}{4}$\\
$x_3$   &0 &$0$ &1   &$0$   &1    &$0$  &$\frac{1}{2}$  &$0$  &$\frac{1}{2}$ &$\frac{1}{2}$&$\frac{1}{2}$ &$\frac{1}{2}$&$\frac{1}{2}$ &$\frac{1}{2}$&$\frac{1}{2}$ &$\frac{1}{2}$&$\frac{1}{4}$ &$\frac{3}{4}$&$0$ &$\frac{1}{2}$ &$0$  &$\frac{1}{2}$ &$\frac{1}{2}$&$\frac{1}{4}$\\
$x_4$   &$\frac{1}{2}$ &$0$ &$\frac{1}{2}$ &$0$   &$\frac{1}{2}$  &$0$  &$1$  &$0$  &$\frac{1}{4}$ &$\frac{1}{4}$&$\frac{1}{4}$ &$\frac{1}{4}$&$\frac{1}{4}$ &$\frac{1}{4}$&$\frac{1}{4}$ &$\frac{1}{4}$&0   &$\frac{1}{2}$&$\frac{1}{4}$   &$\frac{1}{4}$&$\frac{1}{4}$   &$\frac{1}{4}$&$\frac{3}{4}$&$0$ \\
$x_5$   &0 &$0$ &$\frac{1}{2}$ &$\frac{1}{2}$ &$\frac{1}{2}$ &$\frac{1}{2}$ &$\frac{1}{4}$ &$\frac{1}{4}$ &1   &0    &1   &0    &0   &$1$  &0   &$1$  &$\frac{1}{4}$ &$\frac{3}{4}$&0   &$\frac{1}{2}$  &0   &$\frac{1}{2}$  &$\frac{1}{4}$&$\frac{1}{2}$\\
$x_6$   &0 &$0$ &$\frac{1}{2}$ &$\frac{1}{2}$ &$\frac{1}{2}$ &$\frac{1}{2}$ &$\frac{1}{4}$ &$\frac{1}{4}$ &1   &0    &1   &0    &0   &$1$  &0   &$1$  &$\frac{1}{4}$ &$\frac{3}{4}$&0   &$\frac{1}{2}$  &0   &$\frac{1}{2}$  &$\frac{1}{4}$&$\frac{1}{2}$\\
$x_7$   &0 &$0$ &$\frac{1}{2}$ &$\frac{1}{2}$ &$\frac{1}{2}$ &$\frac{1}{2}$ &$\frac{1}{4}$ &$\frac{1}{4}$ &0   &$1$  &0   &$1$  &1   &0     &1   &0   &$\frac{3}{4}$ &$\frac{1}{4}$&$\frac{1}{2}$   &0     &$\frac{1}{2}$   &0     &$\frac{1}{2}$&$\frac{1}{4}$\\
$x_8$   &0 &$0$ &$\frac{1}{2}$ &$\frac{1}{2}$ &$\frac{1}{2}$ &$\frac{1}{2}$ &$\frac{1}{4}$ &$\frac{1}{4}$ &0   &$1$  &0   &$1$  &1   &0     &1   &0   &$\frac{3}{4}$ &$\frac{1}{4}$&$\frac{1}{2}$   &0     &$\frac{1}{2}$   &0     &$\frac{1}{2}$&$\frac{1}{4}$\\
$x_9$   &0 &$0$ &$\frac{1}{4}$ &$\frac{3}{4}$ &$\frac{1}{4}$ &$\frac{3}{4}$ &0   &$\frac{1}{2}$   &$\frac{1}{4}$ &$\frac{3}{4}$&$\frac{1}{4}$ &$\frac{3}{4}$&$\frac{3}{4}$ &$\frac{1}{4}$&$\frac{3}{4}$ &$\frac{1}{4}$&1  &0     &$\frac{1}{2}$   &0     &$\frac{1}{2}$   &0     &$\frac{1}{4}$&$\frac{1}{2}$\\
$x_{10}$&$\frac{1}{2}$ &$0$ &0   &$\frac{1}{2}$ &0   &$\frac{1}{2}$ &$\frac{1}{4}$   &$\frac{1}{4}$ &0   &$\frac{1}{2}$&0   &$\frac{1}{2}$&$\frac{1}{2}$ &0     &$\frac{1}{2}$ &0     &$\frac{1}{2}$ &0     &1   &0     &1   &0     &$\frac{1}{2}$&$\frac{1}{4}$\\
$x_{11}$&$\frac{1}{2}$ &$0$ &0   &$\frac{1}{2}$ &0   &$\frac{1}{2}$ &$\frac{1}{4}$   &$\frac{1}{4}$ &0   &$\frac{1}{2}$&0   &$\frac{1}{2}$&$\frac{1}{2}$ &0     &$\frac{1}{2}$ &0     &$\frac{1}{2}$ &0     &1   &0     &1   &0     &$\frac{1}{2}$&$\frac{1}{4}$\\
$x_{12}$&$\frac{1}{4}$ &$0$ &$\frac{1}{2}$ &$\frac{1}{4}$ &$\frac{1}{2}$ &$\frac{1}{4}$ &$\frac{3}{4}$   &0     &$\frac{1}{4}$ &$\frac{1}{2}$&$\frac{1}{4}$ &$\frac{1}{2}$&$\frac{1}{2}$ &$\frac{1}{4}$&$\frac{1}{2}$ &$\frac{1}{4}$&$\frac{1}{4}$ &$\frac{1}{2}$&$\frac{1}{2}$ &$\frac{1}{4}$&$\frac{1}{2}$ &$\frac{1}{4}$&$1$  &$0$\\
\hline
\end{tabular}}
\caption{The values of $\Phi_{J}^{= Y}$ and $\Phi_{J}^{\asymp Y}$ with respect to Table \ref{tab:situation_table_example2}}
\label{tab:example_alliance_conflict_function_Jxy_Y_values}
\end{table}

\begin{table}[ht!]
\centering
\renewcommand\arraystretch{1.2}
\setlength{\tabcolsep}{1.5mm}
\scalebox{0.65}{
\begin{tabular}{c|cc|cc|cc|cc|cc|cc|cc|cc|cc|cc|cc|cc}
\hline
  \multirow{2}{*}{{\diagbox[width=4em,height=3.4em]{2nd}{1st}}}&\multicolumn{2}{c|}{$x_1$}&\multicolumn{2}{c|}{$x_2$}&\multicolumn{2}{c|}{$x_3$}&\multicolumn{2}{c|}{$x_4$}&
  \multicolumn{2}{c|}{$x_5$}&\multicolumn{2}{c|}{$x_6$}&\multicolumn{2}{c|}{$x_7$}&\multicolumn{2}{c|}{$x_8$}&\multicolumn{2}{c|}{$x_{9}$}
  &\multicolumn{2}{c|}{$x_{10}$}&\multicolumn{2}{c|}{$x_{11}$}&\multicolumn{2}{c}{$x_{12}$}\\\cline{2-25}
&$\Phi_{J^{+-}_{x_1}}^=$&$\Phi_{J^{+-}_{x_1}}^{\asymp}$&$\Phi_{J^{+-}_{x_2}}^=$&$\Phi_{J^{+-}_{x_2}}^{\asymp}$&$\Phi_{J^{+-}_{x_3}}^=$&$\Phi_{J^{+-}_{x_3}}^{\asymp}$&$\Phi_{J^{+-}_{x_4}}^=$&$\Phi_{J^{+-}_{x_4}}^{\asymp}$ &$\Phi_{J^{+-}_{x_5}}^=$&$\Phi_{J^{+-}_{x_5}}^{\asymp}$&$\Phi_{J^{+-}_{x_6}}^=$&$\Phi_{J^{+-}_{x_6}}^{\asymp}$&$\Phi_{J^{+-}_{x_7}}^=$&$\Phi_{J^{+-}_{x_7}}^{\asymp}$ &$\Phi_{J^{+-}_{x_8}}^=$&$\Phi_{J^{+-}_{x_8}}^{\asymp}$&$\Phi_{J^{+-}_{x_9}}^=$&$\Phi_{J^{+-}_{x_9}}^{\asymp}$&$\Phi_{J^{+-}_{x_{10}}}^=$&$\Phi_{J^{+-}_{x_{10}}}^{\asymp}$&$\Phi_{J^{+-}_{x_{11}}}^=$&$\Phi_{J^{+-}_{x_{11}}}^{\asymp}$&$\Phi_{J^{+-}_{x_{12}}}^=$&$\Phi_{J^{+-}_{x_{12}}}^{\asymp}$\\
\hline
$x_1$   &1 &$0$ &$0$ &$0$   &$0$  &$0$  &$0$  &$0$  &$0$  &$0$ &$0$  &$0$  &$0$  &$0$  &$0$  &$0$  &$0$ &$0$  &$0$  &$0$  &$0$  &$0$   &$0$  &$0$ \\
$x_2$   &0 &$0$ &1   &$0$   &1    &$0$  &$1$  &$0$  &$\frac{1}{2}$ &$\frac{1}{2}$&$\frac{1}{2}$ &$\frac{1}{2}$&$\frac{1}{2}$ &$\frac{1}{2}$&$\frac{1}{2}$ &$\frac{1}{2}$&$\frac{1}{4}$ &$\frac{3}{4}$&$0$ &$1$ &$0$  &$1$ &$\frac{2}{3}$&$\frac{1}{3}$\\
$x_3$   &0 &$0$ &1   &$0$   &1    &$0$  &$1$  &$0$  &$\frac{1}{2}$ &$\frac{1}{2}$&$\frac{1}{2}$ &$\frac{1}{2}$&$\frac{1}{2}$ &$\frac{1}{2}$&$\frac{1}{2}$ &$\frac{1}{2}$&$\frac{1}{4}$ &$\frac{3}{4}$&$0$ &$1$ &$0$  &$1$ &$\frac{2}{3}$&$\frac{1}{3}$\\
$x_4$   &0 &$0$ &$\frac{1}{2}$ &$0$   &$\frac{1}{2}$  &$0$  &$1$  &$0$  &$\frac{1}{4}$ &$\frac{1}{4}$&$\frac{1}{4}$ &$\frac{1}{4}$&$\frac{1}{4}$ &$\frac{1}{4}$&$\frac{1}{4}$ &$\frac{1}{4}$&0   &$\frac{1}{2}$&0   &$\frac{1}{2}$&0   &$\frac{1}{2}$&$\frac{2}{3}$&$0$ \\
$x_5$   &0 &$0$ &$\frac{1}{2}$ &$\frac{1}{2}$ &$\frac{1}{2}$ &$\frac{1}{2}$ &$\frac{1}{2}$ &$\frac{1}{2}$ &1   &0    &1   &0    &0   &$1$  &0   &$1$  &$\frac{1}{4}$ &$\frac{3}{4}$&0   &$1$  &0   &$1$  &$\frac{1}{3}$&$\frac{2}{3}$\\
$x_6$   &0 &$0$ &$\frac{1}{2}$ &$\frac{1}{2}$ &$\frac{1}{2}$ &$\frac{1}{2}$ &$\frac{1}{2}$ &$\frac{1}{2}$ &1   &0    &1   &0    &0   &$1$  &0   &$1$  &$\frac{1}{4}$ &$\frac{3}{4}$&0   &$1$  &0   &$1$  &$\frac{1}{3}$&$\frac{2}{3}$\\
$x_7$   &0 &$0$ &$\frac{1}{2}$ &$\frac{1}{2}$ &$\frac{1}{2}$ &$\frac{1}{2}$ &$\frac{1}{2}$ &$\frac{1}{2}$ &0   &$1$  &0   &$1$  &1   &0     &1   &0   &$\frac{3}{4}$ &$\frac{1}{4}$&1   &0     &1   &0     &$\frac{2}{3}$&$\frac{1}{3}$\\
$x_8$   &0 &$0$ &$\frac{1}{2}$ &$\frac{1}{2}$ &$\frac{1}{2}$ &$\frac{1}{2}$ &$\frac{1}{2}$ &$\frac{1}{2}$ &0   &$1$  &0   &$1$  &1   &0     &1   &0   &$\frac{3}{4}$ &$\frac{1}{4}$&1   &0     &1   &0     &$\frac{2}{3}$&$\frac{1}{3}$\\
$x_9$   &0 &$0$ &$\frac{1}{4}$ &$\frac{3}{4}$ &$\frac{1}{4}$ &$\frac{3}{4}$ &0   &$1$   &$\frac{1}{4}$ &$\frac{3}{4}$&$\frac{1}{4}$ &$\frac{3}{4}$&$\frac{3}{4}$ &$\frac{1}{4}$&$\frac{3}{4}$ &$\frac{1}{4}$&1  &0     &1   &0     &1   &0     &$\frac{1}{3}$&$\frac{2}{3}$\\
$x_{10}$&0 &$0$ &0   &$\frac{1}{2}$ &0   &$\frac{1}{2}$ &0   &$\frac{1}{2}$ &0   &$\frac{1}{2}$&0   &$\frac{1}{2}$&$\frac{1}{2}$ &0     &$\frac{1}{2}$ &0     &$\frac{1}{2}$ &0     &1   &0     &1   &0     &$\frac{1}{3}$&$\frac{1}{3}$\\
$x_{11}$&0 &$0$ &0   &$\frac{1}{2}$ &0   &$\frac{1}{2}$ &0   &$\frac{1}{2}$ &0   &$\frac{1}{2}$&0   &$\frac{1}{2}$&$\frac{1}{2}$ &0     &$\frac{1}{2}$ &0     &$\frac{1}{2}$ &0     &1   &0     &1   &0     &$\frac{1}{3}$&$\frac{1}{3}$\\
$x_{12}$&0 &$0$ &$\frac{1}{2}$ &$\frac{1}{4}$ &$\frac{1}{2}$ &$\frac{1}{4}$ &1   &0     &$\frac{1}{4}$ &$\frac{1}{2}$&$\frac{1}{4}$ &$\frac{1}{2}$&$\frac{1}{2}$ &$\frac{1}{4}$&$\frac{1}{2}$ &$\frac{1}{4}$&$\frac{1}{4}$ &$\frac{1}{2}$&$\frac{1}{2}$ &$\frac{1}{2}$&$\frac{1}{2}$ &$\frac{1}{2}$&$1$  &$0$\\
\hline
\end{tabular}}
\caption{The values of $\Phi_{J^{+-}}^{=}$ and $\Phi_{J^{+-}}^{\asymp}$ with respect to Table \ref{tab:situation_table_example2}}
\label{tab:example_alliance_conflict_function_J+-xy_PY_values}
\end{table}

\end{example}

\section{Alliance sets and their decisions}
\label{sec:alliance_set_and_decision}

\begin{figure}[!ht]
\centering
\scalebox{0.85}{
\begin{tikzpicture}[auto, >=stealth', node distance=1em,
block_multilines1/.style ={rectangle, draw=black, fill=white, text width=12.5em, text centered, minimum height=3em,rounded corners=4pt,thick},
block_multilines2/.style ={rectangle, draw=black, fill=white, text width=28em, text centered, minimum height=3em,rounded corners=4pt,thick},
block_multilines3/.style ={rectangle, draw=white, fill=white,text centered, minimum height=1.8em,rounded corners=4pt,thick},
block_multilines4/.style ={rectangle, draw=black, fill=white,text width=28em,text centered, minimum height=10em,rounded corners=4pt,thick},
block_multilines5/.style ={rectangle, dotted, draw=black, fill=white, text width=26em, text centered, minimum height=3.8em,rounded corners=4pt,thick},
block_multilines6/.style ={rectangle, draw=black, fill=white, text width=3.8em, minimum height=3em,text centered,rounded corners=4pt,thick},
block_multilines7/.style ={rectangle, draw=black, fill=white, text width=26em, text centered, minimum height=3.8em,rounded corners=4pt,thick},
block_multilines8/.style ={rectangle, text width=10em, minimum height=3em,text centered,rounded corners=4pt,thick},
noborder_center/.style ={rectangle,text centered, minimum height=2em}
]
\node [block_multilines1] (TT) {A three-valued situation table};
\node [block_multilines2,below=2em of TT] (A) {Trisecting agent pairs\\(Definition~\ref{def:trisection_agentpair_J_our})\\~\\~\\~\\~};
\node [block_multilines3, below =4.8em of TT] (AA) {$A\times A$};
\node [block_multilines3, below left=-0.5em and 5.4em of AA] (DJ) {(Equation~\eqref{equa:trisection_agentpair_J_our})};
\node [block_multilines3, below right=-0.5em and 5.4em of AA] (DJ+-) {(Equation~\eqref{equa:trisection_agentpair_J+-_our})};
\node [block_multilines3, below =2em of AA] (NR) {Neutrality};
\node [block_multilines3, below left=2em and 1em of AA] (AR) {Alliance};
\node [block_multilines3, below right=2em and 1em of AA] (CR) {Conflict};
\node [block_multilines4, below =3em of A] (n1) {};
\node [block_multilines5, below =3.5em of A] (ASJ) {Alliance set of an agent \\(Definition~\ref{def:alliance_set})};
\node [block_multilines5, below =1em of ASJ] (MAJ) {Maximal consistent alliance set\\(Definition~\ref{def:maximal_consistent_alliance_set})};
\node [block_multilines6, below left=3.9em and -6.5em of A] (AJx) {$AS_J(x)$};
\node [block_multilines6, below right=3.9em and -6.5em of A] (AJ+-x) {$AS_{J^{+-}}(x)$};
\node [block_multilines6, below =1.8em of AJx] (MJx) {$\mathcal{M\!A}_J$};
\node [block_multilines6, below =1.8em of AJ+-x] (MJ+-x) {$\mathcal{M\!A}_{J^{+-}}$};
\node [block_multilines1,below =2em of n1] (A3) {Trisecting issues\\(Definition~\ref{def:trisection_issue_X_our})\\~\\~\\~\\~};
\node [block_multilines3, below =4.8em of n1] (I) {$J$};
\node [block_multilines3, below =2em of I] (IN) {$J_X^0$};
\node [block_multilines3, below left=2em and 2em of I] (IP) {$J_X^+$};
\node [block_multilines3, below right=2em and 2em of I] (INE) {$J_X^-$};
\node [block_multilines7, below =2em of A3] (A4) {Decision of an alliance set $X$\\(Definition~\ref{def:decision_maximal_consistent_alliance_set})};
\node [block_multilines6, left=-5.5em of A4] (DJx) {${\rm Des}_J(X)$};
\node [block_multilines6, right=-5.5em of A4] (DJ+-x) {${\rm Des}_{J^{+-}}(X)$};
\node [block_multilines8, above=8.5em of AJx](n10){\color{orange}{\large\textcircled{\small 1}} with respect to \\$J\subseteq I$};
\node [block_multilines8, above=8.5em of AJ+-x](n11){\color{blue}{\large\textcircled{\small 2}} with respect to \\$J^{+-}_x\subseteq J$};
\draw [->,thick] (TT.south) -- (A.north);
\draw [->,thick] (A.south) -- (n1.north);
\draw [->,thick] (n1.south) -- (A3.north);
\draw [->,thick] (A3.south) -- (A4.north);
\draw [->,thick] (AA.south) -- (AR.north);
\draw [->,thick] (AA.south) -- (NR.north);
\draw [->,thick] (AA.south) -- (CR.north);
\draw [->,thick] (I.south) -- (IP.north);
\draw [->,thick] (I.south) -- (IN.north);
\draw [->,thick] (I.south) -- (INE.north);
\draw [->,thick] (AJx.south) -- (MJx.north);
\draw [->,thick] (AJ+-x.south) -- (MJ+-x.north);
\draw [dashed,thick,orange](-13.8em,-7.8em) rectangle (-7em,-42.35em);
\draw [dashed,thick,blue](7em,-7.8em) rectangle (13.8em,-42.35em);
\end{tikzpicture}}
\caption{A framework of the discussion in Section \ref{sec:alliance_set_and_decision}}
\label{fig:framework_alliance_set_and_decision}
\end{figure}
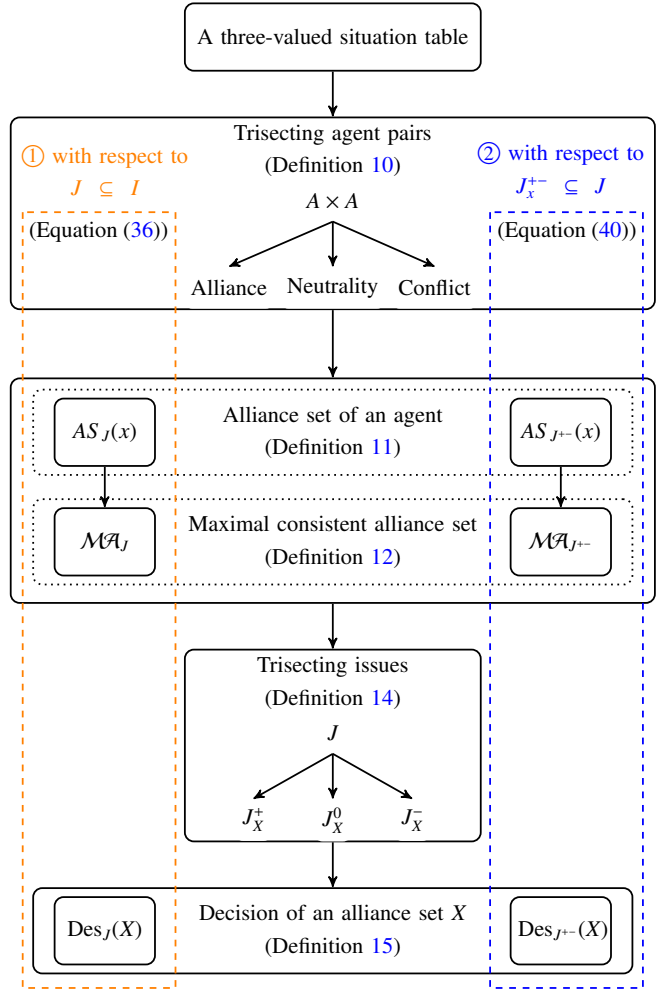

Based on the alliance and conflict functions, we present the concept of alliance sets and investigate the description of their decisions. As shown in Figure~\ref{fig:framework_alliance_set_and_decision}, our discussion includes two streams, one with respect to a set of issues $J\subseteq I$ and another with respect to the non-neutral issues $J_x^{+-}\subseteq J$ regarding an agent $x$. Intuitively, an alliance set for a specific agent is the group of agents that are in the alliance relation with it. In a special case that the agents in an alliance set are allied with each other, we can derive a concept of the maximal consistent alliance set. The agents in an alliance set are in close co-operation. Thus, it is useful to analyze their ratings of issues as a whole, which is called the decision of the alliance set. Examples are given at the end to illustrate the presented concepts.

\subsection{Alliance sets: Trisecting agent pairs with alliance and conflict functions}
\label{sec:alliance_set}

We study the relationships between agents based on the proposed alliance and conflict functions. By combining the two opposite viewpoints represented in the two functions, we define the alliance, conflict, and neutrality relations as follows.

\begin{definition}
\label{def:trisection_agentpair_J_our}
Suppose $\Phi_J^=$ and $\Phi_J^{\asymp}$ are the alliance and conflict functions with respect to a subset of issues $J\subseteq I$.
Given two pairs of thresholds $(l_a,h_a)$ and $(l_c,h_c)$ with $0\leq l_a< h_a\leq 1$ and $0\leq l_c< h_c\leq 1$,
the alliance relation $R_J^{=}(\Phi^{=},\Phi^{\asymp})$, conflict relation $R_J^{\asymp}(\Phi^{=},\Phi^{\asymp})$, and neutrality relation $R_J^{\approx}(\Phi^{=},\Phi^{\asymp})$ are defined as:
\begin{eqnarray}
\label{equa:trisection_agentpair_J_our}
R_J^{=}(\Phi^{=},\Phi^{\asymp})&=&\{(x,y)\in A\times A\mid \Phi_J^{=}(x,y)\geq h_a\wedge \Phi_J^{\asymp}(x,y)\leq l_c\},\nonumber\\
R_J^{\asymp}(\Phi^{=},\Phi^{\asymp})&=&\{(x,y)\in A\times A\mid \Phi_J^{=}(x,y)\leq l_a\wedge \Phi_J^{\asymp}(x,y)\geq h_c\},\nonumber\\
R_J^{\approx}(\Phi^{=},\Phi^{\asymp})&=&(R_J^{=}(\Phi^{=},\Phi^{\asymp})\cup R_J^{\asymp}(\Phi^{=},\Phi^{\asymp}))^c.
\end{eqnarray}
\end{definition}

The above trisection of agent pairs can also be explained through the elevation and reduction operations in shadowed sets that are discussed at the end of Section \ref{sec:review_auxiliary}. If the alliance degree is high enough and at the same time, the conflict degree is low enough, that is, $\Phi_J^{=}(x,y)\geq h_a\wedge \Phi_J^{\asymp}(x,y)\leq l_c$, we elevate the alliance degree to 1 and reduce the conflict degree to 0. Accordingly, the two agents $x$ and $y$ are determined to be allied. One can similarly explain the conflict and neutrality relations. A difference from the existing definitions of these relations is that we apply both elevation and reduction operations in constructing one relation. 

Definition~\ref{def:trisection_agentpair_J_our} actually gives a trisection of agent pairs in $A \times A$ with respect to $J$. One can easily define the trisection with respect to a single issue $i \in I$ by degenerating $J$ to $\{i\}$. Such a trisection is actually equivalent to the trisection induced from an auxiliary function (i.e., Definition~\ref{def:trisection_agentpair_i_existing}).

\begin{theorem}
\label{theorem:trisection_agentpair_i_equivalent}
With respect to an issue $i\in I$, the trisections of agent pairs with an auxiliary function and with {a pair of} alliance and conflict functions are equivalent, that is:
\begin{eqnarray}
\label{equa:trisection_agentpair_i_equivalent}
R_i^{=}(\Phi^{=},\Phi^{\asymp})&=&R_i^{=},\nonumber\\
R_i^{\asymp}(\Phi^{=},\Phi^{\asymp})&=&R_i^{\asymp},\nonumber\\
R_i^{\approx}(\Phi^{=},\Phi^{\asymp})&=&R_i^{\approx}.
\end{eqnarray}
\end{theorem}
\begin{proof}
By Definition~\ref{def:trisection_agentpair_J_our}, we get the trisection of agent pairs with respect to a single issue $i \in I$ as:
\begin{eqnarray}
R_i^{=}(\Phi^{=},\Phi^{\asymp})&=&\{(x,y)\in A\times A\mid \Phi_i^{=}(x,y)\geq h_a\wedge \Phi_i^{\asymp}(x,y)\leq l_c\},\nonumber\\
R_i^{\asymp}(\Phi^{=},\Phi^{\asymp})&=&\{(x,y)\in A\times A\mid \Phi_i^{=}(x,y)\leq l_a\wedge \Phi_i^{\asymp}(x,y)\geq h_c\},\nonumber\\
R_i^{\approx}(\Phi^{=},\Phi^{\asymp})&=&(R_i^{=}(\Phi^{=},\Phi^{\asymp})\cup R_i^{\asymp}(\Phi^{=},\Phi^{\asymp}))^c,
\end{eqnarray}
where we use $i$ to denote the singleton set $\{i\}$ for simplicity. Since $0\leq l_a< h_a\leq 1$ and $0\leq l_c< h_c\leq 1$, we have $l_a<1$, $h_a>0$, $l_c<1$, and $h_c>0$. According to Definition~\ref{def:alliance_conflict_function_ixy}, the values of alliance and conflict functions $\Phi_i^=$ and $\Phi_i^{\asymp}$ can only take $1$ or 0. Therefore, we have:
\begin{eqnarray}
R_i^{=}(\Phi^{=},\Phi^{\asymp})&=&\{(x,y)\in A\times A\mid \Phi_i^{=}(x,y) > 0 \wedge  \Phi_i^{\asymp}(x,y) < 1\}\nonumber\\
&=&\{(x,y)\in A\times A\mid \Phi_i^{=}(x,y) =1 \wedge  \Phi_i^{\asymp}(x,y) =0\}\nonumber\\
&=&\{(x,y)\in A\times A\mid \Phi_i(x,y)=+1\}\nonumber\\
&=&R_i^{=},\nonumber\\
R_i^{\asymp}(\Phi^{=},\Phi^{\asymp})&=& \{(x,y)\in A\times A\mid \Phi_i^{=}(x,y) <1 \wedge  \Phi_i^{\asymp}(x,y) >0\}\nonumber\\
&=&\{(x,y)\in A\times A\mid \Phi_i^{=}(x,y) =0 \wedge  \Phi_i^{\asymp}(x,y) =1\}\nonumber\\
&=&\{(x,y)\in A\times A\mid \Phi_i(x,y)=-1\}\nonumber\\
&=&R_i^{\asymp},\nonumber\\
R_i^{\approx}(\Phi^{=},\Phi^{\asymp})&=&(R_i^{=}(\Phi^{=},\Phi^{\asymp}) \cup R_i^{\asymp}(\Phi^{=},\Phi^{\asymp}))^c\nonumber\\
&=& (R_i^{=} \cup R_i^{\asymp})^c=R_i^{\approx}.
\end{eqnarray}
\end{proof}

The alliance and conflict functions enable us to measure the two opposite views of alliance and conflict separately when trisecting agent pairs. These two views are combined to define the three relations of alliance, conflict, and neutrality. Such a separation and combination of the two views may provide an in-depth understanding of the agent relations. As shown in Table~\ref{tab:agent_relation_two_view}, from the alliance view, if the alliance degree of two agents $x$ and $y$ is high enough, i.e., $\Phi_J^=(x,y)\geq h_a$, then they are strongly allied; if the alliance degree is low enough, i.e., $\Phi_J^=(x,y)\leq l_a$, then they are considered not to be allied; otherwise, they are weakly allied. Similarly, from the conflict view, if the conflict degree of $x$ and $y$ is high enough, i.e., $\Phi_J^{\asymp}(x,y)\geq h_c$, then they are in strong conflict; if the conflict degree is low enough, i.e., $\Phi_J^{\asymp}(x,y)\leq l_c$, then they are not in conflict; otherwise, they are in weak conflict. The combinations of two views give the final definition of the relationship between the two agents. A pair of two agents belongs to the alliance relation if they are strongly allied and not in conflict, and a pair of agents belongs to the conflict relation if they are in strong conflict and not allied. One may also use less restrictive conditions in defining the relations if reasonable in specific situations.

\begin{table}[ht!]
\centering
\setlength{\tabcolsep}{0.6em}
\renewcommand{\arraystretch}{1}
\begin{tabular}{c|c|c|c}
\hline
\multirow{2}{*}{\diagbox[width=7.28em,height=2.9em]{$\Phi_J^{\asymp}$}{$\Phi_J^{=}$}}& Non-alliance & Weak-alliance  & Strong-alliance  \\
&$[0,l_a]$ & $(l_a,h_a)$  &  $[h_a,1]$  \\ \hline
 Non-conflict& \multirow{2}{*}{Neutrality} & \multirow{2}{*}{Neutrality} &\multirow{2}{*}{Alliance}\\
$[0,l_c]$&&&\\
Weak-conflict &\multirow{2}{*}{Neutrality} & \multirow{2}{*}{Neutrality} &\multirow{2}{*}{Neutrality} \\
$(l_c,h_c)$&&&\\
Strong-conflict &\multirow{2}{*}{Conflict} &\multirow{2}{*}{Neutrality} &\multirow{2}{*}{Neutrality}\\
$[h_c,1]$&&&\\ \hline
\end{tabular}
\caption{Interpreting agent relations by combining two views of alliance and conflict}
\label{tab:agent_relation_two_view}
\end{table}

Based on Definition \ref{def:trisection_agentpair_J_our}, we can easily define the alliance, conflict, and neutrality relations with respect to the non-neutral issues as:
\begin{eqnarray}
\label{equa:trisection_agentpair_J+-_our}
R_{J^{+-}}^{=}(\Phi^{=},\Phi^{\asymp})&=&\{(x,y)\in A\times A\mid \Phi_{J^{+-}_{x}}^=(x,y)\geq h_a\wedge \Phi_{J^{+-}_{x}}^{\asymp}(x,y)\leq l_c\},\nonumber\\
R_{J^{+-}}^{\asymp}(\Phi^{=},\Phi^{\asymp})&=&\{(x,y)\in A\times A\mid \Phi_{J^{+-}_{x}}^=(x,y)\leq l_a\wedge \Phi_{J^{+-}_{x}}^{\asymp}(x,y)\geq h_c\},\nonumber\\
R_{J^{+-}}^{\approx}(\Phi^{=},\Phi^{\asymp})&=&(R_{J^{+-}}^{=}(\Phi^{=},\Phi^{\asymp})\cup R_{J^{+-}}^{\asymp}(\Phi^{=},\Phi^{\asymp}))^c.
\end{eqnarray}

By grouping the agents allied with a specific agent $x$, one can get the alliance set of $x$.

\begin{definition}
\label{def:alliance_set}
The alliance sets of an agent $x\in A$ with respect to a subset of issues $J\subseteq I$ and the non-neutral issues $J_x^{+-}\subseteq J$ are respectively defined as:
\begin{eqnarray}
AS_J(x)&=&\{y\in A\mid (x,y)\in R_J^{=}(\Phi^{=},\Phi^{\asymp})\},\nonumber\\
AS_{J^{+-}}(x)&=&\{y\in A\mid (x,y)\in R_{J^{+-}}^{=}(\Phi^{=},\Phi^{\asymp})\}.
\end{eqnarray}
\end{definition}

By Equation~\eqref{equa:trisection_agentpair_J_our}, the alliance relation $R_{J}^{=}(\Phi^{=},\Phi^{\asymp})$ is reflexive and symmetric, but not transitive. In other words, if $(x,y)\in R_{J}^{=}(\Phi^{=},\Phi^{\asymp})$ and $(y,z)\in R_{J}^{=}(\Phi^{=},\Phi^{\asymp})$, we may not have $(x,z)\in R_{J}^{=}(\Phi^{=},\Phi^{\asymp})$. 
By Equation~\eqref{equa:trisection_agentpair_J+-_our}, the alliance relation $R_{J^{+-}}^{=}(\Phi^{=},\Phi^{\asymp})$ is reflexive, but not symmetric or transitive. 
In other words, if $(x,y)\in R_{J^{+-}}^{=}(\Phi^{=},\Phi^{\asymp})$ and $(y,z)\in R_{J^{+-}}^{=}(\Phi^{=},\Phi^{\asymp})$, we may not have $(y,x), (z,y),$ or $(x,z)\in R_{J^{+-}}^{=}(\Phi^{=},\Phi^{\asymp})$. As a result, although the agents in an alliance set $AS_J(x)$ are all allied with $x$, they are not necessarily allied with each other, which may induce some difficulties in analyzing the alliance sets and agent relations. To solve this problem, we present the definition of maximal consistent alliance sets by adopting the ideas in the concept of maximal consistent blocks proposed by Leung~\cite{Leung_2003}.

\begin{definition}
\label{def:maximal_consistent_alliance_set}
A subset of agents $M\subseteq A$ is a maximal consistent alliance set with respect to a subset of issues $J\subseteq I$ if it satisfies the following conditions:
\begin{eqnarray*}
&(1)&\forall\, x,y\in M, (x,y)\in R_{J}^{=}(\Phi^{=},\Phi^{\asymp}),\\
&(2)&\forall\, M'\supset M, \exists\, x,y\in M', (x,y)\notin R_{J}^{=}(\Phi^{=},\Phi^{\asymp}).
\end{eqnarray*}
The family of maximal consistent alliance sets with respect to $J$ is denoted as $\mathcal{MA}_J$.
\end{definition}

Intuitively, a maximal consistent alliance set is a maximal set where every two agents are allied with each other with respect to a set of issues. By taking $R_{J^{+-}}^{=}(\Phi^{=},\Phi^{\asymp})$ instead of $R_{J}^{=}(\Phi^{=},\Phi^{\asymp})$ in Definition~\ref{def:maximal_consistent_alliance_set}, one can  easily define the concept of a maximal consistent alliance set with respect to non-neutral issues. Accordingly, the family of maximal consistent alliance sets with respect to non-neutral issues is denoted as $\mathcal{MA}_{J^{+-}}$. 

\subsection{The decisions of alliance sets: Trisecting issues with alliance and conflict functions}
\label{sec:alliance_set}

Intuitively, the agents in an alliance set share approximately the same attitudes towards a set of issues. These shared attitudes represent the decision or description of these agents as a whole. An agent in an alliance set supports or at least does not oppose this decision. To represent such decisions, we first define the formal description of an agent $x\in A$ regarding a subset of issues $J\subseteq I$ in terms of the ratings, which is also called the decision of $x$ on $J$. Recall that a subset of issues $J$ can be trisected with respect to an agent $x$ into three parts, namely, positive issues $J_x^+$, negative issues $J_x^-$, and neutral issues $J_x^0$ (see Definition~\ref{def:trisection_issue_x_existing}). The description or decision of an agent on a set of issues is formally defined as follows.

\begin{definition}
For a subset of issues $J \subseteq I$, the description or decision of an agent $x \in A$ on $J$ is defined as:
\begin{eqnarray}
\label{equa:description_agent}
{\rm Des}_J(x)&=&\bigwedge_{i\in J}\langle i,r(x,i)\rangle\nonumber\\
&=&(\bigwedge_{i\in J_x^+}\langle i,+1\rangle)\wedge(\bigwedge_{i\in J_x^-}\langle i,-1\rangle)\wedge(\bigwedge_{i\in J_x^0}\langle i,0\rangle).
\end{eqnarray}
\end{definition}

In the alliance set $AS_J(x)$ of an agent $x\in A$, all the agents are allied with $x$ by sharing attitudes with $x$ towards the subset of issues $J$. In other words, the decision of the alliance set $AS_J(x)$ can be represented by the description or decision of $x$:
\begin{eqnarray}
{\rm Des}(AS_J(x))={\rm Des}_J(x).
\end{eqnarray}
Similarly, the decision of alliance set $AS_{J^{+-}}(x)$ is the description of an agent $x\in A$ with respect to the non-neutral issues $J_x^{+-}$:
\begin{eqnarray}
{\rm Des}(AS_{J^{+-}}(x))={\rm Des}_{J_x^{+-}}(x)=\bigwedge_{i\in J_x^{+-}}\langle i,r(x,i)\rangle=(\bigwedge_{i\in J_x^+}\langle i,+1\rangle)\wedge(\bigwedge_{i\in J_x^-}\langle i,-1\rangle).
\end{eqnarray}
It should be noted that an agent with a neutral rating on every issue under consideration does not have a description with respect to the non-neutral issues. Therefore, there is no valid decision of alliance set $AS_{J^{+-}}(x)$ if $J_x^{+-} = \emptyset$.

When it comes to a maximal consistent alliance set, there is no such ``primary'' agent that can represent the decision of the whole set. Nevertheless, we may follow the format in Equation~\eqref{equa:description_agent} by trisecting a set of issues $J \subseteq I$ into three subsets of positive, negative, and neutral issues. This requires an aggregation of the ratings from the agents in the maximal consistent alliance set on $J$. Under the framework of our study, we obtain the aggregation of ratings and the trisection of $J$ based on the proposed alliance and conflict functions. Instead of simply taking average to aggregate the ratings, we propose another approach by introducing two imaginary agents. A positive imaginary agent $x^+$ has a positive rating $+1$ on all issues in $I$ and a negative imaginary agent $x^-$ has a negative rating $-1$ on all issues in $I$. These two agents represent two extremes of positive and negative attitudes on $I$. Formally, the two agents $x^+$ and $x^-$ are described over $I$ as:
\begin{eqnarray}
{\rm Des}_I(x^+)&=&\bigwedge_{i\in I}\langle i,+1\rangle,\nonumber\\
{\rm Des}_I(x^-)&=&\bigwedge_{i\in I}\langle i,-1\rangle.
\end{eqnarray}
Then for a given set of agents $X$, we compare it with $x^+$ and $x^-$ from both views of alliance and conflict. More specifically, for $i\in I$, we compute the following aggregated alliance and conflict functions regarding $x^+/x^-$ and $X$ by Definition~\ref{def:alliance_conflict_function_JXY} as follows:
\begin{eqnarray}
\label{equa:alliance_conflict_function_ix+-X}
\Phi_i^{=}(x^+,X)&=&\frac{\sum\limits_{y\in X}\Phi_i^=(x^+,y)}{|X|}=\frac{\sum\limits_{y\in X}\Phi^=(+1,r(y,i))}{|X|},\nonumber\\
\Phi_i^{\asymp}(x^+,X)&=&\frac{\sum\limits_{y\in X}\Phi_i^{\asymp}(x^+,y)}{|X|}=\frac{\sum\limits_{y\in X}\Phi^{\asymp}(+1,r(y,i))}{|X|};\nonumber\\
\Phi_i^{=}(x^-,X)&=&\frac{\sum\limits_{y\in X}\Phi_i^=(x^-,y)}{|X|}=\frac{\sum\limits_{y\in X}\Phi^=(-1,r(y,i))}{|X|},\nonumber\\
\Phi_i^{\asymp}(x^-,X)&=&\frac{\sum\limits_{y\in X}\Phi_i^{\asymp}(x^-,y)}{|X|}=\frac{\sum\limits_{y\in X}\Phi^{\asymp}(-1,r(y,i))}{|X|}.
\end{eqnarray}
Intuitively, these four functions describe two views of the rating of $X$ as a whole regarding the issue $i$. The functions $\Phi_i^{=}(x^+,X)$ and $\Phi_i^{\asymp}(x^-,X)$ both indicate how positive $X$ is, but from two different views. The function $\Phi_i^{=}(x^+,X)$ shows how $X$ supports $x^+$ (i.e., a positive rating on $i$) and $\Phi_i^{\asymp}(x^-,X)$ shows how $X$ opposes $x^-$ (i.e., a negative rating on $i$). Similarly, the other two functions $\Phi_i^{\asymp}(x^+,X)$ and $\Phi_i^{=}(x^-,X)$ indicate how negative $X$ is. A combination of one function from each view gives a full understanding of the rating of $X$. There are obviously four combinations as given in Table~\ref{tab:combine_alliance_conflict_function_ix+-X}. One can trisect a subset of issues $J \subseteq I$ with respect to a given set of agents using any of the four combinations.

\begin{table}[ht!]
\centering
\setlength{\tabcolsep}{0.6em}
\renewcommand{\arraystretch}{1.2}
\begin{tabular}{c|cc}
\hline
\diagbox[width=13em,height=2.5em]{\small Positive view}{\footnotesize Negative view}& $\Phi_i^{\asymp}(x^+,X)$ & $\Phi_i^{=}(x^-,X)$\\ \hline
$\Phi_i^{=}(x^+,X)$& $\circled{1}$ &$\circled{2}$\\
$\Phi_i^{\asymp}(x^-,X)$ &$\circled{3}$&$\circled{4}$\\ \hline
\end{tabular}
\caption{Four combinations of the alliance and conflict functions}
\label{tab:combine_alliance_conflict_function_ix+-X}
\end{table}

\begin{definition}
\label{def:trisection_issue_X_our}
For a subset of agents $X\subseteq A$ and a subset of issues $J\subseteq I$. Given two pairs of thresholds $(l_p,h_p)$ and $(l_n,h_n)$ with $0\leq l_p< h_p\leq 1$ and $0\leq l_n< h_n\leq 1$, we define four trisections of $J$ with respect to $X$ as:

$(1)$ by using combination $\circled{1}$ in Table~\ref{tab:combine_alliance_conflict_function_ix+-X}:
\begin{eqnarray}
J_{X}^{+}(\Phi^{=},\Phi^{\asymp})&=&\{i\in J\mid \Phi_i^{=}(x^+,X)\geq h_p\wedge \Phi_i^{\asymp}(x^+,X)\leq l_n\},\nonumber\\
J_{X}^{-}(\Phi^{=},\Phi^{\asymp})&=&\{i\in J\mid \Phi_i^{=}(x^+,X)\leq l_p\wedge \Phi_i^{\asymp}(x^+,X)\geq h_n\},\nonumber\\
J_{X}^{0}(\Phi^{=},\Phi^{\asymp})&=&(J_{X}^{+}(\Phi^{=},\Phi^{\asymp})\cup J_{X}^{-}(\Phi^{=},\Phi^{\asymp}))^c.
\end{eqnarray}

$(2)$ by using combination $\circled{2}$ in Table~\ref{tab:combine_alliance_conflict_function_ix+-X}:
\begin{eqnarray}
J_{X}^{+}(\Phi^{=},\Phi^{=})&=&\{i\in J\mid \Phi_i^{=}(x^+,X)\geq h_p\wedge \Phi_i^{=}(x^-,X)\leq l_n\},\nonumber\\
J_{X}^{-}(\Phi^{=},\Phi^{=})&=&\{i\in J\mid \Phi_i^{=}(x^+,X)\leq l_p\wedge \Phi_i^{=}(x^-,X)\geq h_n\},\nonumber\\
J_{X}^{0}(\Phi^{=},\Phi^{=})&=&(J_{X}^{+}(\Phi^{=},\Phi^{=})\cup J_{X}^{-}(\Phi^{=},\Phi^{=}))^c.
\end{eqnarray}

$(3)$ by using combination $\circled{3}$ in Table~\ref{tab:combine_alliance_conflict_function_ix+-X}:
\begin{eqnarray}
J_{X}^{+}(\Phi^{\asymp},\Phi^{\asymp})&=&\{i\in J\mid \Phi_i^{\asymp}(x^-,X)\geq h_p\wedge \Phi_i^{\asymp}(x^+,X)\leq l_n\},\nonumber\\
J_{X}^{-}(\Phi^{\asymp},\Phi^{\asymp})&=&\{i\in J\mid  \Phi_i^{\asymp}(x^-,X)\leq l_p\wedge\Phi_i^{\asymp}(x^+,X)\geq h_n\},\nonumber\\
J_{X}^{0}(\Phi^{\asymp},\Phi^{\asymp})&=&(J_{X}^{+}(\Phi^{\asymp},\Phi^{\asymp})\cup J_{X}^{-}(\Phi^{\asymp},\Phi^{\asymp}))^c.
\end{eqnarray}

$(4)$ by using combination $\circled{4}$ in Table~\ref{tab:combine_alliance_conflict_function_ix+-X}:
\begin{eqnarray}
J_{X}^{+}(\Phi^{\asymp},\Phi^{=})&=&\{i\in J\mid \Phi_i^{\asymp}(x^-,X)\geq h_p\wedge \Phi_i^{=}(x^-,X)\leq l_n\},\nonumber\\
J_{X}^{-}(\Phi^{\asymp},\Phi^{=})&=&\{i\in J\mid \Phi_i^{\asymp}(x^-,X)\leq l_p\wedge \Phi_i^{=}(x^-,X)\geq h_n\},\nonumber\\
J_{X}^{0}(\Phi^{\asymp},\Phi^{=})&=&(J_{X}^{+}(\Phi^{\asymp},\Phi^{=})\cup J_{X}^{-}(\Phi^{\asymp},\Phi^{=}))^c.
\end{eqnarray}
\end{definition}

\begin{table}[ht!]
\centering
\setlength{\tabcolsep}{0.7em}
\renewcommand{\arraystretch}{1.2}
\begin{tabular}{c|c|c}
\hline
& $x^+$ and $X$ & $x^-$ and $X$   \\ \hline
Strong-alliance & $\Phi_i^{=}(x^+,X)\geq h_p$ & $\Phi_i^{=}(x^-,X)\geq h_n$ \\
Weak-alliance &   $l_p<\Phi_i^{=}(x^+,X)<h_p$ & $l_n<\Phi_i^{=}(x^-,X)<h_n$  \\
Non-alliance &  $\Phi_i^{=}(x^+,X)\leq l_p$ & $\Phi_i^{=}(x^-,X)\leq l_n$ \\ \hline
Strong-conflict & $\Phi_i^{\asymp}(x^+,X)\geq h_n$ & $\Phi_i^{\asymp}(x^-,X)\geq h_p$ \\
Weak-conflict &   $l_n<\Phi_i^{\asymp}(x^+,X)<h_n$ & $l_p<\Phi_i^{\asymp}(x^-,X)<h_p$  \\
Non-conflict &  $\Phi_i^{\asymp}(x^+,X)\leq l_n$ & $\Phi_i^{\asymp}(x^-,X)\leq l_p$ \\ \hline
\end{tabular}
\caption{The cases of alliance and conflict between $x^+/x^-$ and $X$}
\label{tab:alliance_conflict_x+-X}
\end{table}

One may also understand the four trisections in Definition~\ref{def:trisection_issue_X_our} from the views of alliance and conflict relations between $x^+/x^-$ and $X$. As given in Table~\ref{tab:alliance_conflict_x+-X}, by applying a pair of thresholds to an alliance function, we consider the three cases of strong-alliance, weak-alliance, and non-alliance, and similarly for conflict. Accordingly, for the trisection in Definition~\ref{def:trisection_issue_X_our} (1), $J_X^+(\Phi^{=},\Phi^{\asymp})$ includes the issues where $x^+$ and $X$ are strongly allied but not in conflict. Thus, $X$ can be reasonably considered to have a positive rating $+1$ on these issues. Similarly, $J_X^-(\Phi^{=},\Phi^{\asymp})$ includes the issues where $x^+$ and $X$ are strongly in conflict but not allied. Thus, $X$ can be reasonably considered to have a negative rating $-1$ on these issues. Otherwise, $X$ has a neutral rating $0$. One can easily make a similar analysis about the other three trisections in Definition \ref{def:trisection_issue_X_our}.

According to Definition~\ref{def:trisection_issue_X_our}, one can easily define the corresponding four trisections of issues with respect to a single agent $x\in A$ by degenerating $X$ to $\{x\}$. These trisections are equivalent with each other and with the corresponding trisection based on a rating function (i.e., Definition~\ref{def:trisection_issue_x_existing}).

\begin{theorem}
\label{theorem:trisection_issue_x_our_relation}
With respect to a single agent $x\in A$, the trisections of a set of issues $J \subseteq I$ with a rating function and with any pair of alliance and conflict functions in Table \ref{tab:combine_alliance_conflict_function_ix+-X} are equivalent, that is:
\begin{eqnarray}
\label{equa:trisection_issue_x_our_relation}
&&J_{x}^{+}(\Phi^{=},\Phi^{\asymp})=J_{x}^{+}(\Phi^{=},\Phi^{=})=J_{x}^{+}(\Phi^{\asymp},\Phi^{\asymp})=J_{x}^{+}(\Phi^{\asymp},\Phi^{=})=J_{x}^{+},\nonumber\\
&&J_{x}^{-}(\Phi^{=},\Phi^{\asymp})=J_{x}^{-}(\Phi^{=},\Phi^{=})=J_{x}^{-}(\Phi^{\asymp},\Phi^{\asymp})=J_{x}^{-}(\Phi^{\asymp},\Phi^{=})=J_{x}^{-},\nonumber\\
&&J_{x}^{0}(\Phi^{=},\Phi^{\asymp})=J_{x}^{0}(\Phi^{=},\Phi^{=})=J_{x}^{0}(\Phi^{\asymp},\Phi^{\asymp})=J_{x}^{0}(\Phi^{\asymp},\Phi^{=})=J_{x}^{0},
\end{eqnarray}
where we use $x$ to represent the singleton set $\{x\}$ for simplicity.
\end{theorem}
\begin{proof}
Since $0\leq l_p< h_p\leq 1$ and $0\leq l_n< h_n\leq 1$, we have $l_p<1$, $h_p>0$, $l_n<1$, and $h_n>0$. 
By Definitions~\ref{def:trisection_issue_X_our} and \ref{def:alliance_conflict_function_ixy}, we get the first trisection of issues with respect to $x$ as:
\begin{eqnarray}
J_{x}^{+}(\Phi^{=},\Phi^{\asymp})&=&\{i\in J\mid \Phi_i^{=}(x^+,x)\geq h_p > 0\wedge \Phi_i^{\asymp}(x^+,x)\leq l_n < 1\},\nonumber\\
&=&\{i\in J\mid \Phi_i^{=}(x^+,x)=1\wedge \Phi_i^{\asymp}(x^+,x)=0\},\nonumber\\
&=&\{i\in J\mid \Phi_i(x^+,x) = \Phi(r(x^+,i),r(x,i)) = \Phi(+1,r(x,i)) =+1\},\nonumber\\
&=&\{i\in J\mid r(x,i) =+1\},\nonumber\\
&=& J_x^+,\nonumber\\
J_{x}^{-}(\Phi^{=},\Phi^{\asymp})&=&\{i\in J\mid \Phi_i^{=}(x^+,x)\leq l_p < 1 \wedge \Phi_i^{\asymp}(x^+,x)\geq h_n > 0\},\nonumber\\
&=&\{i\in J\mid \Phi_i^{=}(x^+,x)=0\wedge \Phi_i^{\asymp}(x^+,x)=1\},\nonumber\\
&=&\{i\in J\mid \Phi_i(x^+,x) = \Phi(r(x^+,i),r(x,i)) = \Phi(+1,r(x,i)) =-1\},\nonumber\\
&=&\{i\in J\mid r(x,i) =-1\},\nonumber\\
&=& J_x^-,\nonumber\\
J_{x}^{0}(\Phi^{=},\Phi^{\asymp})&=&(J_{x}^{+}(\Phi^{=},\Phi^{\asymp})\cup J_{x}^{-}(\Phi^{=},\Phi^{\asymp}))^c\nonumber\\
&=&(J_x^+ \cup J_x^-)^c\nonumber\\
&=& J_x^0.
\end{eqnarray}
Similarly, one can prove that the other three trisections are also equivalent with the trisection of $J_x^+$, $J_x^-$, and $J_x^0$.
\end{proof}

Using a maximal consistent alliance set $M \in \mathcal{MA}_J$ as the set $X$ in Definition \ref{def:trisection_issue_X_our}, a subset of issues $J\subseteq I$ can be trisected into three pair-wise disjoint parts. Accordingly, we can formulate the decision of a maximal consistent alliance set as follows.

\begin{definition}
\label{def:decision_maximal_consistent_alliance_set}
Given a maximal consistent alliance set $M\in \mathcal{MA}_J$, let $J_M^+(\Phi^*,\Phi^\bullet)$, $J_M^-(\Phi^*,\Phi^\bullet)$, and $J_M^0(\Phi^*,\Phi^\bullet)$ denote a trisection of a set of issues $J \subseteq I$ where $*$ and $\bullet$ represent either $=$ or $\asymp$. Then the decision of $M$ is represented as:
\begin{eqnarray}
{\rm Des}_J(M)&=&(\bigwedge_{i\in J_{M}^{+}(\Phi^*,\Phi^\bullet)}\langle {i},+1\rangle)\wedge(\bigwedge_{i\in J_{M}^{-}(\Phi^*,\Phi^\bullet)}\langle {i},-1\rangle)\wedge(\bigwedge_{i\in J_{M}^{0}(\Phi^*,\Phi^\bullet)}\langle {i},0\rangle).
\end{eqnarray}
\end{definition}

Similarly, with respect to the non-neutral issues $J^{+-}_M= J_{M}^{+}(\Phi^*,\Phi^\bullet)\cup J_{M}^{-}(\Phi^*,\Phi^\bullet)$, one can easily get the decision of $M$ as:
\begin{eqnarray}
{\rm Des}_{J_M^{+-}}(M)&=&(\bigwedge_{i\in J_{M}^{+}(\Phi^*,\Phi^\bullet)}\langle {i},+1\rangle)\wedge(\bigwedge_{i\in J_{M}^{-}(\Phi^*,\Phi^\bullet)}\langle {i},-1\rangle).
\end{eqnarray}

\subsection{Examples}

We give two examples to illustrate the concepts introduced in this section. Particularly, Example \ref{example:alliance_conflict_function_ix+-y_PY} uses Pawlak's and Yao's auxiliary functions to construct the alliance and conflict functions regarding $x^+$ and $x^-$ defined in Equation \eqref{equa:alliance_conflict_function_ix+-X}. Based on these functions, we continue with our discussion in Example \ref{example:alliance_conflict_function_PY_computation} in Section \ref{sec:functions} to illustrate the concepts of alliance sets, maximal consistent alliance sets, and the representation of their decisions, which is given in Example \ref{example:alliance_and_decision}.

\begin{example}[\textbf{Alliance and conflict functions regarding $x^+$ and $x^-$ based on Pawlak's and Yao's auxiliary functions}]
\label{example:alliance_conflict_function_ix+-y_PY}
 
We illustrate the alliance and conflict functions defined in Equation \eqref{equa:alliance_conflict_function_ix+-X} with Pawlak's and Yao's auxiliary functions. We have defined the unaggregated alliance and conflict functions with both Pawlak's and Yao's auxiliary functions in Example \ref{example:alliance_conflict_function_PY_definition} in Equation \eqref{equa:alliance_conflict_function_ixy_PY}. By using either $x^+$ or $x^-$ as the first parameter, we can compute the following unaggregated functions: for $i\in I$ and $y \in A$,
\begin{eqnarray}
\Phi^{= P}_i(x^+,y)=\Phi^{= Y}_i(x^+,y)
=\Phi^{\asymp P}_i(x^-,y)=\Phi^{\asymp Y}_i(x^-,y)
=\left\{ {\begin{array}{*{20}{lcl}}
1, & ~r(y,i)=+1,\\
& &\\
0,& \mbox{otherwise};\\
\end{array}} \right.\nonumber\\
\Phi^{= P}_i(x^-,y)=\Phi^{= Y}_i(x^-,y)=\Phi^{\asymp P}_i(x^+,y)=\Phi^{\asymp Y}_i(x^+,y)
=\left\{ {\begin{array}{*{20}{lcl}}
1,& ~r(y,i)=-1,\\
& &\\
0, & \mbox{otherwise}.\\
\end{array}} \right.
\end{eqnarray}
We use the rating $r(y,i)$ as the conditions, which can be easily induced from the original conditions regarding $\Phi^P_i$ or $\Phi^Y_i$ in Equation~\eqref{equa:alliance_conflict_function_ixy_PY}. Then we aggregate these functions as defined in Equation~\eqref{equa:alliance_conflict_function_ix+-X}, which leads to the following alliance and conflict functions: for $X \subseteq A$,
\begin{eqnarray}
\label{equa:alliance_conflict_function_ix+-X_PY}
&&\Phi_i^{=P}(x^+,X)=\Phi_i^{=Y}(x^+,X)=\Phi_i^{\asymp P}(x^-,X)=\Phi_i^{\asymp Y}(x^-,X)=\frac{|X_i^+|}{|X|},\nonumber\\
&&\Phi_i^{= P}(x^-,X)=\Phi_i^{= Y}(x^-,X)=\Phi_i^{\asymp P}(x^+,X)=\Phi_i^{\asymp Y}(x^+,X)=\frac{|X_i^-|}{|X|},
\end{eqnarray}
where $X_i^+$ and $X_i^-$ are as defined in Definition~\ref{def:trisection_agent_i_existing}
(i.e., $X_i^+=\{x\in X\mid r(x,i)=+1\}$ and $X_i^-=\{x\in X\mid r(x,i)=-1\}$).
Intuitively,
$\Phi_i^{=}(x^+,X)$ and $\Phi_i^{\asymp}(x^-,X)$ are the proportions of the agents in $X$ with a positive rating $+1$ on the issue $i$;
and $\Phi_i^{=}(x^-,X)$ and $\Phi_i^{\asymp}(x^+,X)$ are the proportions of those with a negative rating $-1$.
Accordingly, based on Pawlak's and Yao's auxiliary functions, we may get a combination of the positive and negative views by using $\frac{|X_i^+|}{|X|}$ and $\frac{|X_i^-|}{|X|}$.
\end{example}

\begin{example}[\textbf{Alliance sets, maximal consistent alliance sets, and their decisions}]
\label{example:alliance_and_decision}

Using the alliance and conflict functions defined in Example \ref{example:alliance_conflict_function_ix+-y_PY}, we continue with our discussion in Example~\ref{example:alliance_conflict_function_PY_computation} to illustrate the concepts of alliance sets, maximal consistent alliance sets, and their decisions. For simplicity, we illustrate these concepts with respect to the non-neutral issues. Let us consider a set of issues $J = I = \{i_1,i_2,i_3,i_4\}$.

Firstly, we illustrate the concept of alliance sets and their decisions. We apply two thresholds $h_a=\frac{1}{2}$ and $l_c=\frac{1}{3}$ in Equation~\eqref{equa:trisection_agentpair_J+-_our} to construct the alliance relation and accordingly, compute the alliance sets by Definition~\ref{def:alliance_set}. We get the following alliance sets with respect to the non-neutral issues:
\begin{alignat}{5}
\label{equa:example_alliance_sets}
&AS_{J^{+-}}(x_1)=\{x_1\},&\qquad\qquad&AS_{J^{+-}}(x_2)=\{x_2,x_3,x_4,x_{12}\},\nonumber\\
&AS_{J^{+-}}(x_3)=\{x_2,x_3,x_4,x_{12}\},&&AS_{J^{+-}}(x_4)=\{x_2,x_3,x_4,x_{12}\},\nonumber\\
&AS_{J^{+-}}(x_5)=\{x_5,x_6\},&&AS_{J^{+-}}(x_6)=\{x_5,x_6\},\nonumber\\
&AS_{J^{+-}}(x_7)=\{x_7,x_8,x_9,x_{10},x_{11},x_{12}\},&&AS_{J^{+-}}(x_8)=\{x_7,x_8,x_9,x_{10},x_{11},x_{12}\},\nonumber\\
&AS_{J^{+-}}(x_9)=\{x_7,x_8,x_9,x_{10},x_{11}\},&&AS_{J^{+-}}(x_{10})=\{x_7,x_8,x_9,x_{10},x_{11}\},\nonumber\\
&AS_{J^{+-}}(x_{11})=\{x_7,x_8,x_9,x_{10},x_{11}\},&&AS_{J^{+-}}(x_{12})=\{x_2,x_3,x_4,x_7,x_8,x_{12}\}.
\end{alignat}
Then we can simply represent the decision of an alliance set $AS_{J^{+-}}(x_i)$ as the description of $x_i$ with respect to its non-neutral issues $J^{+-}_{x_i}$. For example, we have:
\begin{eqnarray}
{\rm Des}(AS_{J^{+-}}(x_4))={\rm Des}_{J^{+-}}(x_4)=\langle i_1,+1\rangle\wedge\langle i_4,-1\rangle.
\end{eqnarray}

Secondly, we illustrate the concept of maximal consistent alliance sets. By Definition~\ref{def:maximal_consistent_alliance_set} and Equation \eqref{equa:example_alliance_sets}, we compute the maximal consistent alliance sets with respect to the non-neutral issues as:
\begin{alignat}{5}
\label{equa:example_maximal_consistent_alliance_sets}
&X_1=\{x_1\},
&\qquad&X_2=\{x_2,x_3,x_4,x_{12}\},&\qquad&X_3=\{x_5,x_{6}\},\nonumber \\
&X_4=\{x_7,x_8,x_9,x_{10},x_{11}\}, && X_5=\{x_7,x_8,x_{12}\}.&&
\end{alignat}

Thirdly, we illustrate the decisions of maximal consistent alliance sets. Let us consider the alliance and conflict functions induced from Pawlak's auxiliary function, and the first case of combination in Definition~\ref{def:trisection_issue_X_our}. It should be noted that according to Equation \eqref{equa:alliance_conflict_function_ix+-X_PY}, one can actually get the same results using any other combination in Definition~\ref{def:trisection_issue_X_our}. By applying a pair of thresholds $(\frac{1}{|X_k|},\frac{1}{2})$ $(k=1,2,3,4)$, we get the trisections of $J$ regarding each maximal consistent alliance set in Equation \eqref{equa:example_maximal_consistent_alliance_sets} as follows:
\begin{alignat}{5}
&(1)&\quad&J_{X_1}^{+}(\Phi_i^{=P},\Phi_i^{\asymp P})=\emptyset, &\qquad&J_{X_1}^{-}(\Phi_i^{=P},\Phi_i^{\asymp P})=\emptyset, &\qquad&J_{X_1}^{0}(\Phi_i^{=P},\Phi_i^{\asymp P})=\{i_1,i_2,i_3,i_4\};\nonumber\\
&(2)&&J_{X_2}^{+}(\Phi_i^{=P},\Phi_i^{\asymp P})=\{i_1,i_2\},&& J_{X_2}^{-}(\Phi_i^{=P},\Phi_i^{\asymp P})=\{i_3,i_4\},
 &&J_{X_2}^{0}(\Phi_i^{=P},\Phi_i^{\asymp P})=\emptyset;\nonumber\\
&(3)&&J_{X_3}^{+}(\Phi_i^{=P},\Phi_i^{\asymp P})=\{i_1,i_2,i_3,i_4\}, &&J_{X_3}^{-}(\Phi_i^{=P},\Phi_i^{\asymp P})=\emptyset, &&J_{X_3}^{0}(\Phi_i^{=P},\Phi_i^{\asymp P})=\emptyset;\nonumber\\
&(4)&&J_{X_4}^{+}(\Phi_i^{=P},\Phi_i^{\asymp P})=\emptyset,&& J_{X_4}^{-}(\Phi_i^{=P},\Phi_i^{\asymp P})=\{i_1,i_2,i_3\}, &&J_{X_4}^{0}(\Phi_i^{=P},\Phi_i^{\asymp P})=\{i_4\};\nonumber\\
&(5)&&J_{X_5}^{+}(\Phi_i^{=P},\Phi_i^{\asymp P})=\emptyset,&& J_{X_5}^{-}(\Phi_i^{=P},\Phi_i^{\asymp P})=\{i_1,i_2,i_3,i_4\}, &&J_{X_5}^{0}(\Phi_i^{=P},\Phi_i^{\asymp P})=\emptyset.
\end{alignat}
Then according to Definition~\ref{def:decision_maximal_consistent_alliance_set}, the decisions of the maximal consistent alliance sets in Equation \eqref{equa:example_maximal_consistent_alliance_sets} with respect to the non-neutral issues are given as:
\begin{eqnarray}
{\rm Des}_{J^{+-}}(X_1)&=&\emptyset,\nonumber\\
{\rm Des}_{J^{+-}}(X_2)&=&\langle i_1,+1\rangle\wedge\langle i_2,+1\rangle\wedge\langle i_3,-1\rangle\wedge\langle i_4,-1\rangle,\nonumber\\
{\rm Des}_{J^{+-}}(X_3)&=&\langle i_1,+1\rangle\wedge\langle i_2,+1\rangle\wedge\langle i_3,+1\rangle\wedge\langle i_4,+1\rangle,\nonumber\\
{\rm Des}_{J^{+-}}(X_4)&=&\langle i_1,-1\rangle\wedge\langle i_2,-1\rangle\wedge\langle i_3,-1\rangle,\nonumber\\
{\rm Des}_{J^{+-}}(X_5)&=&\langle i_1,-1\rangle\wedge\langle i_2,-1\rangle\wedge\langle i_3,-1\rangle\wedge\langle i_4,-1\rangle.
\end{eqnarray}
We use the empty set $\emptyset$ to denote a representation with no condition. This can be justified by the fact that a description can be easily transformed into an equivalent set representation, in particular, a set of issue-rating pairs with the conjunctive relation assumed.
\end{example}

\section{Strategies and their relationships with agents}
\label{sec:strategy}

\begin{figure}[!ht]
\centering
\scalebox{0.75}{
\begin{tikzpicture}[auto, >=stealth', node distance=1em,
block_multilines1/.style ={rectangle, draw=black, fill=white, text width=16em, text centered, minimum height=3em,rounded corners=4pt,thick},
block_multilines2/.style ={rectangle, draw=black, fill=white,text width=22em,text centered, minimum height=4.5em,rounded corners=4pt,thick},
block_multilines3/.style ={rectangle, draw=white, fill=white,text centered, minimum height=1.8em,rounded corners=4pt,thick},
block_multilines4/.style ={rectangle,dashed, draw=black, fill=white, text width=7em, minimum height=3em,text centered,rounded corners=4pt,thick},
noborder_center/.style ={rectangle,text centered, minimum height=2em}
]
\node [block_multilines1] (TT) {A three-valued situation table};
\node [block_multilines2,below=3em of TT] (A) {The family of strategies with respect to\\~\\~\\~};
\node [block_multilines4,below left=5em and -7em of TT] (J){\color{orange}a subset of issues $J\subseteq I$};
\node [block_multilines4,below right=5em and -7em of TT] (J+-){\color{blue}the non-neutral issues $J^{+-}\subseteq J$};
\node [block_multilines1,below =3em of A] (A3) {Trisecting agents $X\subseteq A$ based on a strategy\\~\\~\\~\\~\\~};
\node [block_multilines3, below =5.7em of A] (I) {$X$};
\node [block_multilines3, below =3em of I] (IN) {Neutral};
\node [block_multilines3, below left=3em and 2.4em of I] (IP) {Supporting};
\node [block_multilines3, below right=3em and 1.8em of I] (INE) {Opposing};
\draw [->,thick] (TT.south) -- (A.north);
\draw [->,thick] (A.south) -- (A3.north);
\draw [->,thick] (I.south) -- (IN.north);
\draw [->,thick] (I.south) -- (IP.north);
\draw [->,thick] (I.south) -- (INE.north);
\end{tikzpicture}}
\caption{A framework of the discussion in Section \ref{sec:strategy}}
\label{fig:framework_strategy}
\end{figure}
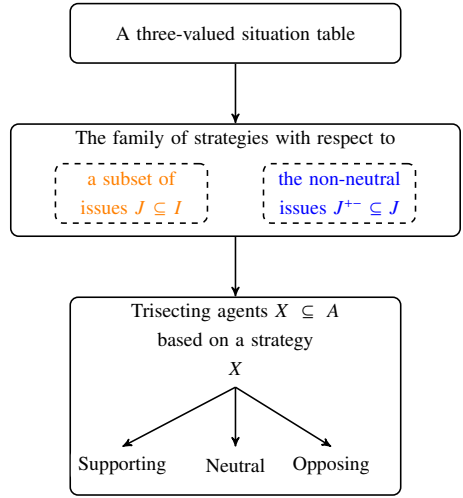

In this section, we apply the proposed alliance and conflict functions in studying another essential topic in conflict analysis, that is, strategy. Specifically, we present a formal representation of a strategy as a logic conjunction of issue-rating pairs. Based on it, we investigate the families of strategies with respect to a subset of issues and the non-neutral issues, as shown in Figure~\ref{fig:framework_strategy}. For each strategy, we discuss its relationship with the agents in a given subset $X\subseteq A$. Particularly, an agent may support, oppose, or be neutral to the strategy, which leads to a trisection of $X$. These attitudes from agents in $X$ on the strategy indicate the feasibility of a strategy in resolving the conflict. Examples are given at the end to illustrate the presented concepts.

\subsection{Describing strategies and their relationships with agents}

A strategy is usually defined as a set of issues in the existing works. The issues in the strategy are implicitly associated with a positive rating, and those not included in the strategy are associated with a non-positive rating that could be either negative or neutral. To further clarify these non-positive ratings, we present to list all the three types of positive, negative, and neutral ratings in a formal representation of a strategy. Particularly, we discuss two equivalent formats: a set representation and a logic representation. A set representation uses a set of issue-rating pairs to formulate a strategy $\mathcal{S}$ as follows:
\begin{eqnarray}
\mathcal{S}=\{\langle i,r_{i}\rangle\mid i\in J_{\mathcal{S}},r_{i}\in \{+1,-1,0\}\},
\end{eqnarray}
where $J_{\mathcal{S}}$ is the set of issues involved in $\mathcal{S}$ and $r_{i}$ is a rating on the issue $i\in J_{\mathcal{S}}$. The issue-rating pairs in the set are assumed to have a logic conjunction relation. Accordingly, we can alternatively adopt the following logic representation:
\begin{eqnarray}
\mathcal{S}=\bigwedge_{i\in J_{\mathcal{S}}}\langle i,r_{i}\rangle.
\end{eqnarray}
Furthermore, according to the ratings, the issues in $J_{\mathcal{S}}$ can be trisected into positive $J_{\mathcal{S}}^+$, negative $J_{\mathcal{S}}^-$, and neutral $J_{\mathcal{S}}^0$ sets. Thus, the strategy $\mathcal{S}$ can also be represented in the following format:
\begin{eqnarray}
\label{equa:strategy_logic_representation_+-0}
\mathcal{S}=(\bigwedge_{i\in J_{\mathcal{S}}^{+}}\langle {i},+1\rangle)\wedge(\bigwedge_{i\in J_{\mathcal{S}}^{-}}\langle {i},-1\rangle)\wedge(\bigwedge_{i\in J_{\mathcal{S}}^{0}}\langle {i},0\rangle).
\end{eqnarray}
The family of strategies is denoted as $\mathbb{S}$. For a given subset $J \subseteq I$, one can formulate $3^{|J|}$ strategies. Thus, for $|I|=n$, one can get the cardinality of $\mathbb{S}$ as:
\begin{eqnarray}
\label{equa:cardinality_strategy}
|\mathbb{S}|=C^0_{n}\times 3^0 + C^1_{n}\times 3^1+C^2_{n}\times 3^2+C^3_{n}\times 3^3+\cdots+C^{n}_{n}\times 3^{n}=4^{n},
\end{eqnarray}
where the empty strategy is also considered as valid.

Both the set and logic representations can also be recursively formulated through the concept of atomic strategies. An atomic strategy is defined as a strategy that involves only one issue. We can formulate three atomic strategies of positive, negative, and neutral ratings for every single issue. Any other strategy can be formulated through atomic strategies regarding the involved issues. We formally define strategies following this idea in terms of logic representations. One can easily formulate the equivalent set representations.

\begin{definition}
\label{def:atomic_composite_strategy}
The logic representation of a strategy can be defined by the following cases:
\begin{enumerate}[label=(\arabic*)]
\item The empty strategy is represented as $\emptyset$.
\item Atomic strategies: $\forall i \in I$, we can formulate three atomic strategies,
\begin{eqnarray*}
{\rm (+)} && {\rm A ~positive ~atomic ~strategy:~} \mathcal{S}_i^+=\langle i,+1\rangle;\\
{\rm (-)} && {\rm A ~negative ~atomic ~strategy:~} \mathcal{S}_i^-=\langle i,-1\rangle;\\
{\rm (0)} && {\rm A ~neutral ~atomic ~strategy:~} \mathcal{S}_i^0=\langle i,0\rangle.
\end{eqnarray*}
\item Composite strategies: $\forall J \subseteq I$, one can formulate a composite strategy on $J$ as,
\begin{eqnarray}
\mathcal{S} = \bigwedge_{i \in J} \mathcal{S}_i^*,
\end{eqnarray}
where $*$ represent $+$, $-$, or 0.
\end{enumerate}
\end{definition}

According to Definition \ref{def:atomic_composite_strategy}, we can further represent the strategy in Equation \eqref{equa:strategy_logic_representation_+-0} as:
\begin{eqnarray}
{\mathcal{S}}=(\bigwedge \limits_{i\in J_{{\mathcal{S}}}^{+}}{\mathcal{S}}_i^+)
\wedge(\bigwedge \limits_{i\in J_{{\mathcal{S}}}^{-}}{\mathcal{S}}_i^-)\wedge(\bigwedge \limits_{i\in J_{{\mathcal{S}}}^{0}}{\mathcal{S}}_i^0).
\end{eqnarray}
For example, suppose ${\mathcal{S}}=\langle i_1,+1\rangle\wedge\langle i_2,-1\rangle\wedge\langle i_3,0\rangle$. Then ${\mathcal{S}}$ is a composition of three atomic strategies ${\mathcal{S}}_{i_1}^+=\langle i_1,+1\rangle$, ${\mathcal{S}}_{i_2}^-=\langle i_2,-1\rangle$, and ${\mathcal{S}}_{i_3}^0=\langle i_3,0\rangle$, and can be represented as:
\begin{eqnarray}
{\mathcal{S}}={\mathcal{S}}_{i_1}^+\wedge{\mathcal{S}}_{i_2}^-\wedge{\mathcal{S}}_{i_3}^0.
\end{eqnarray}
By Definition \ref{def:atomic_composite_strategy}, one can compose a set of atomic strategies on different issues into a composite strategy and decompose a given strategy into a set of atomic strategies. The decomposition enables us to study a strategy through a set of atomic strategies. 

We arrive at a uniform formulation of describing a strategy in Equation \eqref{equa:strategy_logic_representation_+-0} and an agent in Equation \eqref{equa:description_agent}, that is, a conjunction of issue-rating pairs. Although taking the same format, these two descriptions have different semantics. The construction of a strategy may not involve any specific agent. The uniform representation suggests that we may consider a strategy as an imaginary ``agent''. Following this idea, we may explore the relationships between a strategy and an agent by using the same approaches to the relationships between agents that are discussed in Sections \ref{sec:functions} and \ref{sec:alliance_set_and_decision}. By modifying Definition \ref{def:alliance_conflict_function_ixy}, we define the unaggregated alliance and conflict functions regarding a strategy and an agent as given in the following definition.

\begin{definition}
Given an auxiliary function $\Phi_i$ regarding a single issue $i \in I$, the alliance function $\Phi_i^=: \mathbb{S} \times A \rightarrow \{0,1\}$ and the conflict function $\Phi_i^\asymp: \mathbb{S} \times A\rightarrow \{0,1\}$ are defined as:
\begin{eqnarray}
\label{equa:alliance_conflict_function_iSx}
\Phi_i^=(\mathcal{S},x)&=&
\left \{
\begin{array}{lcl}
1, & &\Phi_i(\mathcal{S},x)=+1,\\
& &\\
0,& &\Phi_i(\mathcal{S},x)\neq +1;\\
\end{array}
\right.\nonumber\\
~\nonumber\\
\Phi_i^{\asymp}(\mathcal{S},x)&=&
\left \{
\begin{array}{lcl}
1,& &\Phi_i(\mathcal{S},x)=-1,\\
& &\\
0, & &\Phi_i(\mathcal{S},x)\neq -1,
\end{array}
\right.
\end{eqnarray}
where $\Phi_i(\mathcal{S},x)$ is computed as if $\mathcal{S}$ were an agent.
\end{definition}

One can apply the same approaches discussed in Section \ref{sec:functions} to define the aggregated alliance and conflict functions regarding strategies and agents. In particular, we give the following aggregated functions that will be used in our following discussion:
\begin{eqnarray}
\label{equa:alliance_conflict_function_JSx}
\Phi_{J_{\mathcal{S}}}^=(\mathcal{S},x)&=&\frac{\sum\limits_{i\in {J_\mathcal{S}}}\Phi_i^{=}(\mathcal{S},x)}{|J_\mathcal{S}|}, \nonumber\\
\Phi_{J_{\mathcal{S}}}^{\asymp}(\mathcal{S},x)&=&\frac{\sum\limits_{i\in {J_\mathcal{S}}}\Phi_i^{\asymp}(\mathcal{S},x)}{|J_\mathcal{S}|}.
\end{eqnarray}

Accordingly, we may trisect a subset of agents as given in the following definition.

\begin{definition}
\label{def:trisection_agent_S_our}
For a subset of agents $X\subseteq A$ and a strategy $\mathcal{S}\in \mathbb{S}$, given two pairs of supporting and opposing thresholds $(l_s,h_s)$ and $(l_o,h_o)$ with $0\leq l_s< h_s\leq 1$ and $0\leq l_o< h_o\leq 1$, $X$ can be trisected into the supporting $X_{J_{\mathcal{S}}}^+(\Phi^=,\Phi^{\asymp})$, opposing $X_{J_{\mathcal{S}}}^-(\Phi^=,\Phi^{\asymp})$, and neutral $X_{J_{\mathcal{S}}}^0(\Phi^=,\Phi^{\asymp})$ sets regarding $\mathcal{S}$ as follows:
\begin{eqnarray}
\label{equa:trisection_agent_S_our}
X_{J_{\mathcal{S}}}^+(\Phi^=,\Phi^{\asymp})&=&\{x\in X\mid \Phi_{J_{\mathcal{S}}}^=(\mathcal{S},x)\geq h_s \wedge \Phi_{J_{\mathcal{S}}}^{\asymp}(\mathcal{S},x)\leq l_o\},\nonumber\\
X_{J_{\mathcal{S}}}^-(\Phi^=,\Phi^{\asymp})&=&\{x\in X\mid \Phi_{J_{\mathcal{S}}}^=(\mathcal{S},x)\leq l_s \wedge \Phi_{J_{\mathcal{S}}}^{\asymp}(\mathcal{S},x)\geq h_o\},\nonumber\\
X_{J_{\mathcal{S}}}^0(\Phi^=,\Phi^{\asymp})&=&(X_{J_{\mathcal{S}}}^+(\Phi^=,\Phi^{\asymp})\cup X_{J_{\mathcal{S}}}^-(\Phi^=,\Phi^{\asymp}))^c.
\end{eqnarray}
\end{definition}

One may also apply Definition~\ref{def:alliance_set} of alliance sets by considering a strategy $\mathcal{S}$ as an imaginary agent. Then the supporting set $X_{J_{\mathcal{S}}}^+(\Phi^=,\Phi^{\asymp})$ will induce the alliance set of $\mathcal{S}$ with respect to ${J_{\mathcal{S}}}$.

The alliance and conflict functions regarding a strategy and an agent in Equations \eqref{equa:alliance_conflict_function_iSx} and \eqref{equa:alliance_conflict_function_JSx} can also be computed through the functions regarding the involved atomic strategies.

\begin{theorem}
\label{theorem:alliance_conflict_function_JSx}
The alliance function $\Phi_{J_{{\mathcal{S}}}}^=: \mathbb{S} \times A \rightarrow \{0,1\}$ and the conflict function $\Phi_{J_{{\mathcal{S}}}}^\asymp: \mathbb{S} \times A\rightarrow \{0,1\}$ can be computed as: for $\mathcal{S} \in \mathbb{S}$ and $x \in A$,
\begin{eqnarray}
\Phi_{J_{{\mathcal{S}}}}^{=}({\mathcal{S}},x)&=&\frac{\sum\limits_{i \in J_{\mathcal{S}}}\Phi_{i}^{=}({\mathcal{S}}_{i}^*,x)}{n},\nonumber\\
\Phi_{J_{{\mathcal{S}}}}^{\asymp}({\mathcal{S}},x)&=&\frac{\sum\limits_{i \in J_{\mathcal{S}}}\Phi_{i}^{\asymp}({\mathcal{S}}_{i}^*,x)}{n},
\end{eqnarray}
where $*$ denotes $+$, $-$, or $0$.
\end{theorem}

For instance, for the strategy ${\mathcal{S}}={\mathcal{S}}_{i_1}^+\wedge{\mathcal{S}}_{i_2}^-\wedge{\mathcal{S}}_{i_3}^0$, we have:
\begin{eqnarray}
\Phi_{J_{{\mathcal{S}}}}^{=}({\mathcal{S}},x)&=&
\frac{\Phi_{i_1}^{=}({\mathcal{S}}_{i_1}^+,x)+
\Phi_{i_2}^{=}({\mathcal{S}}_{i_2}^-,x)+
\Phi_{i_3}^{=}({\mathcal{S}}_{i_3}^0,x)}{3},\nonumber\\
\Phi_{J_{{\mathcal{S}}}}^{\asymp}({\mathcal{S}},x)&=&
\frac{\Phi_{i_1}^{\asymp}({\mathcal{S}}_{i_1}^+,x)+
\Phi_{i_2}^{\asymp}({\mathcal{S}}_{i_2}^-,x)+
\Phi_{i_3}^{\asymp}({\mathcal{S}}_{i_3}^0,x)}{3}.
\end{eqnarray}

Our above discussion can be easily applied with respect to the non-neutral issues. For a given strategy $\mathcal{S} \in \mathbb{S}$, we have the set of its non-neutral issue as:
\begin{equation}
J_{\mathcal{S}}^{+-}=J_{\mathcal{S}}^{+}\cup J_{\mathcal{S}}^{-},
\end{equation}
that is, the set of issues with a non-neutral rating involved in $\mathcal{S}$. Then one can induce a non-neutral strategy $\widetilde{\mathcal{S}}$ from $\mathcal{S}$ as:
\begin{eqnarray}
\widetilde{\mathcal{S}}=(\bigwedge_{i\in J_{\mathcal{S}}^{+}}\langle {i},+1\rangle)\wedge(\bigwedge_{i\in J_{\mathcal{S}}^{-}}\langle {i},-1\rangle)=(\bigwedge \limits_{i\in J_{{\mathcal{S}}}^{+}}{\mathcal{S}}_i^+)
\wedge(\bigwedge \limits_{i\in J_{{\mathcal{S}}}^{-}}{\mathcal{S}}_i^-),
\end{eqnarray}
which removes all the issue-rating pairs with a neutral rating from $\mathcal{S}$. The family of non-neutral strategies is denoted as $\widetilde{\mathbb{S}}$. For a given subset $J \subseteq I$, one can formulate $2^{|J|}$ non-neutral strategies. Thus, for $|I|=n$, one can get the cardinality of $\widetilde{\mathbb{S}}$ as:
\begin{eqnarray}
|\widetilde{\mathbb{S}}|=C^0_{n}\times 2^0+C^1_{n}\times 2^1+C^2_{n}\times 2^2+C^3_{n}\times 2^3+\cdots+C^{n}_{n}\times 2^{n}=3^{n},
\end{eqnarray}
where the empty strategy is also considered as valid.

One may analyze the relationships between strategies and agents with respect to the non-neutral issues. In particular, we apply Definition~\ref{def:trisection_agent_S_our} with $J_{\widetilde{\mathcal{S}}}={J_{\mathcal{S}}^{+-}}=J_{\mathcal{S}}^{+}\cup J_{\mathcal{S}}^{-}$ instead of $J_{\mathcal{S}}$ to trisect a subset of agents $X\subseteq A$ as:
\begin{eqnarray}
\label{equa:trisection_agent_S+-_our}
X_{J_{\widetilde{\mathcal{S}}}}^+(\Phi^=,\Phi^{\asymp})&=&\{x\in X\mid \Phi_{J_{\widetilde{\mathcal{S}}}}^=({\widetilde{\mathcal{S}}},x)\geq h_s \wedge \Phi_{J_{\widetilde{\mathcal{S}}}}^{\asymp}({\widetilde{\mathcal{S}}},x)\leq l_o\},\nonumber\\
X_{J_{\widetilde{\mathcal{S}}}}^-(\Phi^=,\Phi^{\asymp})&=&\{x\in X\mid \Phi_{J_{\widetilde{\mathcal{S}}}}^=({\widetilde{\mathcal{S}}},x)\leq l_s \wedge \Phi_{J_{\widetilde{\mathcal{S}}}}^{\asymp}(\widetilde{\mathcal{S}},x)\geq h_o\},\nonumber\\
X_{J_{\widetilde{\mathcal{S}}}}^0(\Phi^=,\Phi^{\asymp})&=&(X_{J_{\widetilde{\mathcal{S}}}}^+(\Phi^=,\Phi^{\asymp})\cup X_{J_{\widetilde{\mathcal{S}}}}^-(\Phi^=,\Phi^{\asymp}))^c,
\end{eqnarray}
where the alliance and conflict functions are computed as:
\begin{eqnarray}
\label{equa:alliance_conflict_function_S+-x}
\Phi_{J_{\widetilde{\mathcal{S}}}}^{=}(\widetilde{\mathcal{S}},x)&=&\frac{\sum\limits_{i\in {J_{\widetilde{\mathcal{S}}}}}\Phi_i^{=}(\widetilde{\mathcal{S}},x)}{|J_{\widetilde{\mathcal{S}}}|},\nonumber\\
\Phi_{J_{\widetilde{\mathcal{S}}}}^{\asymp}(\widetilde{\mathcal{S}},x)&=&\frac{\sum\limits_{i\in {J_{\widetilde{\mathcal{S}}}}}\Phi_i^{\asymp}(\widetilde{\mathcal{S}},x)}{|J_{\widetilde{\mathcal{S}}}|}.
\end{eqnarray}
By Theorem \ref{theorem:alliance_conflict_function_JSx}, the two functions $\Phi_{J_{\widetilde{\mathcal{S}}}}^{=}(\widetilde{\mathcal{S}},x)$ and $\Phi_{J_{\widetilde{\mathcal{S}}}}^{\asymp}(\widetilde{\mathcal{S}},x)$ can be computed as the sum of the functions regarding the atomic non-neutral strategies involved in $\widetilde{\mathcal{S}}$.

\subsection{Examples}

We give two examples to illustrate the concepts introduced in this section. Particularly, Example \ref{example:alliance_conflict_function_strategy_PY} uses Pawlak's and Yao's auxiliary functions to construct the alliance and conflict functions regarding strategies and agents. Based on these functions, Example \ref{example:relation_strategy_agent} continues with our discussion in Example \ref{example:alliance_and_decision} in Section \ref{sec:alliance_set_and_decision} to illustrate the relationships between strategies and agents with respect to non-neutral issues.

\begin{example}[\textbf{Alliance and conflict functions regarding strategies and agents based on Pawlak's and Yao's auxiliary functions}]
\label{example:alliance_conflict_function_strategy_PY}
 
Based on Pawlak's and Yao's auxiliary functions, the alliance and conflict functions for a strategy $\mathcal{S}\in \mathbb{S}$ and an agent $x\in A$ with respect to a subset of issues $J\subseteq I$ are respectively defined as:
\begin{eqnarray}
\label{equa:alliance_conflict_function_strategy_PY}
\Phi_{J_\mathcal{S}}^{= P}(\mathcal{S},x)=\frac{\sum\limits_{i\in J_\mathcal{S}}\Phi_i^{= P}(\mathcal{S},x)}{|J_\mathcal{S}|}, \nonumber\\
\Phi_{J_\mathcal{S}}^{\asymp P}(\mathcal{S},x)=\frac{\sum\limits_{i\in J_\mathcal{S}}\Phi_i^{\asymp P}(\mathcal{S},x)}{|J_\mathcal{S}|}, \nonumber\\
\Phi_{J_\mathcal{S}}^{= Y}(\mathcal{S},x)=\frac{\sum\limits_{i\in J_\mathcal{S}}\Phi_i^{= Y}(\mathcal{S},x)}{|J_\mathcal{S}|}, \nonumber\\
\Phi_{J_\mathcal{S}}^{\asymp Y}(\mathcal{S},x)=\frac{\sum\limits_{i\in J_\mathcal{S}}\Phi_i^{\asymp Y}(\mathcal{S},x)}{|J_\mathcal{S}|}.
\end{eqnarray}

Similarly, the alliance and conflict functions with respect to a non-neutral strategy $\widetilde{\mathcal{S}}\in \widetilde{\mathbb{S}}$ and an agent $x\in A$ are defined as:
\begin{eqnarray}
\label{equa:alliance_conflict_function_S+-x_PY}
&&\Phi_{J_{\widetilde{\mathcal{S}}}}^{= Y}(\widetilde{\mathcal{S}},x)=\Phi_{J_{\widetilde{\mathcal{S}}}}^{= P}(\widetilde{\mathcal{S}},x)=\frac{\sum\limits_{i\in {J_{\widetilde{\mathcal{S}}}}}\Phi_i^{= P}(\widetilde{\mathcal{S}},x)}{|J_{\widetilde{\mathcal{S}}}|},\nonumber\\
&&\Phi_{J_{\widetilde{\mathcal{S}}}}^{\asymp Y}(\widetilde{\mathcal{S}},x)=\Phi_{J_{\widetilde{\mathcal{S}}}}^{\asymp P}(\widetilde{\mathcal{S}},x)=\frac{\sum\limits_{i\in {J_{\widetilde{\mathcal{S}}}}}\Phi_i^{\asymp P}(\widetilde{\mathcal{S}},x)}{|J_{\widetilde{\mathcal{S}}}|},
\end{eqnarray}
where
\begin{eqnarray}
\Phi_i^{=P}(\widetilde{\mathcal{S}},x)&=&
\left \{
\begin{array}{lcl}
1, & &\Phi^P(r(x_{\widetilde{\mathcal{S}}},i),r(x,i))=+1,\\
& &\\
0,& &\Phi^P(r(x_{\widetilde{\mathcal{S}}},i),r(x,i))\neq +1;\\
\end{array}
\right.\nonumber\\\nonumber\\
\Phi_i^{\asymp P}(\widetilde{\mathcal{S}},x)&=&
\left \{
\begin{array}{lcl}
1, & &\Phi^P(r(x_{\widetilde{\mathcal{S}}},i),r(x,i))=-1,\\
& &\\
0,& &\Phi^P(r(x_{\widetilde{\mathcal{S}}},i),r(x,i))\neq -1.\\
\end{array}
\right.
\end{eqnarray}
Furthermore, by the definition of Pawlak's auxiliary function given in Equation~\eqref{equa:auxiliary_ixy_Pawlak}, for an agent $x\in A$, we have the following duality:
\begin{eqnarray}
\label{equa:duality_auxiliaryP}
\Phi^P(+1,r(x,i))=-\Phi^P(-1,r(x,i)).
\end{eqnarray}
Accordingly, for any two strategies, if the set of issues with a positive rating $+1$ in one strategy is the same as the set of issues with a negative rating $-1$ in another and vice versa, they can be considered as two dual strategies. Formally, two dual strategies can be represented as:
\begin{eqnarray}
\widetilde{\mathcal{S}}&=&(\bigwedge_{i\in J_{\widetilde{\mathcal{S}}_{i}}^{+}}\langle {i},+1\rangle)\wedge(\bigwedge_{i\in J_{\widetilde{\mathcal{S}}_{i}}^{-}}\langle {i},-1\rangle), \nonumber\\
{\overline{\widetilde{\mathcal{S}}}}&=&(\bigwedge_{i\in J_{\widetilde{\mathcal{S}}_{i}}^{+}}\langle {i},-1\rangle)\wedge(\bigwedge_{i\in J_{\widetilde{\mathcal{S}}_{i}}^{-}}\langle {i},+1\rangle).
\end{eqnarray}

Then by Equations~\eqref{equa:alliance_conflict_function_S+-x_PY} and \eqref{equa:duality_auxiliaryP}, we have:
\begin{eqnarray}
\label{equa:duality_alliance_conflict_function_P}
\Phi_{J_{\widetilde{\mathcal{S}}}}^{= P}(\widetilde{\mathcal{S}},x)&=&\Phi_{J_{\overline{\widetilde{\mathcal{S}}}}}^{\asymp P}(\overline{\widetilde{\mathcal{S}}},x),\nonumber\\
\Phi_{J_{\widetilde{\mathcal{S}}}}^{\asymp P}(\widetilde{\mathcal{S}},x)&=&\Phi_{J_{\overline{\widetilde{\mathcal{S}}}}}^{= P}(\overline{\widetilde{\mathcal{S}}},x),
\end{eqnarray}
which also holds for those induced from Yao's auxiliary function. Therefore, based on Pawlak's and Yao's auxiliary functions, one can trisect a subset of agents $X\subseteq A$ with respect to the non-neutral issues $J_{\widetilde{\mathcal{S}}}$ by using the values of alliance functions only, which will significantly simplify our computation in the following Example \ref{example:relation_strategy_agent}. 
\end{example}

\begin{example}[\textbf{Relationships between strategies and agents}]
\label{example:relation_strategy_agent}

We continue with Example \ref{example:alliance_and_decision} to illustrate the non-neutral strategies and their relationships with agents based on the alliance and conflict functions defined in the above Example~\ref{example:alliance_conflict_function_strategy_PY}. 
Firstly, Table~\ref{example:strategy+-} lists the family of non-neutral strategies $\widetilde{\mathbb{S}}=\{\widetilde{\mathcal{S}}_0,\widetilde{\mathcal{S}}_1,\widetilde{\mathcal{S}}_2,\widetilde{\mathcal{S}}_3,\cdots,\widetilde{\mathcal{S}}_{80}\}$, where $\widetilde{\mathcal{S}}_0 = \emptyset$. We omit the empty strategy $\widetilde{\mathcal{S}}_0$ in Table~\ref{example:strategy+-} for simplicity.

\begin{table}[ht!]
\centering
\setlength{\tabcolsep}{0.5em}
\renewcommand{\arraystretch}{1.2}
\scalebox{0.75}{
\begin{tabular}{|c|c||c|c||c|c||c|c|}
\hline
Label&Representation&Label&Representation &Label&Representation &Label&Representation\\
\hline
$\widetilde{\mathcal{S}}_1$&$\langle i_1,+\rangle$
& $\widetilde{\mathcal{S}}_{21}$&$\langle i_2,+\rangle\wedge \langle i_3,+\rangle$
& $\widetilde{\mathcal{S}}_{41}$&$\langle i_1,+\rangle\wedge \langle i_2,+\rangle\wedge \langle i_4,+\rangle$
& $\widetilde{\mathcal{S}}_{61}$&$\langle i_2,+\rangle\wedge \langle i_3,-\rangle\wedge \langle i_4,+\rangle$\\
$\widetilde{\mathcal{S}}_2$&$\langle i_1,-\rangle$
& $\widetilde{\mathcal{S}}_{22}$&$\langle i_2,-\rangle\wedge \langle i_3,-\rangle$
& $\widetilde{\mathcal{S}}_{42}$&$\langle i_1,-\rangle\wedge \langle i_2,-\rangle\wedge \langle i_4,-\rangle$
& $\widetilde{\mathcal{S}}_{62}$&$\langle i_2,-\rangle\wedge \langle i_3,+\rangle\wedge \langle i_4,-\rangle$\\
$\widetilde{\mathcal{S}}_3$&$\langle i_2,+\rangle$
& $\widetilde{\mathcal{S}}_{23}$&$\langle i_2,+\rangle\wedge \langle i_3,-\rangle$
& $\widetilde{\mathcal{S}}_{43}$&$\langle i_1,+\rangle\wedge \langle i_2,+\rangle\wedge \langle i_4,-\rangle$
& $\widetilde{\mathcal{S}}_{63}$&$\langle i_2,-\rangle\wedge \langle i_3,+\rangle\wedge \langle i_4,+\rangle$\\
$\widetilde{\mathcal{S}}_4$&$\langle i_2,-\rangle$
& $\widetilde{\mathcal{S}}_{24}$&$\langle i_2,-\rangle\wedge \langle i_3,+\rangle$
& $\widetilde{\mathcal{S}}_{44}$&$\langle i_1,-\rangle\wedge \langle i_2,-\rangle\wedge \langle i_4,+\rangle$
& $\widetilde{\mathcal{S}}_{64}$&$\langle i_2,+\rangle\wedge \langle i_3,-\rangle\wedge \langle i_4,-\rangle$\\
$\widetilde{\mathcal{S}}_5$&$\langle i_3,+\rangle$
& $\widetilde{\mathcal{S}}_{25}$&$\langle i_2,+\rangle\wedge \langle i_4,+\rangle$
& $\widetilde{\mathcal{S}}_{45}$&$\langle i_1,+\rangle\wedge \langle i_2,-\rangle\wedge \langle i_4,+\rangle$
& $\widetilde{\mathcal{S}}_{65}$&$\langle i_1,+\rangle\wedge\langle i_2,+\rangle\wedge \langle i_3,+\rangle\wedge \langle i_4,+\rangle$\\
$\widetilde{\mathcal{S}}_6$&$\langle i_3,-\rangle$
& $\widetilde{\mathcal{S}}_{26}$&$\langle i_2,-\rangle\wedge \langle i_4,-\rangle$
& $\widetilde{\mathcal{S}}_{46}$&$\langle i_1,-\rangle\wedge \langle i_2,+\rangle\wedge \langle i_4,-\rangle$
& $\widetilde{\mathcal{S}}_{66}$&$\langle i_1,-\rangle\wedge\langle i_2,-\rangle\wedge \langle i_3,-\rangle\wedge \langle i_4,-\rangle$\\
$\widetilde{\mathcal{S}}_7$&$\langle i_4,+\rangle$
& $\widetilde{\mathcal{S}}_{27}$&$\langle i_2,+\rangle\wedge \langle i_4,-\rangle$
& $\widetilde{\mathcal{S}}_{47}$&$\langle i_1,-\rangle\wedge \langle i_2,+\rangle\wedge \langle i_4,+\rangle$
& $\widetilde{\mathcal{S}}_{67}$&$\langle i_1,+\rangle\wedge \langle i_2,+\rangle\wedge \langle i_3,+\rangle\wedge \langle i_4,-\rangle$\\
$\widetilde{\mathcal{S}}_8$&$\langle i_4,-\rangle$
& $\widetilde{\mathcal{S}}_{28}$&$\langle i_2,-\rangle\wedge \langle i_4,+\rangle$
& $\widetilde{\mathcal{S}}_{48}$&$\langle i_1,+\rangle\wedge \langle i_2,-\rangle\wedge \langle i_4,-\rangle$
& $\widetilde{\mathcal{S}}_{68}$&$\langle i_1,-\rangle\wedge \langle i_2,-\rangle\wedge \langle i_3,-\rangle\wedge \langle i_4,+\rangle$\\
$\widetilde{\mathcal{S}}_9$&$\langle i_1,+\rangle\wedge \langle i_2,+\rangle$
& $\widetilde{\mathcal{S}}_{29}$&$\langle i_3,+\rangle\wedge \langle i_4,+\rangle$
& $\widetilde{\mathcal{S}}_{49}$&$\langle i_1,+\rangle\wedge \langle i_3,+\rangle\wedge \langle i_4,+\rangle$
& $\widetilde{\mathcal{S}}_{69}$&$\langle i_1,+\rangle\wedge \langle i_2,+\rangle\wedge \langle i_3,-\rangle\wedge \langle i_4,+\rangle$\\
$\widetilde{\mathcal{S}}_{10}$&$\langle i_1,-\rangle\wedge \langle i_2,-\rangle$
& $\widetilde{\mathcal{S}}_{30}$&$\langle i_3,-\rangle\wedge \langle i_4,-\rangle$
& $\widetilde{\mathcal{S}}_{50}$&$\langle i_1,-\rangle\wedge \langle i_3,-\rangle\wedge \langle i_4,-\rangle$
& $\widetilde{\mathcal{S}}_{70}$&$\langle i_1,-\rangle\wedge \langle i_2,-\rangle\wedge \langle i_3,+\rangle\wedge \langle i_4,-\rangle$\\
$\widetilde{\mathcal{S}}_{11}$&$\langle i_1,+\rangle\wedge \langle i_2,-\rangle$
& $\widetilde{\mathcal{S}}_{31}$&$\langle i_3,+\rangle\wedge \langle i_4,-\rangle$
& $\widetilde{\mathcal{S}}_{51}$&$\langle i_1,+\rangle\wedge \langle i_3,+\rangle\wedge \langle i_4,-\rangle$
& $\widetilde{\mathcal{S}}_{71}$&$\langle i_1,+\rangle\wedge \langle i_2,-\rangle\wedge \langle i_3,+\rangle\wedge \langle i_4,+\rangle$\\
$\widetilde{\mathcal{S}}_{12}$&$\langle i_1,-\rangle\wedge \langle i_2,+\rangle$
& $\widetilde{\mathcal{S}}_{32}$&$\langle i_3,-\rangle\wedge \langle i_4,+\rangle$
& $\widetilde{\mathcal{S}}_{52}$&$\langle i_1,-\rangle\wedge \langle i_3,-\rangle\wedge \langle i_4,+\rangle$
& $\widetilde{\mathcal{S}}_{72}$&$\langle i_1,-\rangle\wedge \langle i_2,+\rangle\wedge \langle i_3,-\rangle\wedge \langle i_4,-\rangle$\\
$\widetilde{\mathcal{S}}_{13}$&$\langle i_1,+\rangle\wedge \langle i_3,+\rangle$
& $\widetilde{\mathcal{S}}_{33}$&$\langle i_1,+\rangle\wedge i_2,+\rangle\wedge\langle i_3,+\rangle$
& $\widetilde{\mathcal{S}}_{53}$&$\langle i_1,+\rangle\wedge \langle i_3,-\rangle\wedge \langle i_4,+\rangle$
& $\widetilde{\mathcal{S}}_{73}$&$\langle i_1,-\rangle\wedge \langle i_2,+\rangle\wedge \langle i_3,+\rangle\wedge \langle i_4,+\rangle$\\
$\widetilde{\mathcal{S}}_{14}$&$\langle i_1,-\rangle\wedge \langle i_3,-\rangle$
& $\widetilde{\mathcal{S}}_{34}$&$\langle i_1,-\rangle\wedge \langle i_2,-\rangle\wedge \langle i_3,-\rangle$
& $\widetilde{\mathcal{S}}_{54}$&$\langle i_1,-\rangle\wedge \langle i_3,+\rangle\wedge \langle i_4,-\rangle$
& $\widetilde{\mathcal{S}}_{74}$&$\langle i_1,+\rangle\wedge \langle i_2,-\rangle\wedge \langle i_3,-\rangle\wedge \langle i_4,-\rangle$\\
$\widetilde{\mathcal{S}}_{15}$&$\langle i_1,+\rangle\wedge \langle i_3,-\rangle$
& $\widetilde{\mathcal{S}}_{35}$&$\langle i_1,+\rangle\wedge \langle i_2,+\rangle\wedge \langle i_3,-\rangle$
& $\widetilde{\mathcal{S}}_{55}$&$\langle i_1,-\rangle\wedge \langle i_3,+\rangle\wedge \langle i_4,+\rangle$
& $\widetilde{\mathcal{S}}_{75}$&$\langle i_1,+\rangle\wedge \langle i_2,+\rangle\wedge \langle i_3,-\rangle\wedge \langle i_4,-\rangle$\\
$\widetilde{\mathcal{S}}_{16}$&$\langle i_1,-\rangle\wedge \langle i_3,+\rangle$
& $\widetilde{\mathcal{S}}_{36}$&$\langle i_1,-\rangle\wedge \langle i_2,-\rangle\wedge \langle i_3,+\rangle$
& $\widetilde{\mathcal{S}}_{56}$&$\langle i_1,+\rangle\wedge \langle i_3,-\rangle\wedge \langle i_4,-\rangle$
& $\widetilde{\mathcal{S}}_{76}$&$\langle i_1,-\rangle\wedge \langle i_2,-\rangle\wedge \langle i_3,+\rangle\wedge \langle i_4,+\rangle$\\
$\widetilde{\mathcal{S}}_{17}$&$\langle i_1,+\rangle\wedge \langle i_4,+\rangle$
& $\widetilde{\mathcal{S}}_{37}$&$\langle i_1,+\rangle\wedge \langle i_2,-\rangle\wedge \langle i_3,+\rangle$
& $\widetilde{\mathcal{S}}_{57}$&$\langle i_2,+\rangle\wedge \langle i_3,+\rangle\wedge \langle i_4,+\rangle$
& $\widetilde{\mathcal{S}}_{77}$&$\langle i_1,+\rangle\wedge \langle i_2,-\rangle\wedge \langle i_3,-\rangle\wedge \langle i_4,+\rangle$\\
$\widetilde{\mathcal{S}}_{18}$&$\langle i_1,-\rangle\wedge \langle i_4,-\rangle$
& $\widetilde{\mathcal{S}}_{38}$&$\langle i_1,-\rangle\wedge \langle i_2,+\rangle\wedge \langle i_3,-\rangle$
& $\widetilde{\mathcal{S}}_{58}$&$\langle i_2,-\rangle\wedge \langle i_3,-\rangle\wedge \langle i_4,-\rangle$
& $\widetilde{\mathcal{S}}_{78}$&$\langle i_1,-\rangle\wedge \langle i_2,+\rangle\wedge \langle i_3,+\rangle\wedge \langle i_4,-\rangle$\\
$\widetilde{\mathcal{S}}_{19}$&$\langle i_1,+\rangle\wedge \langle i_4,-\rangle$
& $\widetilde{\mathcal{S}}_{39}$&$\langle i_1,-\rangle\wedge \langle i_2,+\rangle\wedge \langle i_3,+\rangle$
& $\widetilde{\mathcal{S}}_{59}$&$\langle i_2,+\rangle\wedge \langle i_3,+\rangle\wedge \langle i_4,-\rangle$
& $\widetilde{\mathcal{S}}_{79}$&$\langle i_1,+\rangle\wedge \langle i_2,-\rangle\wedge \langle i_3,+\rangle\wedge \langle i_4,-\rangle$\\
$\widetilde{\mathcal{S}}_{20}$&$\langle i_1,-\rangle\wedge \langle i_4,+\rangle$
& $\widetilde{\mathcal{S}}_{40}$&$\langle i_1,+\rangle\wedge \langle i_2,-\rangle\wedge \langle i_3,-\rangle$
& $\widetilde{\mathcal{S}}_{60}$&$\langle i_2,-\rangle\wedge \langle i_3,-\rangle\wedge \langle i_4,+\rangle$
& $\widetilde{\mathcal{S}}_{80}$&$\langle i_1,-\rangle\wedge \langle i_2,+\rangle\wedge \langle i_3,-\rangle\wedge \langle i_4,+\rangle$\\
\hline
\end{tabular}
}
\caption{The family of nonempty non-neutral strategies $\widetilde{\mathbb{S}}$}
\label{example:strategy+-}
\end{table}

Secondly, by using the values computed from Equation~\eqref{equa:duality_alliance_conflict_function_P}, one can get a trisection of $X=A$ with respect to each strategy in Table~\ref{example:strategy+-}. The values of the alliance functions regarding each non-neutral strategy and each agent are shown in Table~\ref{tab:example_alliance_strategy_agent}. We take $\widetilde{\mathcal{S}}_{11}=\widetilde{\mathcal{S}}_{1}\wedge \widetilde{\mathcal{S}}_{4}$ as an illustration of the computation:
\begin{eqnarray}
\Phi_{J_{\widetilde{\mathcal{S}}_{11}}}^{= P}(\widetilde{\mathcal{S}}_{11},x_2)=\frac{\Phi_{J_{\widetilde{\mathcal{S}}_{1}}}^{= P}(\widetilde{\mathcal{S}}_{1},x_2)+\Phi_{J_{\widetilde{\mathcal{S}}_{4}}}^{= P}(\widetilde{\mathcal{S}}_{4},x_2)}{2}
=\frac{1+0}{2}=\frac{1}{2}.
\end{eqnarray}

\begin{table}
\centering
\setlength{\tabcolsep}{0.5em}
\renewcommand{\arraystretch}{1.2}
\scalebox{0.75}{
\begin{tabular}{|c|cccccccccccc||c|cccccccccccc|}
\hline
{Label}&$x_1$&$x_2$&$x_3$&$x_4$&$x_5$&$x_6$&$x_7$&$x_8$&$x_9$&$x_{10}$&$x_{11}$&$x_{12}$&{Label}
&$x_1$&$x_2$&$x_3$&$x_4$&$x_5$&$x_6$&$x_7$&$x_8$&$x_9$&$x_{10}$&$x_{11}$&$x_{12}$\\
\hline
$\Phi_{J_{\widetilde{\mathcal{S}}_1}}^{= P}(\widetilde{\mathcal{S}}_1,x)$
&$0$&1&1&1&1&1&$0$&$0$&$0$&$0$&$0$&1&
$\Phi_{J_{\widetilde{\mathcal{S}}_{41}}}^{= P}(\widetilde{\mathcal{S}}_{41},x)
$&$0$&$\frac{2}{3}$&$\frac{2}{3}$&$\frac{1}{3}$&1&1&$0$&$0$&$\frac{1}{3}$&$0$&$0$&$\frac{1}{3}$  \\
$\Phi_{J_{\widetilde{\mathcal{S}}_2}}^{= P}(\widetilde{\mathcal{S}}_2,x)$
&$0$&$0$&$0$&$0$&$0$&$0$&1&1&1&1&1&$0$&
$\Phi_{J_{\widetilde{\mathcal{S}}_{42}}}^{= P}(\widetilde{\mathcal{S}}_{42},x)$
&$0$&$\frac{1}{3}$&$\frac{1}{3}$&$\frac{1}{3}$&$0$&$0$&1&1&$\frac{2}{3}$&$\frac{2}{3}$&$\frac{2}{3}$&$\frac{2}{3}$  \\
$\Phi_{J_{\widetilde{\mathcal{S}}_3}}^{= P}(\widetilde{\mathcal{S}}_3,x)$
&$0$&1&1&$0$&1&1&$0$&$0$&$0$&$0$&$0$&$0$&
$\Phi_{J_{\widetilde{\mathcal{S}}_{43}}}^{= P}(\widetilde{\mathcal{S}}_{43},x)$
&$0$&1&1&$\frac{2}{3}$&$\frac{2}{3}$&$\frac{2}{3}$&$\frac{1}{3}$&$\frac{1}{3}$&$0$&$0$&$0$&$\frac{2}{3}$ \\
$\Phi_{J_{\widetilde{\mathcal{S}}_4}}^{= P}(\widetilde{\mathcal{S}}_4,x)$
&$0$&$0$&$0$&$0$&$0$&$0$&1&1&1&1&1&1&
$\Phi_{J_{\widetilde{\mathcal{S}}_{44}}}^{= P}(\widetilde{\mathcal{S}}_{44},x)$
&$0$&$0$&$0$&$0$&$\frac{1}{3}$&$\frac{1}{3}$&$\frac{2}{3}$&$\frac{2}{3}$&1&$\frac{2}{3}$&$\frac{2}{3}$&$\frac{1}{3}$  \\
$\Phi_{J_{\widetilde{\mathcal{S}}_5}}^{= P}(\widetilde{\mathcal{S}}_5,x)$
&$0$&$0$&$0$&$0$&1&1&$0$&$0$&$0$&$0$&$0$&$0$&
$\Phi_{J_{\widetilde{\mathcal{S}}_{45}}}^{= P}(\widetilde{\mathcal{S}}_{45},x)$
&$0$&$\frac{1}{3}$&$\frac{1}{3}$&$\frac{1}{3}$&$\frac{2}{3}$&$\frac{2}{3}$&$\frac{1}{3}$&$\frac{1}{3}$&$\frac{2}{3}$&$\frac{1}{3}$&$\frac{1}{3}$&$\frac{2}{3}$  \\
$\Phi_{J_{\widetilde{\mathcal{S}}_6}}^{= P}(\widetilde{\mathcal{S}}_6,x)$
&$0$&1&1&$0$&$0$&$0$&1&1&1&$0$&$0$&$0$&
$\Phi_{J_{\widetilde{\mathcal{S}}_{46}}}^{= P}(\widetilde{\mathcal{S}}_{46},x)$
&$0$&$\frac{2}{3}$&$\frac{2}{3}$&$\frac{1}{3}$&$\frac{1}{3}$&$\frac{1}{3}$&$\frac{2}{3}$&$\frac{2}{3}$&$\frac{1}{3}$&$\frac{1}{3}$&$\frac{1}{3}$&$\frac{1}{3}$  \\
$\Phi_{J_{\widetilde{\mathcal{S}}_7}}^{= P}(\widetilde{\mathcal{S}}_7,x)$
&$0$&$0$&$0$&$0$&1&1&$0$&$0$&1&$0$&$0$&$0$&
$\Phi_{J_{\widetilde{\mathcal{S}}_{47}}}^{= P}(\widetilde{\mathcal{S}}_{47},x)$
&$0$&$\frac{1}{3}$&$\frac{1}{3}$&$0$&$\frac{2}{3}$&$\frac{2}{3}$&$\frac{1}{3}$&$\frac{1}{3}$&$\frac{2}{3}$&$\frac{1}{3}$&$\frac{1}{3}$&0 \\
$\Phi_{{J_{\widetilde{\mathcal{S}}_8}^{= P}}}(\widetilde{\mathcal{S}}_8,x)$
&$0$&1&1&1&$0$&$0$&1&1&$0$&$0$&$0$&1&
$\Phi_{J_{\widetilde{\mathcal{S}}_{48}}}^{= P}(\widetilde{\mathcal{S}}_{48},x)$
&$0$&$\frac{2}{3}$&$\frac{2}{3}$&$\frac{2}{3}$&$\frac{1}{3}$&$\frac{1}{3}$&$\frac{2}{3}$&$\frac{2}{3}$&$\frac{1}{3}$&$\frac{1}{3}$&$\frac{1}{3}$&1  \\
$\Phi_{J_{\widetilde{\mathcal{S}}_9}}^{= P}(\widetilde{\mathcal{S}}_9,x)$
&$0$&1&1&$\frac{1}{2}$&1&1&$0$&$0$&$0$&$0$&$0$&$\frac{1}{2}$&
$\Phi_{J_{\widetilde{\mathcal{S}}_{49}}}^{= P}(\widetilde{\mathcal{S}}_{49},x)$
&$0$&$\frac{1}{3}$&$\frac{1}{3}$&$\frac{1}{3}$&1&1&$0$&$0$&$\frac{1}{3}$&$0$&$0$&$\frac{1}{3}$  \\
$\Phi_{J_{\widetilde{\mathcal{S}}_{10}}}^{= P}(\widetilde{\mathcal{S}}_{10},x)$
&$0$&$0$&$0$&$0$&$0$&$0$&1&1&1&1&1&$\frac{1}{2}$&
$\Phi_{J_{\widetilde{\mathcal{S}}_{50}}}^{= P}(\widetilde{\mathcal{S}}_{50},x)$
&$0$&$\frac{2}{3}$&$\frac{2}{3}$&$\frac{1}{3}$&$0$&$0$&1&1&$\frac{2}{3}$&$\frac{1}{3}$&$\frac{1}{3}$&$\frac{1}{3}$  \\
$\Phi_{J_{\widetilde{\mathcal{S}}_{11}}}^{= P}(\widetilde{\mathcal{S}}_{11},x)$
&$0$&$\frac{1}{2}$&$\frac{1}{2}$&$\frac{1}{2}$&$\frac{1}{2}$&$\frac{1}{2}$&$\frac{1}{2}$&$\frac{1}{2}$&$\frac{1}{2}$&$\frac{1}{2}$&$\frac{1}{2}$&1&
$\Phi_{J_{\widetilde{\mathcal{S}}_{51}}}^{= P}(\widetilde{\mathcal{S}}_{51},x)$
&$0$&$\frac{2}{3}$&$\frac{2}{3}$&$\frac{2}{3}$&$\frac{2}{3}$&$\frac{2}{3}$&$\frac{1}{3}$&$\frac{1}{3}$&$0$&$0$&$0$&$\frac{2}{3}$  \\
$\Phi_{J_{\widetilde{\mathcal{S}}_{12}}}^{= P}(\widetilde{\mathcal{S}}_{12},x)$
&$0$&$\frac{1}{2}$&$\frac{1}{2}$&$0$&$\frac{1}{2}$&$\frac{1}{2}$&$\frac{1}{2}$&$\frac{1}{2}$&$\frac{1}{2}$&$\frac{1}{2}$&$\frac{1}{2}$&$0$&
$\Phi_{J_{\widetilde{\mathcal{S}}_{52}}}^{= P}(\widetilde{\mathcal{S}}_{52},x)$
&$0$&$\frac{1}{3}$&$\frac{1}{3}$&$0$&$\frac{1}{3}$&$\frac{1}{3}$&$\frac{2}{3}$&$\frac{2}{3}$&1&$\frac{1}{3}$&$\frac{1}{3}$&$0$  \\
$\Phi_{J_{\widetilde{\mathcal{S}}_{13}}}^{= P}(\widetilde{\mathcal{S}}_{13},x)$
&$0$&$\frac{1}{2}$&$\frac{1}{2}$&$\frac{1}{2}$&1&1&$0$&$0$&$0$&$0$&$0$&$\frac{1}{2}$&
$\Phi_{J_{\widetilde{\mathcal{S}}_{53}}}^{= P}(\widetilde{\mathcal{S}}_{53},x)$
&$0$&$\frac{2}{3}$&$\frac{2}{3}$&$\frac{1}{3}$&$\frac{2}{3}$&$\frac{2}{3}$&$\frac{1}{3}$&$\frac{1}{3}$&$\frac{2}{3}$&$0$&$0$&$\frac{1}{3}$  \\
$\Phi_{J_{\widetilde{\mathcal{S}}_{14}}}^{= P}(\widetilde{\mathcal{S}}_{14},x)$
&$0$&$\frac{1}{2}$&$\frac{1}{2}$&$0$&$0$&$0$&1&1&1&$\frac{1}{2}$&$\frac{1}{2}$&$0$&
$\Phi_{J_{\widetilde{\mathcal{S}}_{54}}}^{= P}(\widetilde{\mathcal{S}}_{54},x)$
&$0$&$\frac{1}{3}$&$\frac{1}{3}$&$\frac{1}{3}$&$\frac{1}{3}$&$\frac{1}{3}$&$\frac{2}{3}$&$\frac{2}{3}$&$\frac{1}{3}$&$\frac{1}{3}$&$\frac{1}{3}$&$\frac{1}{3}$  \\
$\Phi_{J_{\widetilde{\mathcal{S}}_{15}}}^{= P}(\widetilde{\mathcal{S}}_{15},x)$
&$0$&1&1&$\frac{1}{2}$&$\frac{1}{2}$&$\frac{1}{2}$&$\frac{1}{2}$&$\frac{1}{2}$&$\frac{1}{2}$&$0$&$0$&$\frac{1}{2}$&
$\Phi_{J_{\widetilde{\mathcal{S}}_{55}}}^{= P}(\widetilde{\mathcal{S}}_{55},x)$
&$0$&$0$&$0$&$0$&$\frac{2}{3}$&$\frac{2}{3}$&$\frac{1}{3}$&$\frac{1}{3}$&$\frac{2}{3}$&$\frac{1}{3}$&$\frac{1}{3}$&0 \\
$\Phi_{J_{\widetilde{\mathcal{S}}_{16}}}^{= P}(\widetilde{\mathcal{S}}_{16},x)$
&$0$&$0$&$0$&$0$&$\frac{1}{2}$&$\frac{1}{2}$&$\frac{1}{2}$&$\frac{1}{2}$&$\frac{1}{2}$&$\frac{1}{2}$&$\frac{1}{2}$&$0$&
$\Phi_{J_{\widetilde{\mathcal{S}}_{56}}}^{= P}(\widetilde{\mathcal{S}}_{56},x)$
&$0$&1&1&$\frac{2}{3}$&$\frac{1}{3}$&$\frac{1}{3}$&$\frac{2}{3}$&$\frac{2}{3}$&$\frac{1}{3}$&$0$&$0$&$\frac{2}{3}$  \\
$\Phi_{J_{\widetilde{\mathcal{S}}_{17}}}^{= P}(\widetilde{\mathcal{S}}_{17},x)$
&$0$&$\frac{1}{2}$&$\frac{1}{2}$&$\frac{1}{2}$&1&1&$0$&$0$&$\frac{1}{2}$&$0$&$0$&$\frac{1}{2}$&
$\Phi_{J_{\widetilde{\mathcal{S}}_{57}}}^{= P}(\widetilde{\mathcal{S}}_{57},x)$
&$0$&$\frac{1}{3}$&$\frac{1}{3}$&$0$&1&1&$0$&$0$&$\frac{1}{3}$&$0$&$0$&0  \\
$\Phi_{J_{\widetilde{\mathcal{S}}_{18}}}^{= P}(\widetilde{\mathcal{S}}_{18},x)$
&$0$&$\frac{1}{2}$&$\frac{1}{2}$&$\frac{1}{2}$&$0$&$0$&1&1&$\frac{1}{2}$&$\frac{1}{2}$&$\frac{1}{2}$&$\frac{1}{2}$&
$\Phi_{J_{\widetilde{\mathcal{S}}_{58}}}^{= P}(\widetilde{\mathcal{S}}_{58},x)$
&$0$&$\frac{2}{3}$&$\frac{2}{3}$&$\frac{1}{3}$&$0$&$0$&1&1&$\frac{2}{3}$&$\frac{1}{3}$&$\frac{1}{3}$&$\frac{2}{3}$  \\
$\Phi_{J_{\widetilde{\mathcal{S}}_{19}}}^{= P}(\widetilde{\mathcal{S}}_{19},x)$
&$0$&1&1&1&$\frac{1}{2}$&$\frac{1}{2}$&$\frac{1}{2}$&$\frac{1}{2}$&$0$&$0$&$0$&1&
$\Phi_{J_{\widetilde{\mathcal{S}}_{59}}}^{= P}(\widetilde{\mathcal{S}}_{59},x)$
&$0$&$\frac{2}{3}$&$\frac{2}{3}$&$\frac{1}{3}$&$\frac{2}{3}$&$\frac{2}{3}$&$\frac{1}{3}$&$\frac{1}{3}$&$0$&$0$&$0$&$\frac{1}{3}$  \\
$\Phi_{J_{\widetilde{\mathcal{S}}_{20}}}^{= P}(\widetilde{\mathcal{S}}_{20},x)$
&$0$&$0$&$0$&$0$&$\frac{1}{2}$&$\frac{1}{2}$&$\frac{1}{2}$&$\frac{1}{2}$&1&$\frac{1}{2}$&$\frac{1}{2}$&$0$&
$\Phi_{J_{\widetilde{\mathcal{S}}_{60}}}^{= P}(\widetilde{\mathcal{S}}_{60},x)$
&$0$&$\frac{1}{3}$&$\frac{1}{3}$&$0$&$\frac{1}{3}$&$\frac{1}{3}$&$\frac{2}{3}$&$\frac{2}{3}$&1&$\frac{1}{3}$&$\frac{1}{3}$&$\frac{1}{3}$ \\
$\Phi_{J_{\widetilde{\mathcal{S}}_{21}}}^{= P}(\widetilde{\mathcal{S}}_{21},x)$
&$0$&$\frac{1}{2}$&$\frac{1}{2}$&$0$&1&1&$0$&$0$&$0$&0&0&0&
$\Phi_{J_{\widetilde{\mathcal{S}}_{61}}}^{= P}(\widetilde{\mathcal{S}}_{61},x)$
&$0$&$\frac{2}{3}$&$\frac{2}{3}$&$0$&$\frac{2}{3}$&$\frac{2}{3}$&$\frac{1}{3}$&$\frac{1}{3}$&$\frac{2}{3}$&$0$&$0$&0  \\
$\Phi_{J_{\widetilde{\mathcal{S}}_{22}}}^{= P}(\widetilde{\mathcal{S}}_{22},x)$
&$0$&$\frac{1}{2}$&$\frac{1}{2}$&$0$&$0$&$0$&1&1&1&$\frac{1}{2}$&$\frac{1}{2}$&$\frac{1}{2}$&
$\Phi_{J_{\widetilde{\mathcal{S}}_{62}}}^{= P}(\widetilde{\mathcal{S}}_{62},x)$
&$0$&$\frac{1}{3}$&$\frac{1}{3}$&$\frac{1}{3}$&$\frac{1}{3}$&$\frac{1}{3}$&$\frac{2}{3}$&$\frac{2}{3}$&$\frac{1}{3}$&$\frac{1}{3}$&$\frac{1}{3}$&$\frac{2}{3}$  \\
$\Phi_{J_{\widetilde{\mathcal{S}}_{23}}}^{= P}(\widetilde{\mathcal{S}}_{23},x)$
&$0$&1&1&$0$&$\frac{1}{2}$&$\frac{1}{2}$&$\frac{1}{2}$&$\frac{1}{2}$&$\frac{1}{2}$&$0$&$0$&$0$&
$\Phi_{J_{\widetilde{\mathcal{S}}_{63}}}^{= P}(\widetilde{\mathcal{S}}_{63},x)$
&$0$&$0$&$0$&$0$&$\frac{2}{3}$&$\frac{2}{3}$&$\frac{1}{3}$&$\frac{1}{3}$&$\frac{2}{3}$&$\frac{1}{3}$&$\frac{1}{3}$&$\frac{1}{3}$  \\
$\Phi_{J_{\widetilde{\mathcal{S}}_{24}}}^{= P}(\widetilde{\mathcal{S}}_{24},x)$
&$0$&$0$&$0$&$0$&$\frac{1}{2}$&$\frac{1}{2}$&$\frac{1}{2}$&$\frac{1}{2}$&$\frac{1}{2}$&$\frac{1}{2}$&$\frac{1}{2}$&$\frac{1}{2}$&
$\Phi_{J_{\widetilde{\mathcal{S}}_{64}}}^{= P}(\widetilde{\mathcal{S}}_{64},x)$
&$0$&1&1&$\frac{1}{3}$&$\frac{1}{3}$&$\frac{1}{3}$&$\frac{2}{3}$&$\frac{2}{3}$&$\frac{1}{3}$&$0$&$0$&$\frac{1}{3}$  \\
$\Phi_{J_{\widetilde{\mathcal{S}}_{25}}}^{= P}(\widetilde{\mathcal{S}}_{25},x)$
&$0$&$\frac{1}{2}$&$\frac{1}{2}$&$0$&1&1&$0$&$0$&$\frac{1}{2}$&$0$&$0$&$0$&
$\Phi_{J_{\widetilde{\mathcal{S}}_{65}}}^{= P}(\widetilde{\mathcal{S}}_{65},x)$
&$0$&$\frac{1}{2}$&$\frac{1}{2}$&$\frac{1}{4}$&1&1&$0$&$0$&$\frac{1}{4}$&$0$&$0$&$\frac{1}{4}$  \\
$\Phi_{J_{\widetilde{\mathcal{S}}_{26}}}^{= P}(\widetilde{\mathcal{S}}_{26},x)$
&$0$&$\frac{1}{2}$&$\frac{1}{2}$&$\frac{1}{2}$&$0$&$0$&1&1&$\frac{1}{2}$&$\frac{1}{2}$&$\frac{1}{2}$&1&
$\Phi_{J_{\widetilde{\mathcal{S}}_{66}}}^{= P}(\widetilde{\mathcal{S}}_{66},x)$
&$0$&$\frac{1}{2}$&$\frac{1}{2}$&$\frac{1}{4}$&$0$&$0$&1&1&$\frac{3}{4}$&$\frac{1}{2}$&$\frac{1}{2}$&$\frac{1}{2}$  \\
$\Phi_{J_{\widetilde{\mathcal{S}}_{27}}}^{= P}(\widetilde{\mathcal{S}}_{27},x)$
&$0$&1&1&$\frac{1}{2}$&$\frac{1}{2}$&$\frac{1}{2}$&$\frac{1}{2}$&$\frac{1}{2}$&$0$&$0$&$0$&$\frac{1}{2}$&
$\Phi_{J_{\widetilde{\mathcal{S}}_{67}}}^{= P}(\widetilde{\mathcal{S}}_{67},x)$
&$0$&$\frac{3}{4}$&$\frac{3}{4}$&$\frac{1}{2}$&$\frac{3}{4}$&$\frac{3}{4}$&$\frac{1}{4}$&$\frac{1}{4}$&$0$&$0$&$0$&$\frac{1}{2}$  \\
$\Phi_{J_{\widetilde{\mathcal{S}}_{28}}}^{= P}(\widetilde{\mathcal{S}}_{28},x)$
&$0$&$0$&$0$&$0$&$\frac{1}{2}$&$\frac{1}{2}$&$\frac{1}{2}$&$\frac{1}{2}$&1&$\frac{1}{2}$&$\frac{1}{2}$&$\frac{1}{2}$&
$\Phi_{J_{\widetilde{\mathcal{S}}_{68}}}^{= P}(\widetilde{\mathcal{S}}_{68},x)$
&$0$&$\frac{1}{4}$&$\frac{1}{4}$&$0$&$\frac{1}{4}$&$\frac{1}{4}$&$\frac{3}{4}$&$\frac{3}{4}$&1&$\frac{1}{2}$&$\frac{1}{2}$&$\frac{1}{4}$  \\
$\Phi_{J_{\widetilde{\mathcal{S}}_{29}}}^{= P}(\widetilde{\mathcal{S}}_{29},x)$
&$0$&$0$&$0$&$0$&1&1&$0$&$0$&$\frac{1}{2}$&$0$&$0$&$0$&
$\Phi_{J_{\widetilde{\mathcal{S}}_{69}}}^{= P}(\widetilde{\mathcal{S}}_{69},x)$
&$0$&$\frac{3}{4}$&$\frac{3}{4}$&$\frac{1}{4}$&$\frac{3}{4}$&$\frac{3}{4}$&$\frac{1}{4}$&$\frac{1}{4}$&$\frac{1}{2}$&$0$&$0$&$\frac{1}{4}$  \\
$\Phi_{J_{\widetilde{\mathcal{S}}_{30}}}^{= P}(\widetilde{\mathcal{S}}_{30},x)$
&$0$&1&1&$\frac{1}{2}$&$0$&$0$&1&1&$\frac{1}{2}$&$0$&$0$&$\frac{1}{2}$&
$\Phi_{J_{\widetilde{\mathcal{S}}_{70}}}^{= P}(\widetilde{\mathcal{S}}_{70},x)$
&$0$&$\frac{1}{4}$&$\frac{1}{4}$&$\frac{1}{4}$&$\frac{1}{4}$&$\frac{1}{4}$&$\frac{3}{4}$&$\frac{3}{4}$&$\frac{1}{2}$&$\frac{1}{2}$&$\frac{1}{2}$&$\frac{1}{2}$  \\
$\Phi_{J_{\widetilde{\mathcal{S}}_{31}}}^{= P}(\widetilde{\mathcal{S}}_{31},x)$
&$0$&$\frac{1}{2}$&$\frac{1}{2}$&$\frac{1}{2}$&$\frac{1}{2}$&$\frac{1}{2}$&$\frac{1}{2}$&$\frac{1}{2}$&$0$&$0$&$0$&$\frac{1}{2}$&
$\Phi_{J_{\widetilde{\mathcal{S}}_{71}}}^{= P}(\widetilde{\mathcal{S}}_{71},x)$
&$0$&$\frac{1}{4}$&$\frac{1}{4}$&$\frac{1}{4}$&$\frac{3}{4}$&$\frac{3}{4}$&$\frac{1}{4}$&$\frac{1}{4}$&$\frac{1}{2}$&$\frac{1}{4}$&$\frac{1}{4}$&$\frac{1}{2}$  \\
$\Phi_{J_{\widetilde{\mathcal{S}}_{32}}}^{= P}(\widetilde{\mathcal{S}}_{32},x)$
&$0$&$\frac{1}{2}$&$\frac{1}{2}$&$0$&$\frac{1}{2}$&$\frac{1}{2}$&$\frac{1}{2}$&$\frac{1}{2}$&1&$0$&$0$&$0$&
$\Phi_{J_{\widetilde{\mathcal{S}}_{72}}}^{= P}(\widetilde{\mathcal{S}}_{72},x)$
&$0$&$\frac{3}{4}$&$\frac{3}{4}$&$\frac{1}{4}$&$\frac{1}{4}$&$\frac{1}{4}$&$\frac{3}{4}$&$\frac{3}{4}$&$\frac{1}{2}$&$\frac{1}{4}$&$\frac{1}{4}$&$\frac{1}{4}$  \\
$\Phi_{J_{\widetilde{\mathcal{S}}_{33}}}^{= P}(\widetilde{\mathcal{S}}_{33},x)$
&$0$&$\frac{2}{3}$&$\frac{2}{3}$&$\frac{1}{3}$&1&1&$0$&$0$&$0$&$0$&$0$&$\frac{1}{3}$&
$\Phi_{J_{\widetilde{\mathcal{S}}_{73}}}^{= P}(\widetilde{\mathcal{S}}_{73},x)$
&$0$&$\frac{1}{4}$&$\frac{1}{4}$&$0$&$\frac{3}{4}$&$\frac{3}{4}$&$\frac{1}{4}$&$\frac{1}{4}$&$\frac{1}{2}$&$\frac{1}{4}$&$\frac{1}{4}$&0  \\
$\Phi_{J_{\widetilde{\mathcal{S}}_{34}}}^{= P}(\widetilde{\mathcal{S}}_{34},x)$
&$0$&$\frac{1}{3}$&$\frac{1}{3}$&$0$&$0$&$0$&1&1&1&$\frac{2}{3}$&$\frac{2}{3}$&$\frac{1}{3}$&
$\Phi_{J_{\widetilde{\mathcal{S}}_{74}}}^{= P}(\widetilde{\mathcal{S}}_{74},x)$
&$0$&$\frac{3}{4}$&$\frac{3}{4}$& $\frac{1}{2}$&$\frac{1}{4}$&$\frac{1}{4}$&$\frac{3}{4}$&$\frac{3}{4}$&$\frac{1}{2}$&$\frac{1}{4}$&$\frac{1}{4}$&$\frac{3}{4}$  \\
$\Phi_{J_{\widetilde{\mathcal{S}}_{35}}}^{= P}(\widetilde{\mathcal{S}}_{35},x)$
&$0$&1&1&$\frac{1}{3}$&$\frac{2}{3}$&$\frac{2}{3}$&$\frac{1}{3}$&$\frac{1}{3}$&$\frac{1}{3}$&$0$&$0$&$\frac{1}{3}$&
$\Phi_{J_{\widetilde{\mathcal{S}}_{75}}}^{= P}(\widetilde{\mathcal{S}}_{75},x)$
&$0$&1&1&$\frac{1}{2}$&$\frac{1}{2}$&$\frac{1}{2}$&$\frac{1}{2}$&$\frac{1}{2}$&$\frac{1}{4}$&$0$&$0$&$\frac{1}{2}$ \\
$\Phi_{J_{\widetilde{\mathcal{S}}_{36}}}^{= P}(\widetilde{\mathcal{S}}_{36},x)$
&$0$&$0$&$0$&$0$&$\frac{1}{3}$&$\frac{1}{3}$&$\frac{2}{3}$&$\frac{2}{3}$&$\frac{2}{3}$&$\frac{2}{3}$&$\frac{2}{3}$&$\frac{1}{3}$&
$\Phi_{J_{\widetilde{\mathcal{S}}_{76}}}^{= P}(\widetilde{\mathcal{S}}_{76},x)$
&$0$&$0$&$0$&$0$&$\frac{1}{2}$&$\frac{1}{2}$&$\frac{1}{2}$&$\frac{1}{2}$&$\frac{3}{4}$&$\frac{1}{2}$&$\frac{1}{2}$&$\frac{1}{4}$  \\
$\Phi_{J_{\widetilde{\mathcal{S}}_{37}}}^{= P}(\widetilde{\mathcal{S}}_{37},x)
$&$0$&$\frac{1}{3}$&$\frac{1}{3}$&$\frac{1}{3}$&$\frac{2}{3}$&$\frac{2}{3}$&$\frac{1}{3}$&$\frac{1}{3}$&$\frac{1}{3}$&$\frac{1}{3}$&$\frac{1}{3}$&$\frac{2}{3}$&
$\Phi_{J_{\widetilde{\mathcal{S}}_{77}}}^{= P}(\widetilde{\mathcal{S}}_{77},x)$
&$0$&$\frac{1}{2}$&$\frac{1}{2}$&$\frac{1}{4}$&$\frac{1}{2}$&$\frac{1}{2}$&$\frac{1}{2}$&$\frac{1}{2}$&$\frac{3}{4}$&$\frac{1}{4}$&$\frac{1}{4}$&$\frac{1}{2}$  \\
$\Phi_{J_{\widetilde{\mathcal{S}}_{38}}}^{= P}(\widetilde{\mathcal{S}}_{38},x)
$&$0$&$\frac{2}{3}$&$\frac{2}{3}$&$0$&$\frac{1}{3}$&$\frac{1}{3}$&$\frac{2}{3}$&$\frac{2}{3}$&$\frac{2}{3}$&$\frac{1}{3}$&$\frac{1}{3}$&$0$&
$\Phi_{J_{\widetilde{\mathcal{S}}_{78}}}^{= P}(\widetilde{\mathcal{S}}_{78},x)$
&$0$&$\frac{1}{2}$&$\frac{1}{2}$&$\frac{1}{4}$&$\frac{1}{2}$&$\frac{1}{2}$&$\frac{1}{2}$&$\frac{1}{2}$&$\frac{1}{4}$&$\frac{1}{4}$&$\frac{1}{4}$&$\frac{1}{4}$  \\
$\Phi_{J_{\widetilde{\mathcal{S}}_{39}}}^{= P}(\widetilde{\mathcal{S}}_{39},x)$
&$0$&$\frac{1}{3}$&$\frac{1}{3}$&$0$&$\frac{2}{3}$&$\frac{2}{3}$&$\frac{1}{3}$&$\frac{1}{3}$&$\frac{1}{3}$&$\frac{1}{3}$&$\frac{1}{3}$&$0$&
$\Phi_{J_{\widetilde{\mathcal{S}}_{79}}}^{= P}(\widetilde{\mathcal{S}}_{79},x)$
&$0$&$\frac{1}{2}$&$\frac{1}{2}$&$\frac{1}{2}$&$\frac{1}{2}$&$\frac{1}{2}$&$\frac{1}{2}$&$\frac{1}{2}$&$\frac{1}{4}$&$\frac{1}{4}$&$\frac{1}{4}$&$\frac{3}{4}$  \\
$\Phi_{J_{\widetilde{\mathcal{S}}_{40}}}^{= P}(\widetilde{\mathcal{S}}_{40},x)$
&$0$&$\frac{2}{3}$&$\frac{2}{3}$&$\frac{1}{3}$&$\frac{1}{3}$&$\frac{1}{3}$&$\frac{2}{3}$&$\frac{2}{3}$&$\frac{2}{3}$&$\frac{1}{3}$&$\frac{1}{3}$&$\frac{2}{3}$&
$\Phi_{J_{\widetilde{\mathcal{S}}_{80}}}^{= P}(\widetilde{\mathcal{S}}_{80},x)$
&$0$&$\frac{1}{2}$&$\frac{1}{2}$&$0$&$\frac{1}{2}$&$\frac{1}{2}$&$\frac{1}{2}$&$\frac{1}{2}$&$\frac{3}{4}$&$\frac{1}{4}$&$\frac{1}{4}$&0  \\
\hline
\end{tabular}
}
\caption{The values of the alliance functions regarding each strategy and each agent}
\label{tab:example_alliance_strategy_agent}
\end{table}

Finally, we illustrate the relationships between strategies and agents. By taking the thresholds $l_s=0$, $h_s=\frac{1}{2}$, $l_o=0$, and $h_o=\frac{1}{2}$, we get the trisection of $X$ with respect to $\widetilde{\mathcal{S}}_{10}$ as:
\begin{eqnarray}
X_{J_{\widetilde{\mathcal{S}}_{10}}}^+(\Phi^=,\Phi^{\asymp})&=&
\{x\in X\mid \Phi_{J_{\widetilde{\mathcal{S}}_{10}}}^{=P}(\widetilde{\mathcal{S}}_{10},x)\geq \frac{1}{2} \wedge \Phi_{J_{\widetilde{\mathcal{S}}_{10}}}^{\asymp P}(\widetilde{\mathcal{S}}_{10},x)\leq 0\},\nonumber\\
&=&\{x\in X\mid \Phi_{J_{\widetilde{\mathcal{S}}_{10}}}^{=P}(\widetilde{\mathcal{S}}_{10},x)\geq \frac{1}{2} \wedge \Phi_{J_{\widetilde{\mathcal{S}}_{9}}}^{=P}(\widetilde{\mathcal{S}}_{9},x)\leq 0\},\nonumber\\
&=&\{x_7,x_8,x_9,x_{10},x_{11}\},\nonumber\\
X_{J_{\widetilde{\mathcal{S}}_{10}}}^-(\Phi^=,\Phi^{\asymp})&=&
\{x\in X\mid \Phi_{J_{\widetilde{\mathcal{S}}_{10}}}^{\asymp P}(\widetilde{\mathcal{S}}_{10},x)\geq \frac{1}{2}\wedge \Phi_{J_{\widetilde{\mathcal{S}}_{10}}}^{=P}(\widetilde{\mathcal{S}}_{10},x)\leq 0\},\nonumber\\
&=&\{x\in X\mid \Phi_{J_{\widetilde{\mathcal{S}}_{9}}}^{=P}(\widetilde{\mathcal{S}}_{9},x)\geq \frac{1}{2}\wedge \Phi_{J_{\widetilde{\mathcal{S}}_{10}}}^{=P}(\widetilde{\mathcal{S}}_{10},x)\leq 0\},\nonumber\\
&=&\{x_2,x_3,x_4,x_5,x_6\},\nonumber\\
X_{J_{\widetilde{\mathcal{S}}_{10}}}^0(\Phi^=,\Phi^{\asymp})&=&(X_{J_{\widetilde{\mathcal{S}}_{10}}}^+(\Phi^=,\Phi^{\asymp})\cup X_{J_{\widetilde{\mathcal{S}}_{10}}}^-(\Phi^=,\Phi^{\asymp}))^c\nonumber\\
&=&\{x_1,x_{12}\},
\end{eqnarray}
where $\widetilde{\mathcal{S}}_{9}$ and $\widetilde{\mathcal{S}}_{10}$ are dual strategies. By using the thresholds $l_s=0$, $h_s=\frac{1}{4}$, $l_o=0$, and $h_o=\frac{1}{4}$, we get the trisection of $X$ with respect to $\widetilde{\mathcal{S}}_{65}$ as:
\begin{eqnarray}
X_{J_{\widetilde{\mathcal{S}}_{65}}}^+(\Phi^=,\Phi^{\asymp})&=&\{x\in X\mid \Phi_{J_{\widetilde{\mathcal{S}}_{65}}}^{=P}(\widetilde{\mathcal{S}}_{65},x)\geq \frac{1}{4} \wedge \Phi_{J_{\widetilde{\mathcal{S}}_{65}}}^{\asymp P}(\widetilde{\mathcal{S}}_{65},x)\leq 0\},\nonumber\\
&=&\{x\in X\mid \Phi_{J_{\widetilde{\mathcal{S}}_{65}}}^{=P}(\widetilde{\mathcal{S}}_{65},x)\geq \frac{1}{4} \wedge \Phi_{J_{\widetilde{\mathcal{S}}_{66}}}^{=P}(\widetilde{\mathcal{S}}_{66},x)\leq 0\},\nonumber\\
&=&\{x_5,x_6\},\nonumber\\
X_{J_{\widetilde{\mathcal{S}}_{65}}}^-(\Phi^=,\Phi^{\asymp})&=&\{x\in X\mid \Phi_{J_{\widetilde{\mathcal{S}}_{65}}}^{\asymp P}(\widetilde{\mathcal{S}}_{65},x)\geq \frac{1}{4}\wedge \Phi_{J_{\widetilde{\mathcal{S}}_{65}}}^{=P}(\widetilde{\mathcal{S}}_{65},x)\leq 0\},\nonumber\\
&=&\{x\in X\mid \Phi_{J_{\widetilde{\mathcal{S}}_{66}}}^{=P}(\widetilde{\mathcal{S}}_{66},x)\geq \frac{1}{4}\wedge \Phi_{J_{\widetilde{\mathcal{S}}_{65}}}^{=P}(\widetilde{\mathcal{S}}_{65},x)\leq 0\},\nonumber\\
&=&\{x_7,x_8,x_{10},x_{11}\},\nonumber\\
X_{J_{\widetilde{\mathcal{S}}_{65}}}^0(\Phi^=,\Phi^{\asymp})&=&(X_{J_{\widetilde{\mathcal{S}}_{65}}}^+(\Phi^=,\Phi^{\asymp})\cup X_{J_{\widetilde{\mathcal{S}}_{65}}}^-(\Phi^=,\Phi^{\asymp}))^c\nonumber\\
&=&\{x_1,x_2,x_3,x_4,x_9,x_{12}\},
\end{eqnarray}
where $\widetilde{\mathcal{S}}_{65}$ and $\widetilde{\mathcal{S}}_{66}$ are dual strategies.
\end{example}

\section{Comparative analysis and an illustrative application}
\label{sec:comparative}

This section compares the proposed three-way conflict analysis models with two evaluation functions to the existing models with a single evaluation function. The comparative analysis demonstrates that the proposed model can assist us in gaining an in-depth understanding of conflict analysis. Additionally, we discuss an illustrative example that applies the proposed models to help the government address the conflict problems in making the development plan for Gansu province in China~\cite{Sun_2020}. 

\subsection{Comparative analysis}

In this paper, we investigate the trisections of agents, issues, and agent pairs in three-way conflict analysis, and discuss their applications in solving some common problems regarding conflict. The existing related studies are commonly based on a single evaluation function. Specifically, a rating function is used to trisect agents and issues, and an auxiliary function is used to trisect agent pairs. As discussed above, the models with a single evaluation function may face some challenges in interpreting aggregated ratings, especially when a neutral rating $0$ is involved. These challenges may further lead to difficulties in analyzing the trisections and understanding the corresponding relationships. To provide more precise semantics, we split an auxiliary function that measures alliance and conflict simultaneously into two separate functions of alliance and conflict. Table~\ref{tab:comparative_analysis} summarizes the comparisons between the existing three-way conflict analysis models with a single evaluation function and the proposed models with two evaluation functions. We discuss the comparisons in detail as follows.

\begin{table}[!ht]
\centering
\setlength{\tabcolsep}{0.5em}
\renewcommand{\arraystretch}{1.3}
\scalebox{0.64}{
\begin{tabular}{|c|c||c|c|c||c|c|c|}
  \hline
 \multicolumn{2}{|c||}{\backslashbox{Trisecting}{Model}} &\multicolumn{3}{c||}{A single evaluation function} &\multicolumn{3}{c|}{Two evaluation functions}\\ \hline
Type& w.r.t.& Formulation &  \multicolumn{2}{c||}{Evaluation function}& Formulation &Evaluation function&Special evaluation function\\\hline\hline
 
\multirow{6}{*}{Agents}& \multirow{2}{*}{$i \in I$}& \multirow{2}{*}{Definition~\ref{def:trisection_agent_i_existing}}& \multicolumn{2}{c||}{\multirow{2}*{{$r(x,i)$ in Definition~\ref{def:situation_table}}}}&  \multirow{2}{*}{Definition~\ref{def:trisection_agent_S_our}}&$\Phi_{J_{\mathcal{S}}}^=(\mathcal{S},x)$, $\Phi_{J_{\mathcal{S}}}^{\asymp}(\mathcal{S},x)$ in &$\Phi_i^{= P}(x,y)$, $\Phi_i^{\asymp P}(x,y)$ in Equation \eqref{equa:alliance_conflict_function_ixy_PY}\\\cline{8-8}
&&&\multicolumn{2}{c||}{\multirow{2}*{}}&&Equation \eqref{equa:alliance_conflict_function_JSx} for atomic strategy&$\Phi_i^{= Y}(x,y)$, $\Phi_i^{\asymp Y}(x,y)$ in Equation \eqref{equa:alliance_conflict_function_ixy_PY}\\\cline{2-8}

& \multirow{4}{*}{$J \subseteq I$}& \multirow{4}{*}{Definition~\ref{def:trisection_agent_J_existing}}& \multicolumn{2}{c||}{\multirow{4}*{{$r(x,J)$} in Equation ~\eqref{equa:aggregated_rating_issues}}}&   \multirow{2}{*}{Definition~\ref{def:trisection_agent_S_our}}&$\Phi_{J_{\mathcal{S}}}^=(\mathcal{S},x)$, $\Phi_{J_{\mathcal{S}}}^{\asymp}(\mathcal{S},x)$&$\Phi_{J_{\mathcal{S}}}^{=P}(\mathcal{S},x)$, $\Phi_{J_{\mathcal{S}}}^{\asymp P}(\mathcal{S},x)$ in Equation \eqref{equa:alliance_conflict_function_strategy_PY}\\\cline{8-8}
&&&\multicolumn{2}{c||}{\multirow{3}*{}}&&in Equation \eqref{equa:alliance_conflict_function_JSx}&$\Phi_{J_{\mathcal{S}}}^{=Y}(\mathcal{S},x)$, $\Phi_{J_{\mathcal{S}}}^{\asymp Y}(\mathcal{S},x)$ in Equation \eqref{equa:alliance_conflict_function_strategy_PY}\\
\cline{6-8}
&&&\multicolumn{2}{c||}{\multirow{4}*{}}& \multirow{2}{*}{Equation \eqref{equa:trisection_agent_S+-_our}}&$\Phi_{J_{\widetilde{\mathcal{S}}}}^{=}(\widetilde{\mathcal{S}},x)$, $\Phi_{J_{\widetilde{\mathcal{S}}}}^{\asymp}(\widetilde{\mathcal{S}},x)$&$\Phi_{J_{\widetilde{\mathcal{S}}}}^{= P}(\widetilde{\mathcal{S}},x)=\Phi_{J_{\widetilde{\mathcal{S}}}}^{=Y}(\widetilde{\mathcal{S}},x)$, \\
&&&\multicolumn{2}{c||}{\multirow{4}*{}}&&in Equation \eqref{equa:alliance_conflict_function_S+-x}&$\Phi_{J_{\widetilde{\mathcal{S}}}}^{\asymp P}(\widetilde{\mathcal{S}},x)=\Phi_{J_{\widetilde{\mathcal{S}}}}^{\asymp Y}(\widetilde{\mathcal{S}},x)$ in Equation \eqref{equa:alliance_conflict_function_S+-x_PY}\\\hline\hline

\multirow{4}{*}{Issues}& \multirow{2}{*}{$x \in A$}& \multirow{2}{*}{Definition~\ref{def:trisection_issue_x_existing}}& \multicolumn{2}{c||}{\multirow{2}*{{$r(x,i)$ in Definition~\ref{def:situation_table}}}}&  \multirow{2}{*}{Equation \eqref {equa:trisection_issue_x_our_relation}}&$\Phi_i^{=}(x,y)$, $\Phi_i^{\asymp}(x,y)$&$\Phi_i^{= P}(x,y)$, $\Phi_i^{\asymp P}(x,y)$ in Equation \eqref{equa:alliance_conflict_function_ixy_PY}\\\cline{8-8}
&&&\multicolumn{2}{c||}{\multirow{2}*{}}&&in Definition~\ref{def:alliance_conflict_function_ixy}&$\Phi_i^{= Y}(x,y)$, $\Phi_i^{\asymp Y}(x,y)$ in Equation \eqref{equa:alliance_conflict_function_ixy_PY}\\\cline{2-8}

& \multirow{2}{*}{$X \subseteq A$}& \multirow{2}{*}{Definition~\ref{def:trisection_issue_X_existing}}& \multicolumn{2}{c||}{\multirow{2}{*}{$r(X,i)$ in Equation  \eqref{equa:aggregated_rating_agents}}}&  \multirow{2}{*}{Definition~\ref{def:trisection_issue_X_our}}& Four pairs of&\multirow{2}{*}{$\frac{|X_i^+|}{|X|}$, $\frac{|X_i^-|}{|X|}$ in Equation \eqref{equa:alliance_conflict_function_ix+-X_PY}}\\
&&&\multicolumn{2}{c||}{}&& functions in Table~\ref{tab:combine_alliance_conflict_function_ix+-X}&\\\hline\hline

& \multirow{2}{*}{$i \in I$}& \multirow{2}{*}{Definition~\ref{def:trisection_agentpair_i_existing}}& $\Phi_i(x,y)$& 
$\Phi_i^{P}(x,y)$ in Equation \eqref{equa:auxiliary_ixy_Pawlak}&  \multirow{2}{*}{Equation \eqref{equa:trisection_agentpair_i_equivalent}}&$\Phi_i^{\asymp}(x,y)$, $\Phi_i^{\asymp}(x,y)$ &$\Phi_i^{= P}(x,y)$, $\Phi_i^{\asymp P}(x,y)$ in Equation \eqref{equa:alliance_conflict_function_ixy_PY}\\\cline{5-5}\cline{8-8}
&&&in Equation \eqref{equa:auxiliary_ixy_general}&$\Phi_i^{Y}(x,y)$ in Equation \eqref{equa:auxiliary_ixy_Yao}&&in Definition~\ref{def:alliance_conflict_function_ixy}&$\Phi_i^{= Y}(x,y)$, $\Phi_i^{\asymp Y}(x,y)$ in Equation \eqref{equa:alliance_conflict_function_ixy_PY}\\\cline{2-8}

Agent pairs & \multirow{4}{*}{$J \subseteq I$}& \multirow{4}{*}{Definition~\ref{def:trisection_agentpair_J_existing}}& & 
\multirow{2}{*}{$\Phi_J^{P}(x,y)$ in Equation \eqref{equa:auxiliary_Jxy_Pawlak}}&  \multirow{2}{*}{Definition~\ref{def:trisection_agentpair_J_our}}&$\Phi_J^{=}(x,y)$, $\Phi_J^{\asymp}(x,y)$&$\Phi_J^{=P}(x,y)$, $\Phi_J^{\asymp P}(x,y)$ in Equation \eqref{equa:alliance_conflict_function_Jxy_PY}\\\cline{8-8}
& &&$\Phi_J(x,y)$&&&in Equation \eqref{equa:alliance_conflict_function_Jxy}&$\Phi_J^{=Y}(x,y)$, $\Phi_J^{\asymp Y}(x,y)$ in Equation \eqref{equa:alliance_conflict_function_Jxy_PY}\\
\cline{5-8}
&&&in Equation  \eqref{equa:auxiliary_Jxy_general}&\multirow{2}{*}{$\Phi_J^{Y}(x,y)$ in Equation \eqref{equa:auxiliary_Jxy_Yao}}& \multirow{2}{*}{Equation ~\eqref{equa:trisection_agentpair_J+-_our}}&$\Phi_{J_x^{+-}}^{=}(x,y)$, $\Phi_{J_x^{+-}}^{\asymp}(x,y)$&$\Phi_{J_x^{+-}}^{=P}(x,y)=\Phi_{J_x^{+-}}^{=P}(x,y)$,\\
&&&&&&in Equation \eqref{equa:alliance_conflict_function_J+-xy}&$\Phi_{J_x^{+-}}^{\asymp P}(x,y)=\Phi_{J_x^{+-}}^{\asymp Y}(x,y)$ in Equation \eqref{equa:alliance_conflict_function_J+-xy_PY}\\
  \hline
\end{tabular}}
\caption{Comparisons between existing models with a single evaluation function and proposed models with two evaluation functions}
\label{tab:comparative_analysis}
\end{table}

\begin{enumerate}[label=(\arabic*)]
\item \textbf{Trisecting agents}

The trisections of agents in the existing and proposed models differ in two aspects, that is, formulation and semantics. Firstly, for formulation, with a single evaluation function, the trisections of agents with respect to a single issue $i$ and a subset of issues $J$ are based on the rating function $r(x,i)$ in Definition~\ref{def:situation_table} and its aggregation $r(x,J)$ in Equation~\eqref{equa:aggregated_rating_issues}, respectively. In both cases, the two aspects of alliance and conflict are measured in a single function. With two evaluation functions, we define the alliance and conflict functions regarding a strategy and an agent in Equation~\eqref{equa:alliance_conflict_function_JSx}. Based on these two functions, we define the trisections of agents with respect to a strategy as in Definition~\ref{def:trisection_agent_S_our}. Similarly, we also discuss a special case of using non-neutral strategies to define the alliance and conflict functions in Equation~\eqref{equa:alliance_conflict_function_S+-x} and accordingly, trisect agents in Equation \eqref{equa:trisection_agent_S+-_our}. It should be noted that in our model, we trisect the agents with respect to the issues involved in a strategy. Thus, a trisection of agents with respect to a single issue is a special case with respect to the corresponding atomic strategy.

Secondly, for semantics, we focus on different purposes when trisecting agents in the existing model with a single function and the proposed model with two functions. Specifically, the trisection of agents with a rating function divides a set of agents $X$ into three pair-wise disjoint sets with respect to a set of issues $J$: a set of agents who have a positive rating on most issues in $J$, a set of agents who have a negative rating on most issues in $J$, and a set of the remaining agents. 
In comparison, the trisection of agents with a pair of alliance and conflict functions divides $X$ into three pair-wise disjoint sets: a set of agents supporting the strategy, a set of agents opposing the strategy, and a set of agents neutral to the strategy.

\item \textbf{Trisecting issues}

We compare the trisections of issues in the existing and proposed models in the two aspects of formulation and semantics. Firstly, for formulation, with a single function of either the rating function $r(x,i)$ in Definition~\ref{def:situation_table} or $r(X,i)$ in Equation~\eqref{equa:aggregated_rating_agents}, one can respectively define the trisections of issues with respect to a single agent $x$ and a subset of agents $X$ (i.e, Definitions~\ref{def:trisection_issue_x_existing} and \ref{def:trisection_issue_X_existing}). 
In the proposed model with two functions, we define two pairs of alliance and conflict functions in Table~\ref{tab:combine_alliance_conflict_function_ix+-X} regarding two imaginary agents $x^+$ and $x^-$. Accordingly, we get four types of trisections of issues (i.e., Definition~\ref{def:trisection_issue_X_our}).  
According to Theorem~\ref{theorem:trisection_issue_x_our_relation}, with respect to a single agent, the trisections of issues based on either a single rating function or a pair of alliance and conflict functions are equivalent.

Secondly, for semantics, the rating function $r(X,i)$ aggregates the positive, negative, and neutral attitudes of agents on an issue $i$. As previously discussed, two opposing ratings have the same effect as two neutral ratings, which may create challenges in analyzing the trisection of issues. Therefore, we define the alliance and conflict functions, which can describe the different attitudes of agents to trisect issues. Both the existing model with a single function and the proposed model with two functions divide a set of issues $J$ into three pair-wise disjoint sets with respect to a set of agents $X$: a set of issues on which most of the agents in $X$ have positive attitudes, a set of issues on which most of the agents in $X$ have negative attitudes, and a set of the remaining issues.

\item \textbf{Trisecting agent pairs}

We also compare the trisections of agent pairs in the existing and proposed models in the two aspects of formulation and semantics. Firstly, for formulation, with a single evaluation function, the trisections of agent pairs with respect to a single issue $i$ and a subset of issues $J$ (i.e., Definitions~\ref{def:trisection_agentpair_i_existing} and \ref{def:trisection_agentpair_J_existing}) are defined based on an auxiliary function $\Phi_i(x,y)$ in Equation~\eqref{equa:auxiliary_ixy_general} and its aggregation $\Phi_J(x,y)$ in Equation~\eqref{equa:auxiliary_Jxy_general}, respectively.
In the proposed model with two functions, we define a pair of alliance and conflict functions $\Phi_i^=(x,y)$ and $\Phi_i^{\asymp}(x,y)$ (i.e., Definition~\ref{def:alliance_conflict_function_ixy}) by splitting the auxiliary function $\Phi_i(x,y)$. 
Furthermore, with respect to an arbitrary subset of issues and the set of non-neutral issues, we respectively define two pairs of alliance and conflict functions in Equations \eqref{equa:alliance_conflict_function_Jxy} and \eqref{equa:alliance_conflict_function_J+-xy} to trisect the agent pairs (i.e., Equations \eqref{equa:trisection_agentpair_J_our} and \eqref{equa:trisection_agentpair_J+-_our}).
According to Theorem~\ref{theorem:trisection_agentpair_i_equivalent}, with respect to a single issue, the trisections of agent pairs with either a single auxiliary function or a pair of alliance and conflict functions are equivalent.

Secondly, for semantics, we define the alliance and conflict functions to reduce the difficulty of applying a rating function to trisect agent pairs. The proposed model with two functions and the existing model with a single function serve the same purpose.
They both divide agent relationships into three pair-wise disjoint parts: alliance, conflict, and neutrality relations.
\end{enumerate}

\subsection{An illustrative application of the proposed models}

We discuss a real-world application to illustrate the proposed three-way conflict analysis model with alliance and conflict functions. Specifically, we use the proposed model to help the government of Gansu province in China to address conflict problems. The same application is also used by Sun et al. in~\cite{Sun_2020} as a case study.

The three-valued situation table is given in Table~\ref{tab:situation_table_application}. $A=\{x_1,x_2,x_3,x_4,x_5,x_6,x_7,x_8,x_9,x_{10},x_{11}$, $x_{12},x_{13},x_{14}\}$ is a set of fourteen cities in Gansu province, namely, Lanzhou, Jinchang, Baiyin, Tianshui, Jiayuguan, Wuwei, Zhangye, Pingliang, Jiuquan, Qingyang, Dingxi, Longnan, Linxia, and Gannan. $I=\{i_1, i_2, i_3, i_4, i_5, i_6, i_7, i_8, i_9$, $i_{10},i_{11}\}$ is a set of issues involved in the development plans, namely, the construction of roads, factories, entertainment, educational institutions, the total population of residence, ecology environment, the number of senior intellectuals, the traffic capacity, mineral resources, sustainable development capacity, and water resources carrying capacity. 

\begin{table}[ht!]
\centering
\scalebox{1}{
\begin{tabular}{c|ccccccccccc}
\hline
\diagbox[width=3em,height=2em]{$A$}{$I$}
     &$i_1$&$i_2$&$i_3$&$i_4$&$i_5$&$i_6$&$i_7$&$i_8$&$i_9$&$i_{10}$&$i_{11}$\\
\hline
$x_1$&$+1$ &$-1$ &$0$  &$-1$ &$+1$  &$-1$ &$0$  &$-1$ &$+1$ &$-1$    &$+1$\\
$x_2$&$0$  &$+1$ &$-1$ &$0$  &$0$   &$+1$ &$-1$ &$0$  &$0$  &$+1$    &$-1$\\
$x_3$&$-1$ &$0$  &$-1$ &$-1$ &$-1$  &$+1$ &$+1$ &$-1$ &$-1$ &$0$     &$0$\\
$x_4$&$0$  &$0$  &$-1$ &$+1$ &$+1$  &$-1$ &$-1$ &$+1$ &$0$  &$-1$    &$-1$\\
$x_5$&$-1$ &$+1$ &$-1$ &$0$  &$-1$  &$+1$ &$0$  &$0$  &$-1$ &$+1$    &$+1$\\
$x_6$&$0$  &$+1$ &$0$  &$-1$ &$-1$  &$-1$ &$-1$ &$-1$ &$0$  &$+1$    &$-1$\\
$x_7$&$+1$ &$+1$ &$0$  &$+1$ &$0$   &$+1$ &$0$  &$+1$ &$+1$ &$+1$    &$0$\\
$x_8$&$-1$ &$0$ &$-1$  &$+1$ &$-1$  &$0$  &$+1$ &$+1$ &$-1$ &$0$     &$+1$\\
$x_9$&$+1$ &$+1$ &$0$  &$-1$ &$+1$  &$+1$ &$-1$ &$-1$ &$+1$ &$+1$    &$-1$\\
$x_{10}$&$-1$ &$-1$ &$-1$  &$0$&$+1$  &$-1$ &$+1$  &$0$ &$-1$&$-1$    &$+1$\\
$x_{11}$&$-1$ &$0$ &$-1$  &$-1$&$-1$  &$-1$ &$-1$  &$-1$ &$-1$&$0$    &$-1$\\
$x_{12}$&$0$ &$+1$ &$0$  &$-1$&$+1$  &$+1$ &$+1$  &$-1$ &$0$&$+1$    &$0$\\
$x_{13}$&$-1$ &$0$ &$-1$  &$+1$&$0$  &$0$ &$0$  &$+1$ &$-1$&$0$    &$+1$\\
$x_{14}$&$-1$ &$-1$ &$-1$  &$0$&$-1$  &$-1$ &$-1$  &$0$ &$-1$&$-1$    &$-1$\\
\hline
\end{tabular}}
\caption{A three-valued situation table~\cite{Sun_2020}}
\label{tab:situation_table_application}
\end{table}

\begin{table}[ht!]
\centering
\renewcommand\arraystretch{1.2}
\setlength{\tabcolsep}{1.5mm}
\scalebox{0.9}{
\begin{tabular}{c|c|cccccccccccccc}
\hline
\multicolumn{2}{c|}{\diagbox[width=6em,height=3em]{1st}{2nd}}& $x_1$&$x_2$&$x_3$&$x_4$&$x_5$&$x_6$&$x_7$&$x_8$&$x_9$&$x_{10}$&$x_{11}$&$x_{12}$&$x_{13}$&$x_{14}$\\ \hline
\multirow{2}{*}{$x_1$} &  $\Phi_{J^{+-}_{x_1}}^=$
&1  &$0$            &$\frac{2}{9}$  &$\frac{1}{3}$&$\frac{1}{9}$ &$\frac{1}{3}$&$\frac{2}{9}$&$\frac{1}{9}$ &$\frac{5}{9}$ &$\frac{5}{9}$ &$\frac{1}{3}$ &$\frac{1}{3}$ &$\frac{1}{9}$ &$\frac{1}{3}$\\
& $\Phi_{J^{+-}_{x_1}}^{\asymp}$
&0  &$\frac{4}{9}$  &$\frac{4}{9}$  &$\frac{1}{3}$&$\frac{2}{3}$ &$\frac{4}{9}$&$\frac{5}{9}$&$\frac{5}{9}$ &$\frac{4}{9}$ &$\frac{2}{9}$ &$\frac{4}{9}$ &$\frac{1}{3}$ &$\frac{4}{9}$ &$\frac{4}{9}$\\ \hline
\multirow{2}{*}{$x_2$}  &$\Phi_{J^{+-}_{x_2}}^=$
&0              &$1$   &$\frac{1}{3}$& $\frac{1}{2}$&$\frac{2}{3}$ &$\frac{2}{3}$&$\frac{1}{2}$&$\frac{1}{6}$&$\frac{5}{6}$&$\frac{1}{6}$&$\frac{1}{2}$&$\frac{1}{2}$&$\frac{1}{6}$&$\frac{1}{2}$\\
&$\Phi_{J^{+-}_{x_2}}^{\asymp}$
&$\frac{2}{3}$  &$0$   &$\frac{1}{6}$& $\frac{1}{3}$&$\frac{1}{6}$ &$\frac{1}{6}$&$0$          &$\frac{1}{3}$&$0$          &$\frac{5}{6}$&$\frac{1}{6}$&$\frac{1}{6}$&$\frac{1}{6}$&$\frac{1}{3}$\\ \hline
\multirow{2}{*}{$x_3$}  &$\Phi_{J^{+-}_{x_3}}^=$
&$\frac{1}{4}$ &$\frac{1}{4}$ &$1$& $\frac{1}{8}$ &$\frac{5}{8}$&$\frac{3}{8}$&$\frac{1}{8}$ &$\frac{5}{8}$ &$\frac{3}{8}$  &$\frac{1}{2}$&$\frac{3}{4}$&$\frac{1}{2}$&$\frac{3}{8}$&$\frac{1}{2}$\\
&$\Phi_{J^{+-}_{x_3}}^{\asymp}$
&$\frac{1}{2}$ &$\frac{1}{8}$ &$0$& $\frac{5}{8}$ &$0$          &$\frac{1}{4}$&$\frac{1}{2}$ &$\frac{1}{4}$ &$\frac{1}{2}$  &$\frac{1}{4}$&$\frac{1}{4}$&$\frac{1}{8}$&$\frac{1}{4}$&$\frac{1}{4}$\\
\hline
\multirow{2}{*}{$x_4$}  &$\Phi_{J^{+-}_{x_4}}^=$
&$\frac{3}{8}$ &$\frac{3}{8}$ &$\frac{1}{8}$& $1$ &$\frac{1}{8}$&$\frac{3}{8}$&$\frac{1}{4}$ &$\frac{3}{8}$  &$\frac{3}{8}$&$\frac{1}{2}$&$\frac{1}{2}$&$\frac{1}{8}$ &$\frac{3}{8}$&$\frac{5}{8}$\\
&$\Phi_{J^{+-}_{x_4}}^{\asymp}$
&$\frac{3}{8}$ &$\frac{1}{4}$ &$\frac{5}{8}$& $0$ &$\frac{1}{2}$&$\frac{1}{2}$ &$\frac{1}{4}$ &$\frac{3}{8}$  &$\frac{1}{2}$&$\frac{1}{4}$&$\frac{3}{8}$&$\frac{5}{8}$&$\frac{1}{8}$&$\frac{1}{8}$\\\hline
\multirow{2}{*}{$x_5$}   &$\Phi_{J^{+-}_{x_5}}^=$
&$\frac{1}{8}$ &$\frac{1}{2}$ &$\frac{5}{8}$& $\frac{1}{8}$ &$1$&$\frac{3}{8}$&$\frac{3}{8}$ &$\frac{5}{8}$ &$\frac{3}{8}$  &$\frac{1}{2}$&$\frac{1}{2}$&$\frac{3}{8}$&$\frac{1}{2}$&$\frac{1}{2}$\\
&$\Phi_{J^{+-}_{x_5}}^{\asymp}$
&$\frac{3}{4}$ &$\frac{1}{8}$ &$0$          & $\frac{1}{2}$ &$0$&$\frac{1}{4}$&$\frac{1}{4}$ &$0$           &$\frac{1}{2}$  &$\frac{1}{2}$&$\frac{1}{4}$&$\frac{1}{8}$&$0$          &$\frac{1}{2}$\\\hline
\multirow{2}{*}{$x_6$} &$\Phi_{J^{+-}_{x_6}}^=$
&$\frac{3}{8}$ &$\frac{1}{2}$ &$\frac{3}{8}$& $\frac{3}{8}$ &$\frac{3}{8}$&$1$&$\frac{1}{4}$ &$\frac{1}{8}$ &$\frac{3}{4}$  &$\frac{1}{8}$&$\frac{3}{4}$&$\frac{1}{2}$&$0$          &$\frac{1}{2}$\\
&$\Phi_{J^{+-}_{x_6}}^{\asymp}$ 
&$\frac{1}{2}$ &$\frac{1}{8}$ &$\frac{1}{4}$& $\frac{1}{2}$ &$\frac{1}{4}$&$0$&$\frac{3}{8}$ &$\frac{1}{2}$ &$\frac{1}{4}$  &$\frac{5}{8}$&$0$          &$\frac{3}{8}$&$\frac{3}{8}$&$\frac{1}{4}$\\\hline
\multirow{2}{*}{$x_7$}   &$\Phi_{J^{+-}_{x_7}}^=$
&$\frac{2}{7}$ &$\frac{3}{7}$ &$\frac{1}{7}$& $\frac{2}{7}$ &$\frac{3}{7}$&$\frac{2}{7}$&1&$\frac{2}{7}$ &$\frac{5}{7}$  &$0$          &$0$          &$\frac{3}{7}$&$\frac{2}{7}$&$0$          \\
&$\Phi_{J^{+-}_{x_7}}^{\asymp}$
&$\frac{5}{7}$ &$0$           &$\frac{4}{7}$& $\frac{2}{7}$ &$\frac{2}{7}$&$\frac{3}{7}$&0&$\frac{2}{7}$ &$\frac{2}{7}$  &$\frac{5}{7}$&$\frac{5}{7}$&$\frac{2}{7}$&$\frac{2}{7}$&$\frac{5}{7}$\\\hline
\multirow{2}{*}{$x_8$} &$\Phi_{J^{+-}_{x_8}}^=$
&$\frac{1}{8}$ &$\frac{1}{8}$ &$\frac{5}{8}$& $\frac{3}{8}$ &$\frac{5}{8}$&$\frac{1}{8}$&$\frac{1}{4}$ &$1$ &$0$            &$\frac{5}{8}$&$\frac{1}{2}$&$\frac{1}{8}$&$\frac{3}{4}$&$\frac{1}{2}$\\
&$\Phi_{J^{+-}_{x_8}}^{\asymp}$
&$\frac{5}{8}$ &$\frac{1}{4}$ &$\frac{1}{4}$& $\frac{3}{8}$ &$0$          &$\frac{1}{2}$&$\frac{1}{4}$ &$0$ &$\frac{7}{8}$  &$\frac{1}{8}$&$\frac{1}{2}$&$\frac{3}{8}$&$0$          &$\frac{1}{4}$\\\hline
\multirow{2}{*}{$x_9$}   &$\Phi_{J^{+-}_{x_9}}^=$
&$\frac{1}{2}$ &$\frac{1}{2}$ &$\frac{3}{10}$& $\frac{3}{10}$ &$\frac{3}{10}$&$\frac{3}{5}$&$\frac{1}{2}$ &$0$          &1&$\frac{1}{10}$&$\frac{2}{5}$&$\frac{3}{5}$&$0$           &$\frac{1}{5}$\\
&$\Phi_{J^{+-}_{x_9}}^{\asymp}$
&$\frac{2}{5}$ &$0$           &$\frac{2}{5}$& $\frac{2}{5}$ &$\frac{2}{5}$&$\frac{1}{5}$&$\frac{1}{5}$ &$\frac{7}{10}$&0&$\frac{7}{10}$&$\frac{2}{5}$&$\frac{1}{10}$&$\frac{1}{2}$&$\frac{3}{5}$\\\hline
\multirow{2}{*}{$x_{10}$}&$\Phi_{J^{+-}_{x_{10}}}^=$
&$\frac{5}{9}$ &$\frac{1}{9}$ &$\frac{4}{9}$& $\frac{4}{9}$ &$\frac{4}{9}$&$\frac{1}{9}$&$0$          &$\frac{5}{9}$ &$\frac{1}{9}$  &1&$\frac{4}{9}$&$\frac{2}{9}$&$\frac{4}{9}$&$\frac{2}{3}$\\
&$\Phi_{J^{+-}_{x_{10}}}^{\asymp}$
&$\frac{2}{9}$ &$\frac{5}{9}$ &$\frac{2}{9}$& $\frac{2}{9}$ &$\frac{4}{9}$&$\frac{5}{9}$&$\frac{5}{9}$&$\frac{1}{9}$ &$\frac{7}{9}$  &0&$\frac{1}{3}$&$\frac{1}{3}$&$0$          &$\frac{1}{3}$\\
\hline
\multirow{2}{*}{$x_{11}$}&$\Phi_{J^{+-}_{x_{11}}}^=$
&$\frac{1}{3}$ &$\frac{1}{3}$ &$\frac{2}{3}$& $\frac{4}{9}$ &$\frac{4}{9}$&$\frac{2}{3}$&$0$          &$\frac{4}{9}$ &$\frac{4}{9}$  &$\frac{4}{9}$&1&$\frac{2}{9}$&$\frac{1}{3}$&$\frac{7}{9}$\\
&$\Phi_{J^{+-}_{x_{11}}}^{\asymp}$
&$\frac{4}{9}$ &$\frac{1}{9}$ &$\frac{2}{9}$& $\frac{1}{3}$ &$\frac{2}{9}$&$0$          &$\frac{5}{9}$&$\frac{4}{9}$ &$\frac{4}{9}$  &$\frac{1}{3}$&0&$\frac{1}{3}$&$\frac{1}{3}$&$0$\\
\hline
\multirow{2}{*}{$x_{12}$}&$\Phi_{J^{+-}_{x_{12}}}^{=}$
&$\frac{3}{7}$ &$\frac{3}{7}$ &$\frac{4}{7}$& $\frac{1}{7}$ &$\frac{3}{7}$&$\frac{4}{7}$&$\frac{3}{7}$ &$\frac{1}{7}$ &$\frac{6}{7}$  &$\frac{2}{7}$&$\frac{2}{7}$&1&$0$          &$0$          \\
&$\Phi_{J^{+-}_{x_{12}}}^{\asymp}$
&$\frac{3}{7}$ &$\frac{1}{7}$ &$\frac{1}{7}$& $\frac{5}{7}$ &$\frac{1}{7}$&$\frac{3}{7}$&$\frac{2}{7}$ &$\frac{3}{7}$ &$\frac{1}{7}$  &$\frac{3}{7}$&$\frac{3}{7}$&0&$\frac{2}{7}$&$\frac{5}{7}$\\\hline
\multirow{2}{*}{$x_{13}$}&$\Phi_{J^{+-}_{x_{13}}}^=$
&$\frac{1}{6}$  &$\frac{1}{6}$   &$\frac{1}{2}$& $\frac{1}{2}$   &$\frac{2}{3}$&$0$          &$\frac{1}{3}$&1   &$0$          &$\frac{2}{3}$   &$\frac{1}{2}$&$0$             &1&$\frac{1}{2}$\\ 
&$\Phi_{J^{+-}_{x_{13}}}^{\asymp}$
&$\frac{2}{3}$  &$\frac{1}{6}$   &$\frac{1}{3}$& $\frac{1}{6}$   &$0$          &$\frac{1}{2}$&$\frac{1}{3}$&0   &$\frac{5}{6}$&$0$             &$\frac{1}{2}$&$\frac{1}{3}$   &0&$\frac{1}{6}$\\ \hline
\multirow{2}{*}{$x_{14}$}&$\Phi_{J^{+-}_{x_{14}}}^{=}$
&$\frac{1}{3}$ &$\frac{1}{3}$ &$\frac{4}{9}$& $\frac{5}{9}$ &$\frac{4}{9}$&$\frac{4}{9}$&$0$          &$\frac{4}{9}$ &$\frac{2}{9}$  &$\frac{2}{3}$&$\frac{7}{9}$&$0$          &$\frac{1}{3}$&1\\
&$\Phi_{J^{+-}_{x_{14}}}^{\asymp}$
&$\frac{4}{9}$ &$\frac{1}{3}$ &$\frac{2}{9}$& $\frac{1}{9}$ &$\frac{4}{9}$&$\frac{2}{9}$&$\frac{5}{9}$ &$\frac{2}{9}$ &$\frac{2}{3}$  &$\frac{1}{3}$&$0$          &$\frac{5}{9}$&$\frac{1}{9}$&0\\
\hline
\end{tabular}}
\caption{The alliance and conflict degrees with respect to the non-neutral issues}
\label{tab:alliance_conflict_application}
\end{table}

We consider the non-neutral issues and compute the alliance and conflict degrees between cities according to Equation~\eqref{equa:alliance_conflict_function_J+-xy_PY}, which is given in Table~\ref{tab:alliance_conflict_application}. Then by applying a pair of thresholds $(\frac{1}{|J^{+-}_x|},\frac{1}{4})$ in Definition~\ref{def:trisection_agentpair_J_our}, we can trisect the agent pairs and get the alliance sets as follows:
\begin{alignat}{5}
&AS_{J^{+-}}(x_1)=\{x_1\},&\qquad\qquad&AS_{J^{+-}}(x_2)=\{x_2,x_3,x_5,x_6,x_7,x_9,x_{11},x_{12}\},\nonumber\\
&AS_{J^{+-}}(x_3)=\{x_2,x_3,x_5,x_{12}\},&&AS_{J^{+-}}(x_4)=\{x_4,x_{13},x_{14}\},\nonumber\\
&AS_{J^{+-}}(x_5)=\{x_2,x_3,x_5,x_8,x_{12},x_{13}\},&&AS_{J^{+-}}(x_6)=\{x_2,x_6,x_{11}\},\nonumber\\
&AS_{J^{+-}}(x_7)=\{x_2,x_7\},&&AS_{J^{+-}}(x_8)=\{x_5,x_8,x_{10},x_{13}\},\nonumber\\
&AS_{J^{+-}}(x_9)=\{x_2,x_9,x_{12}\},&&AS_{J^{+-}}(x_{10})=\{x_8,x_{10},x_{13}\},\nonumber\\
&AS_{J^{+-}}(x_{11})=\{x_2,x_6,x_{11},x_{14}\},&&AS_{J^{+-}}(x_{12})=\{x_2,x_3,x_5,x_9,x_{12}\},\nonumber\\
&AS_{J^{+-}}(x_{13})=\{x_4,x_5,x_8,x_{10},x_{13},x_{14}\},&&AS_{J^{+-}}(x_{14})=\{x_4,x_{11},x_{13},x_{14}\}.
\end{alignat}

Accordingly, Lanzhou (i.e., $x_1$), the capital city of Gansu, only allies with itself. Jinchang (i.e., $x_2$) has the biggest alliance set. We consider the non-neutral issues again to formally represent the decisions of the alliance sets. For example, the decisions of the alliance sets regarding Lanzhou and Jinchang are:
\begin{eqnarray}
{\rm Des}(AS_{J^{+-}}(x_1))&=&\langle i_1,+1\rangle\wedge\langle i_2,-1\rangle\wedge\langle i_4,-1\rangle\wedge\langle i_5,+1\rangle\wedge\langle i_6,-1\rangle\wedge\langle i_8,-1\rangle\wedge\langle i_9,+1\rangle\wedge\langle i_{10},-1\rangle\wedge\langle i_{11},+1\rangle,\nonumber\\
{\rm Des}(AS_{J^{+-}}(x_2))&=&\langle i_2,+1\rangle\wedge\langle i_3,-1\rangle\wedge\langle i_6,+1\rangle\wedge\langle i_7,-1\rangle\wedge\langle i_{10},+1\rangle\wedge\langle i_{11},-1\rangle.
\end{eqnarray}

Furthermore, we compute the maximal consistent alliance sets based on Definition~\ref{def:maximal_consistent_alliance_set} as:
\begin{alignat}{5}
&X_1=\{x_1\},&\qquad\qquad&X_2=\{x_2,x_3,x_5,x_{12}\},&\qquad\qquad&X_3=\{x_2,x_{6},x_{11}\}, \nonumber\\
&X_4=\{x_2,x_7\}, && X_5=\{x_2,x_9,x_{12}\}, && X_6=\{x_4,x_{13},x_{14}\}, \nonumber\\
&X_7=\{x_5,x_8,x_{13}\}, && X_8=\{x_8,x_{10},x_{13}\}, && X_9=\{x_{11},x_{14}\}.
\end{alignat}
The cities in a maximal consistent alliance set are allied with each other. Recall that we need to trisect a set of issues $J$ in order to formally represent the decision of a maximal consistent alliance set. Let us consider $J=I$. By applying the pairs of thresholds $(0,\frac{1}{|X_k|})$ $(k=1,2,\cdots,10)$ in Definition~\ref{def:trisection_issue_X_our}, we trisect $J$ regarding each maximal consistent alliance set as follows:
\begin{alignat}{5}
&J_{X_1}^{+}(\Phi^{=},\Phi^{\asymp})=\{i_1,i_5,i_9,i_{11}\}, &~~&J_{X_1}^{-}(\Phi^{=},\Phi^{\asymp})=\{i_2,i_4,i_6,i_8,i_{10}\}, &~~&J_{X_1}^{0}(\Phi^{=},\Phi^{\asymp})=\{i_3,i_7\};\nonumber\\
&J_{X_2}^{+}(\Phi^{=},\Phi^{\asymp})=\{i_2,i_6,i_{10}\},&& J_{X_2}^{-}(\Phi^{=},\Phi^{\asymp})=\{i_1,i_3,i_4,i_8,i_9\},&&J_{X_2}^{0}(\Phi^{=},\Phi^{\asymp})=\{i_5,i_7,i_{11}\};\nonumber\\
&J_{X_3}^{+}(\Phi^{=},\Phi^{\asymp})=\{i_2,i_{10}\}, &&J_{X_3}^{-}(\Phi^{=},\Phi^{\asymp})=\{i_1,i_3,i_4,i_5,i_7,i_8,i_9,i_{11}\}, &&J_{X_3}^{0}(\Phi^{=},\Phi^{\asymp})=\{i_6\};\nonumber\\
&J_{X_4}^{+}(\Phi^{=},\Phi^{\asymp})=\{i_1,i_2,i_4,i_6,i_8,i_9,i_{10}\},&& J_{X_4}^{-}(\Phi^{=},\Phi^{\asymp})=\{i_3,i_7,i_{11}\}, &&J_{X_4}^{0}(\Phi^{=},\Phi^{\asymp})=\{i_5\};\nonumber\\
&J_{X_5}^{+}(\Phi^{=},\Phi^{\asymp})=\{i_1,i_2,i_5,i_6,i_9,i_{10}\},&& J_{X_5}^{-}(\Phi^{=},\Phi^{\asymp})=\{i_3,i_4,i_8,i_{11}\}, &&J_{X_5}^{0}(\Phi^{=},\Phi^{\asymp})=\{i_7\};\nonumber\\
&J_{X_6}^{+}(\Phi^{=},\Phi^{\asymp})=\{i_4,i_8\},&& J_{X_6}^{-}(\Phi^{=},\Phi^{\asymp})=\{i_1,i_2,i_3,i_6,i_7,i_9,i_{10}\},&&J_{X_6}^{0}(\Phi^{=},\Phi^{\asymp})=\{i_5,i_{11}\};\nonumber\\
&J_{X_7}^{+}(\Phi^{=},\Phi^{\asymp})=\{i_2,i_4,i_6,i_7,i_8,i_{10},i_{11}\}, &&J_{X_7}^{-}(\Phi^{=},\Phi^{\asymp})=\{i_1,i_3,i_5,i_9\}, &&J_{X_7}^{0}(\Phi^{=},\Phi^{\asymp})=\emptyset;\nonumber\\
&J_{X_8}^{+}(\Phi^{=},\Phi^{\asymp})=\{i_4,i_7,i_8,i_{11}\},&& J_{X_8}^{-}(\Phi^{=},\Phi^{\asymp})=\{i_1,i_2,i_3,i_6,i_9,i_{10}\}, &&J_{X_8}^{0}(\Phi^{=},\Phi^{\asymp})=\{i_5\};\nonumber\\
&J_{X_9}^{+}(\Phi^{=},\Phi^{\asymp})=\emptyset,&& J_{X_9}^{-}(\Phi^{=},\Phi^{\asymp})=\{i_1,i_2,i_3,i_4,i_5,i_6,i_7,i_8,i_9,i_{10},i_{11}\}, &&J_{X_9}^{0}(\Phi^{=},\Phi^{\asymp})=\emptyset.
\end{alignat}
Accordingly, the decisions of maximal consistent alliance sets with respect to the non-neutral issues are given as follows:
{\small
\begin{eqnarray}
{\rm Des}_{J^{+-}}(X_1)&=&\langle i_1,+1\rangle\wedge\langle i_2,-1\rangle\wedge\langle i_4,-1\rangle\wedge\langle i_5,+1\rangle\wedge\langle i_6,-1\rangle\wedge\langle i_8,-1\rangle\wedge\langle i_9,+1\rangle\wedge\langle i_{10},-1\rangle\wedge\langle i_{11},+1\rangle,\nonumber\\
{\rm Des}_{J^{+-}}(X_2)&=&\langle i_1,-1\rangle\wedge\langle i_2,+1\rangle\wedge\langle i_3,-1\rangle\wedge\langle i_4,-1\rangle\wedge\langle i_6,+1\rangle\wedge\langle i_8,-1\rangle\wedge\langle i_9,-1\rangle\wedge\langle i_{10},+1\rangle,\nonumber\\
{\rm Des}_{J^{+-}}(X_3)&=&\langle i_1,-1\rangle\wedge\langle i_2,+1\rangle\wedge\langle i_3,-1\rangle\wedge\langle i_4,-1\rangle\wedge\langle i_5,-1\rangle\wedge\langle i_7,-1\rangle\wedge\langle i_8,-1\rangle\wedge\langle i_9,-1\rangle\wedge\langle i_{10},+1\rangle\wedge\langle i_{11},-1\rangle,\nonumber\\
{\rm Des}_{J^{+-}}(X_4)&=&\langle i_1,+1\rangle\wedge\langle i_2,+1\rangle\wedge\langle i_3,-1\rangle\wedge\langle i_4,+1\rangle\wedge\langle i_6,+1\rangle\wedge\langle i_7,-1\rangle\wedge\langle i_8,+1\rangle\wedge\langle i_9,+1\rangle\wedge\langle i_{10},+1\rangle\wedge\langle i_{11},-1\rangle,\nonumber\\
{\rm Des}_{J^{+-}}(X_5)&=&\langle i_1,+1\rangle\wedge\langle i_2,+1\rangle\wedge\langle i_3,-1\rangle\wedge\langle i_4,-1\rangle\wedge\langle i_5,+1\rangle\wedge\langle i_6,+1\rangle\wedge\langle i_8,-1\rangle\wedge\langle i_9,+1\rangle\wedge\langle i_{10},+1\rangle\wedge\langle i_{11},-1\rangle,\nonumber\\
{\rm Des}_{J^{+-}}(X_6)&=&\langle i_1,-1\rangle\wedge\langle i_2,-1\rangle\wedge\langle i_3,-1\rangle\wedge\langle i_4,+1\rangle\wedge\langle i_6,-1\rangle\wedge\langle i_7,-1\rangle\wedge\langle i_8,+1\rangle\wedge\langle i_9,-1\rangle\wedge\langle i_{10},-1\rangle,\nonumber\\
{\rm Des}_{J^{+-}}(X_7)&=&\langle i_1,-1\rangle\wedge\langle i_2,+1\rangle\wedge\langle i_3,-1\rangle\wedge\langle i_4,+1\rangle\wedge\langle i_5,-1\rangle\wedge\langle i_6,+1\rangle\wedge\langle i_7,+1\rangle\wedge\langle i_8,+1\rangle\wedge\langle i_9,-1\rangle\wedge\langle i_{10},+1\rangle\wedge\langle i_{11},+1\rangle,\nonumber\\
{\rm Des}_{J^{+-}}(X_8)&=&\langle i_1,-1\rangle\wedge\langle i_2,-1\rangle\wedge\langle i_3,-1\rangle\wedge\langle i_4,+1\rangle\wedge\langle i_6,-1\rangle\wedge\langle i_7,+1\rangle\wedge\langle i_8,+1\rangle\wedge\langle i_9,-1\rangle\wedge\langle i_{10},-1\rangle\wedge\langle i_{11},+1\rangle,\nonumber\\
{\rm Des}_{J^{+-}}(X_9)&=&\langle i_1,-1\rangle\wedge\langle i_2,-1\rangle\wedge\langle i_3,-1\rangle\wedge\langle i_4,-1\rangle\wedge\langle i_5,-1\rangle\wedge\langle i_6,-1\rangle\wedge\langle i_7,-1\rangle\wedge\langle i_8,-1\rangle\wedge\langle i_9,-1\rangle\wedge\langle i_{10},-1\rangle\wedge\langle i_{11},-1\rangle\nonumber\\
&=& \bigwedge_{i \in I} \langle i,-1\rangle.
\end{eqnarray}}

We may consider the decisions of some alliance sets as the candidates of the final strategy. According to Equation~\eqref{equa:cardinality_strategy}, the number of all non-neutral strategies is $3^{|I|}=3^{11}=177147$, which is a huge number. Thus, instead of exploring each non-neutral strategy, we may develop certain heuristics for selecting the candidates. For instance, the government may prefer the strategy induced from the largest alliance set as a candidate strategy, which is $x_2$ Jinchang in this example. The strategy of $x_2$ with respect to the non-neutral issues is:
\begin{equation}
\widetilde{\mathcal{S}}_{\rm J}={\rm Des}_{J^{+-}}(x_2)=\langle i_2,+1\rangle\wedge\langle i_3,-1\rangle\wedge\langle i_6,+1\rangle\wedge\langle i_7,-1\rangle\wedge\langle i_{10},+1\rangle\wedge\langle i_{11},-1\rangle.
\end{equation}
By applying the thresholds $l_s=\frac{1}{7}$, $h_s=\frac{1}{6}$, $l_o=\frac{1}{6}$, and $h_o=\frac{1}{3}$ in Equation~\eqref{equa:trisection_agent_S+-_our}, we get the trisection of $A$ with respect to $\widetilde{\mathcal{S}}_{\rm J}$ as:
\begin{eqnarray}
X_{J_{\widetilde{\mathcal{S}}_{\rm J}}}^+(\Phi^=,\Phi^{\asymp})&=&
\{x\in X\mid \Phi_{J_{\widetilde{\mathcal{S}}_{\rm J}}}^{=P}(\widetilde{\mathcal{S}}_{\rm J},x)\geq \frac{1}{6} \wedge \Phi_{J_{\widetilde{\mathcal{S}}_{\rm J}}}^{\asymp P}(\widetilde{\mathcal{S}}_{\rm J},x)\leq \frac{1}{6}\},\nonumber\\
&=&\{x_2,x_3,x_5,x_6,x_7,x_9,x_{11},x_{12},x_{13}\},\nonumber\\
X_{J_{\widetilde{\mathcal{S}}_{\rm J}}}^-(\Phi^=,\Phi^{\asymp})&=&
\{x\in X\mid \Phi_{J_{\widetilde{\mathcal{S}}_{\rm J}}}^{\asymp P}(\widetilde{\mathcal{S}}_{\rm J},x)\geq \frac{1}{3}\wedge \Phi_{J_{\widetilde{\mathcal{S}}_{\rm J}}}^{=P}(\widetilde{\mathcal{S}}_{\rm J},x)\leq \frac{1}{7}\},\nonumber\\
&=&\{x_1\},\nonumber\\
X_{J_{\widetilde{\mathcal{S}}_{\rm J}}}^0(\Phi^=,\Phi^{\asymp})&=&(X_{J_{\widetilde{\mathcal{S}}_{\rm J}}}^+(\Phi^=,\Phi^{\asymp})\cup X_{J_{\widetilde{\mathcal{S}}_{\rm J}}}^-(\Phi^=,\Phi^{\asymp}))^c\nonumber\\
&=&\{x_4,x_8,x_{10},x_{14}\}.
\end{eqnarray}

In another case, the government may prefer the city with the lowest GDP for the sake of the overall development of the province. In this example, such a heuristic will select the strategy of $x_{14}$ Gannan as a candidate. The strategy of $x_{14}$ with respect to the non-neutral issues is:
\begin{equation}
\widetilde{\mathcal{S}}_{\rm G}={\rm Des}_{J^{+-}}(x_{14})=\langle {i_i},-1\rangle\wedge\langle {i_2},-1\rangle\wedge\langle {i_3},-1\rangle\wedge\langle {i_5},-1\rangle\wedge\langle {i_6},-1\rangle\wedge\langle {i_7},-1\rangle\wedge\langle {i_9},-1\rangle\wedge\langle {i_{10}},-1\rangle\wedge\langle {i_{11}},-1\rangle.
\end{equation} 
By applying the thresholds $l_s=\frac{2}{9}$, $h_s=\frac{1}{3}$, $l_o=\frac{1}{9}$, $h_o=\frac{4}{9}$ in Equation~\eqref{equa:trisection_agent_S+-_our}, we get the trisection of $A$ with respect to $\widetilde{\mathcal{S}}_{\rm G}$ as:
\begin{eqnarray}
X_{J_{\widetilde{\mathcal{S}}_{\rm G}}}^+(\Phi^=,\Phi^{\asymp})&=&
\{x\in X\mid \Phi_{J_{\widetilde{\mathcal{S}}_{\rm G}}}^{=P}(\widetilde{\mathcal{S}}_{\rm G},x)\geq \frac{1}{3} \wedge \Phi_{J_{\widetilde{\mathcal{S}}_{\rm G}}}^{\asymp P}(\widetilde{\mathcal{S}}_{\rm G},x)\leq \frac{1}{9}\},\nonumber\\
&=&\{x_4,x_{11},x_{13},x_{14}\},\nonumber\\
X_{J_{\widetilde{\mathcal{S}}_{\rm G}}}^-(\Phi^=,\Phi^{\asymp})&=&
\{x\in X\mid \Phi_{J_{\widetilde{\mathcal{S}}_{\rm G}}}^{\asymp P}(\widetilde{\mathcal{S}}_{\rm G},x)\geq \frac{4}{9}\wedge \Phi_{J_{\widetilde{\mathcal{S}}_{\rm G}}}^{=P}(\widetilde{\mathcal{S}}_{\rm G},x)\leq \frac{2}{9}\},\nonumber\\
&=&\{x_7,x_9,x_{12}\},\nonumber\\
X_{J_{\widetilde{\mathcal{S}}_{\rm G}}}^0(\Phi^=,\Phi^{\asymp})&=&(X_{J_{\widetilde{\mathcal{S}}_{\rm G}}}^+(\Phi^=,\Phi^{\asymp})\cup X_{J_{\widetilde{\mathcal{S}}_{\rm G}}}^-(\Phi^=,\Phi^{\asymp}))^c\nonumber\\
&=&\{x_1,x_2,x_3,x_5,x_6,x_8,x_{10}\}.
\end{eqnarray}

After determining the final strategy, the government may make efforts to persuade the cities in the neutral set $X_{J_{\mathcal{S}}}^0(\Phi^=,\Phi^{\asymp})$ in order to implement the strategy. 

\section{Conclusions}
\label{sec:conclusion}

This paper proposes the alliance and conflict functions by separating the two aspects of alliance and conflict measured in an auxiliary function. Using the alliance and conflict functions as two evaluation functions, we study two topics of three-way conflict analysis. The first topic is about the relationship between agents. Particularly, we define the alliance sets and the maximal consistent alliance sets. Furthermore, we present the conjunction of issue-rating pairs as a formal representation of the decision or description of an alliance set. In the second topic, we formally define the strategies through issue-rating pairs and investigate their relationships with agents. To verify the effectiveness of the trisections with the proposed alliance and conflict functions, we compare them with the trisections in existing models based on a single evaluation function. Finally, we apply the proposed model in a real-world application to help the government of Gansu province make the development plan.

Determining the thresholds used in the trisections is a fundamental issue that needs further discussion in future work. Moreover, our approaches can be generalized with respect to a few aspects, such as weighted agents, conflict sets of agents instead of alliance sets, incomplete situation tables, and dynamic situation tables. In addition, it is worth investigating the connections and differences between three-way conflict analysis and the concept of bipolarity and shadowed sets, which may inspire exciting results in these closely related topics.

\section*{Acknowledgment}
We would like to thank the anonymous reviewers for their professional comments and valuable suggestions. This work was supported in part by the National Natural Science Foundation of China (Nos. 62076040, 61976130), Ministry of Education in China Project of Humanities and Social Sciences (No.21YJC630092), Hunan Provincial Natural Science Foundation of China (No. 2020JJ3034), and the Scientific Research Fund of Hunan Provincial
Education Department (No. 19B027).


\begin{thebibliography}{10}
\expandafter\ifx\csname url\endcsname\relax
  \def\url#1{\texttt{#1}}\fi
\expandafter\ifx\csname urlprefix\endcsname\relax\def\urlprefix{URL }\fi
\expandafter\ifx\csname href\endcsname\relax
  \def\href#1#2{#2} \def\path#1{#1}\fi

\bibitem{Deja_2002}
R. Deja,
\href{https://doi.org/10.1002/int.10019}{Conflict Analysis}, 
International Journal of Intelligent Systems, 17 (2002) 235-253.

\bibitem{Deng_2014}
X. Deng, Y. Yao, J. Yao,
\href{https://doi.org/10.1007/978-3-319-08326-1_8}{On interpreting three-way decisions through two-way decisions},
in: T. Andreasen et al. (Eds.) Foundations of Intelligent Systems (ISMIS 2014), LNCS 8502, Springer, Cham, 2014, pp. 73-82.

\bibitem{Dubois_2008}
D. Dubois, H. Prade,
\href{https://doi.org/10.1002/int.20297}{An introduction to bipolar representations of information and preference},
International Journal of Intelligent Systems, 23 (2008) 866-877.

\bibitem{Dubois_2012}
D. Dubois, H. Prade,
\href{https://doi.org/10.1016/j.fss.2010.11.007}{Gradualness, uncertainty and bipolarity: making sense of fuzzy sets}, 
Fuzzy sets and Systems, 192 (2012) 3-24.

\bibitem{Fan_2018}
Y. Fan, J. Qi, L. Wei, 
\href{https://doi.org/10.1007/978-3-319-99368-3_41}{A conflict analysis model based on three-way decisions}, 
in: H. Nguyen et al. (Eds.) Rough Sets (IJCRS 2018), LNCS 11103, Springer, Cham, 2018, pp. 522-532.

\bibitem{Hu_2014}
B. Hu,
\href{https://doi.org/10.1016/j.ins.2014.05.015}{Three-way decisions space and three-way decisions},
Information Sciences, 281 (2014) 21-52.

\bibitem{Hu_2021}
M. Hu,
\href{https://doi.org/10.1016/j.ins.2021.05.058}{Three-way data analytics: Preparing and analyzing data in threes},
Information Sciences, 573 (2021) 412-432.

\bibitem{Hu_2019}
M. Hu, Y. Yao,
\href{https://doi.org/10.1016/j.knosys.2018.11.022}{Structured approximations as a basis for three-way decisions in rough set theory},
Knowledge-Based Systems, 165 (2019) 92-109.

\bibitem{Jabbour_2017}
S. Jabbour, Y. Ma, B. Raddaoui, L. Sais,
\href{https://doi.org/10.1016/j.ijar.2016.12.017}{Quantifying conflicts in propositional logic through prime implicates},
International Journal of Approximate Reasoning, 89 (2017) 27-40.

\bibitem{Jiang_2011}
Y. Jiang, Y. Tang, Q. Chen, Z. Cao,
\href{https://doi.org/10.1016/j.camwa.2011.06.036}{Semantic operations of multiple soft sets under conflict}, 
Computers and Mathematics with Applications, 62 (2011) 1923-1939.

\bibitem{Lang_2020_general}
G. Lang,
\href{https://doi.org/10.1007/s13042-020-01100-y}{A general conflict analysis model based on three-way decision},
International Journal of Machine Learning and Cybernetics, 11 (2020) 1083–1094.

\bibitem{Lang_2020_unification}
G. Lang, J. Luo, Y. Yao,
\href{https://doi.org/10.1016/j.knosys.2020.105556}{Three-way conflict analysis: a unification of models based on rough sets and formal concept analysis},
Knowledge-Based Systems, 194 (2020) 105556.
 
\bibitem{Lang_2017}
G. Lang, D. Miao, M. Cai,
\href{https://doi.org/10.1016/j.ins.2017.04.030}{Three-way decision approaches to conflict analysis using decision-theoretic rough set theory},
Information Sciences, 406-407 (2017) 185-207.

\bibitem{Lang_2019}
G. Lang, D. Miao, H. Fujita,
\href{http: //dx.doi.org/10.1109/TFUZZ.2019.2908123}{Three-way group conflict analysis based on Pythagorean fuzzy set theory}, 
IEEE Transactions on Fuzzy Systems, 28 (2019) 447-461.

\bibitem{Leung_2003}
Y. Leung, D. Li,
\href{https://doi.org/10.1016/s0020-0255(03)00061-6}{Maximal consistent block technique for rule acquisition in incomplete information systems},
Information Sciences, 153 (2003) 85-106.

\bibitem{Li_2021}
X. Li, X. Wang, G. Lang, H. Yi,
\href{https://doi.org/10.1016/j.ijar.2020.12.004}{Conflict analysis based on three-way decision for triangular fuzzy information systems}, International Journal of Approximate Reasoning, 132 (2021) 88-106.

\bibitem{Liu_2020}
J. Liu, H. Li, B. Huang, Y. Liu, D. Liu, 
\href{https://doi.org/10.1016/j.ins.2021.06.018}{Convex combination-based consensus analysis for intuitionistic fuzzy three-way group decision},
Information Sciences, 574 (2021) 542-566.

\bibitem{Liu_2015}
Y. Liu, Y. Lin,
\href{https://doi.org/10.1016/j.asoc.2015.02.045}{Intuitionistic fuzzy rough set model based on conflict distance and applications}, 
Applied Soft Computing, 31 (2015) 266-273.

\bibitem{Luo_2020}
J. Luo, M. Hu, K. Qin,
\href{https://doi.org/10.1016/j.ijar.2020.02.005}{Three-way decision with incomplete information based on similarity and satisfiability},
International Journal of Approximate Reasoning, 120 (2020) 151-183.

\bibitem{Maldonado_2020}
S. Maldonado, G. Peters, R. Weber,
\href{https://doi.org/10.1016/j.ins.2018.08.001}{Credit scoring using three-way decisions with probabilistic rough sets},
Information Sciences, 507 (2020) 700-714.

\bibitem{Pawlak_1984}
Z. Pawlak, 
\href{https://doi.org/10.1016/S0020-7373(84)80062-0}{On conflicts},
International Journal of Man-Machine Studies, 21 (1984) 127-134.

\bibitem{Pawlak_1998}
Z. Pawlak, 
\href{https://doi.org/10.1016/s0020-0255(97)10072-x}{An inquiry into anatomy of conflicts}, 
Information Sciences, 109 (1998) 65-78.

\bibitem{Pawlak_2005}
Z. Pawlak, 
\href{https://doi.org/10.1016/j.ejor.2003.09.038}{Some remarks on conflict analysis}, 
European Journal of Operational Research, 166 (2005) 649-654.

\bibitem{Pedrycz_1998}
W. Pedrycz,
\href{https://doi.org/10.1109/3477.658584}{Shadowed sets: representing and processing fuzzy sets},
IEEE Transactions on Systems, Man, and Cybernetics, Part B (Cybernetics), 28 (1998) 103-109.

\bibitem{Pedrycz_2005}
W. Pedrycz,
\href{https://doi.org/10.1016/j.patrec.2005.05.001}{Interpretation of clusters in the framework of shadowed sets},
Pattern recognition letters, 26 (2005) 2439-2449.

\bibitem{Pedrycz_2009}
W. Pedrycz,
\href{https://doi.org/10.1002/int.20323}{From fuzzy sets to shadowed sets: interpretation and computing},
International journal of intelligent systems, 24 (2009) 48-61.

\bibitem{Silva_2016}
L. Silva, A. Almeida-Filho,
\href{https://doi.org/10.1016/j.ins.2016.01.080}{A multicriteria approach for analysis of conflicts in evidence theory},
Information Sciences, 346-347 (2016) 275-285.

\bibitem{Skowron_2002}
A. Skowron, R. Deja,
\href{https://citeseerx.ist.psu.edu/viewdoc/download?doi=10.1.1.107.5063&rep=rep1&type=pdf}{On some conflict models and conflict resolutions},
Romanian Journal of Information Science and Technology, 5 (2002) 69-82.

\bibitem{Skowron_2006}
A. Skowron, S. Ramanna, J. Peters,
\href{https://doi.org/10.1007/11795131_34}{Conflict analysis and information systems: a rough set approach},
Rough Sets and Current Trends in Computing, 4062 (2006) 233-240.

\bibitem{Sun_2020}
B. Sun, X. Chen, L. Zhang, W. Ma,
\href{https://doi.org/10.1016/j.ins.2019.05.080}{Three-way decision making approach to conflict analysis and resolution using probabilistic rough set over two universes},
Information Sciences, 507 (2020) 809-822.

\bibitem{Sun_2015}
B. Sun, W. Ma,
\href{https://doi.org/10.1016/j.ins.2015.03.061}{Rough approximation of a preference relation by multi-decision dominance for a multi-agent conflict analysis problem},
Information Sciences, 315 (2015) 39-53.

\bibitem{Sun_2016}
B. Sun, W. Ma, H. Zhao,
\href{https://doi.org/10.1016/j.ins.2016.08.030}{Rough set-based conflict analysis model and method over two universes}, 
Information Sciences, 372 (2016) 111-125.

\bibitem{Xu_2020}
J. Xu, Y. Zhang, D. Miao,
\href{https://doi.org/10.1016/j.ins.2019.06.064}{Three-way confusion matrix for classification: A measure driven view},
Information Sciences, 507 (2020) 772-794.

\bibitem{Yao_2012}
Y. Yao, 
\href{https://doi.org/10.1007/978-3-642-32115-3_1}{An outline of theory of three-way decisions}, 
in: J. Yao et al. (Eds.) Rough Sets and Current Trends in Computing (RSCTC 2012), LNCS 7413, Springer, Heidelberg, 2012, pp. 1-7.

\bibitem{Yao_2018}
Y. Yao,
\href{https://doi.org/10.1016/j.ijar.2018.09.005}{Three-way decision and granular computing}, 
International Journal of Approximate Reasoning, 103 (2018) 107-123.

\bibitem{Yao_2019}
Y. Yao, 
\href{https://doi.org/10.1016/j.knosys.2019.05.016}{Three-way conflict analysis: Reformulations and extensions of the Pawlak model},
Knowledge-Based Systems, 180 (2019) 26-37.

\bibitem{Yao_2021}
Y. Yao, 
\href{https://doi.org/10.1007/s10489-020-02142-z}{The geometry of three-way decision},
Applied Intelligence, 51 (2021) 6298-6325.

\bibitem{Yao_Hu_2018}
Y. Yao, M. Hu, X. Deng,
\href{https://doi.org/10.1007/978-3-319-91476-3_59}{Modes of sequential three-way classifications},
in: J. Medina et al. (Eds.) Information Processing and Management of Uncertainty in Knowledge-Based Systems (IPMU 2018), CCIS 854, Springer, Cham, 2018, pp. 724-735.

\bibitem{Yao_Wang_2017}
Y. Yao, S. Wang, X. Deng,
\href{https://doi.org/10.1016/j.ins.2017.05.036}{Constructing shadowed sets and three-way approximations of fuzzy sets},
Information Sciences, 412 (2017) 132-153.

\bibitem{Yang_2021}
J. Yang, Y. Yao, 
\href{https://doi.org/10.1016/j.ins.2021.06.065}{A three-way decision based construction of shadowed sets from Atanassov intuitionistic fuzzy sets}, 
Information Sciences, 577 (2021) 1-21.

\bibitem{Yue_2020}
X. Yue, Y. Chen, D. Miao, H. Fujita,
\href{https://doi.org/10.1016/j.ins.2018.07.065}{Fuzzy neighborhood covering for three-way classification},
Information Sciences, 507 (2020) 795-808.

\bibitem{Zhang_2020}
Y. Zhang, J. Yao,
\href{https://doi.org/10.1016/j.ins.2018.07.058}{Game theoretic approach to shadowed sets: a three-way tradeoff perspective},
Information Sciences, 507 (2020) 540-552.

\bibitem{Zhang_2021} 
X. Zhang, H. Yao, Z. Lv, D. Miao, 
\href{https://doi.org/10.1016/j.ins.2021.01.080}{Class-specific information measures and attribute reducts for hierarchy and systematicness}, 
Information Sciences, 563 (2021) 196-225.

\bibitem{Zhao_2020}
X. Zhao, B. Hu,
\href{https://doi.org/10.1016/j.ins.2018.08.024}{Three-way decisions with decision-theoretic rough sets in multiset-valued information tables},
Information Sciences, 507 (2020) 684-699.

\bibitem{Zhao_2019}
X. Zhao, Y. Yao, 
\href{https://doi.org/10.1016/j.ins.2019.05.022}{Three-way fuzzy partitions defined by shadowed sets},
Information Sciences, 497 (2019) 23-37.

\bibitem{Zhi_2019}
H. Zhi, J. Qi, T. Qian, R. Ren,
\href{https://doi.org/10.1016/j.ins.2019.12.065}{Conflict analysis under one-vote veto based on approximate three-way concept lattice},
Information Sciences, 516 (2020) 316-330.

\bibitem{Zhou_2021}
J. Zhou, W. Pedrycz, C. Gao, Z. Lai, X. Yue,
\href{https://doi.org/10.1016/j.fss.2020.06.019}{Principles for constructing three-way approximations of fuzzy sets: A comparative evaluation based on unsupervised learning}, 
Fuzzy Sets and Systems, 413 (2021) 74-98.

\end{thebibliography}
\end{document}